%% file: acl_latex.tex
\documentclass[11pt]{article}

\usepackage[final]{acl}

\usepackage{times}
\usepackage{latexsym}

\usepackage[T1]{fontenc}
\usepackage{iftex}
\ifPDFTeX
  \usepackage[utf8]{inputenc}
\fi
\usepackage{inconsolata}

\usepackage{graphicx}
\usepackage{booktabs}
\usepackage{tabularx}
\usepackage{multirow}
\usepackage{multicol}
\usepackage{makecell}
\usepackage{colortbl}
\usepackage{subcaption}

\usepackage{amsmath}
\usepackage{amssymb}
\usepackage{amsfonts}
\usepackage{nicefrac}
\usepackage{bm}

\usepackage{xcolor}
\usepackage{textcomp}
\usepackage{marvosym}
\usepackage{pifont}
\usepackage{enumitem}
\usepackage{xurl}
\usepackage{placeins} 
\usepackage{stfloats} 

\usepackage[capitalize]{cleveref}

\usepackage{tcolorbox}
\usepackage[stretch=40,shrink=40,step=1]{microtype}

\newcommand{\eg}{\emph{e.g.}}

\newcommand{\cv}{\textit{S}\textsubscript{cv. }}
\newcommand{\avg}{\textit{S}\textsubscript{avg. }}

\title{Benchmarking and Boosting Multilingual Capabilities of LVLMs via OCR-Centric Reinforcement Learning}

\newcommand{\projecturl}{https://github.com/opendatalab/PM4Bench}

\author{
  \textbf{Junyuan Gao\textsuperscript{1,*}},
  \textbf{Jiahe Song\textsuperscript{2,1,*}},
  \textbf{Jiang Wu\textsuperscript{1,\ensuremath{\ddagger}}},
  \textbf{Runchuan Zhu\textsuperscript{3}}
\\
  \textbf{Guanlin Shen\textsuperscript{1}},
  \textbf{Shasha Wang\textsuperscript{1}},
  \textbf{Xingjian Wei\textsuperscript{1}},
  \textbf{Haote Yang\textsuperscript{1}}
\\
  \textbf{Weijia Li\textsuperscript{1,4}},
  \textbf{Bin Wang\textsuperscript{1}},
  \textbf{Lijun Wu\textsuperscript{1}},
  \textbf{Conghui He\textsuperscript{1,\ensuremath{\dagger}}}
\\[2pt]
  \textsuperscript{1}Shanghai Artificial Intelligence Laboratory
\\
  \textsuperscript{2}School of Artificial Intelligence, Shanghai Jiao Tong University
\\
  \textsuperscript{3}National University of Singapore
  \quad
  \textsuperscript{4}SIGS, Tsinghua University
\\[2pt]
  \footnotesize{\texttt{\{gaojunyuan,songjiahe,wujiang,shenguanlin,wangshasha\}@pjlab.org.cn}}
  \\[-2.5pt]
  \footnotesize{\texttt{\{weixingjian,yanghaote,wangbin,wulijun,heconghui\}@pjlab.org.cn}}
  \\[-2.5pt]
  \footnotesize{\texttt{zhurunchuan0828@gmail.com};
  \texttt{liweijia@sz.tsinghua.edu.cn}}
}

\begin{document}
\maketitle
\begingroup
\renewcommand{\thefootnote}{\fnsymbol{footnote}}
\footnotetext[1]{\raggedright Equal contribution.
\textsuperscript{\ensuremath{\dagger}} Corresponding author.
\textsuperscript{\ensuremath{\ddagger}} Project lead. Code and Data:
\href{\projecturl}{\nolinkurl{github.com/opendatalab/PM4Bench}}.}
\endgroup

\begin{abstract}
Evaluating the multilingual capabilities of Large Vision-Language Models (LVLMs) remains challenging because most benchmarks rely on non-parallel corpora, making it unclear whether cross-lingual performance gaps reflect model limitations or dataset inconsistencies. To address this, we introduce \textbf{PM\textsuperscript{4}Bench}, the first multimodal, multilingual, multi-task benchmark built on a strictly parallel 10-language corpus, enabling \textbf{fair, apples-to-apples cross-lingual comparison} of model performance. We further introduce a \textbf{\texttt{vision} setting} that embeds textual inputs directly into images, better approximating deployment scenarios where LVLM-driven agents interact with virtual or physical environments through unified visual observations. Experiments with 10 LVLMs reveal that \textbf{OCR is a key factor behind cross-lingual disparity} when textual content is rendered visually. Motivated by this, we design an \textbf{OCR-centric GRPO training strategy} using fully synthesized, label-free OCR data, without expensive task-specific VQA supervision. The resulting model improves general multilingual VQA capability, reduces cross-lingual disparities under the \texttt{vision} setting, and transfers gains beyond PM\textsuperscript{4}Bench. This methodology offers an efficient, label-free pathway toward more equitable multilingual deployment of LVLM-driven agents.
\end{abstract}

\input{Chapter/1_Introduction.tex}
\input{Chapter/2_Related_Work.tex}

\input{Chapter/3_Benchmark.tex}
\input{Chapter/4_Training}
\input{Chapter/5_Experiment}
\input{Chapter/6_Results}
\input{Chapter/7_Conclusion}

\FloatBarrier 

\setlength{\parskip}{0pt plus 1pt}

\input{Chapter/8_Limitations}

\section*{Acknowledgments}

This work was supported by Shanghai Artificial Intelligence Laboratory.

\paragraph{Use of generative AI.}
Generative AI coding assistants, including Claude Code using Claude Opus 4.6
and OpenAI Codex using GPT-5.6-sol, were used to assist with code and shell scripting, remote
experiment orchestration, result aggregation and checking, and drafting and
editing parts of the manuscript and rebuttal. All generated outputs,
experimental results, references, claims, and final text were reviewed and
verified by the authors, who take full responsibility for the work.

\bibliography{custom}

\clearpage
\appendix
\crefalias{section}{appendix}
\crefalias{subsection}{appendix}
\Crefname{appendix}{Appendix}{Appendices}
\crefname{appendix}{appendix}{appendices}
\input{Chapter/Appendix_A}
\input{Chapter/Appendix_B}
\input{Chapter/Appendix_C}
\input{Chapter/Appendix_D}
\input{Chapter/Appendix_E}
\input{Chapter/Appendix_F}
\input{Chapter/Appendix_G}
\input{Chapter/Appendix_H}

\end{document}

%% file: Chapter/1_Introduction.tex
\section{Introduction}
\begin{figure*}[ht]
  \centering
  \includegraphics[width=\textwidth]{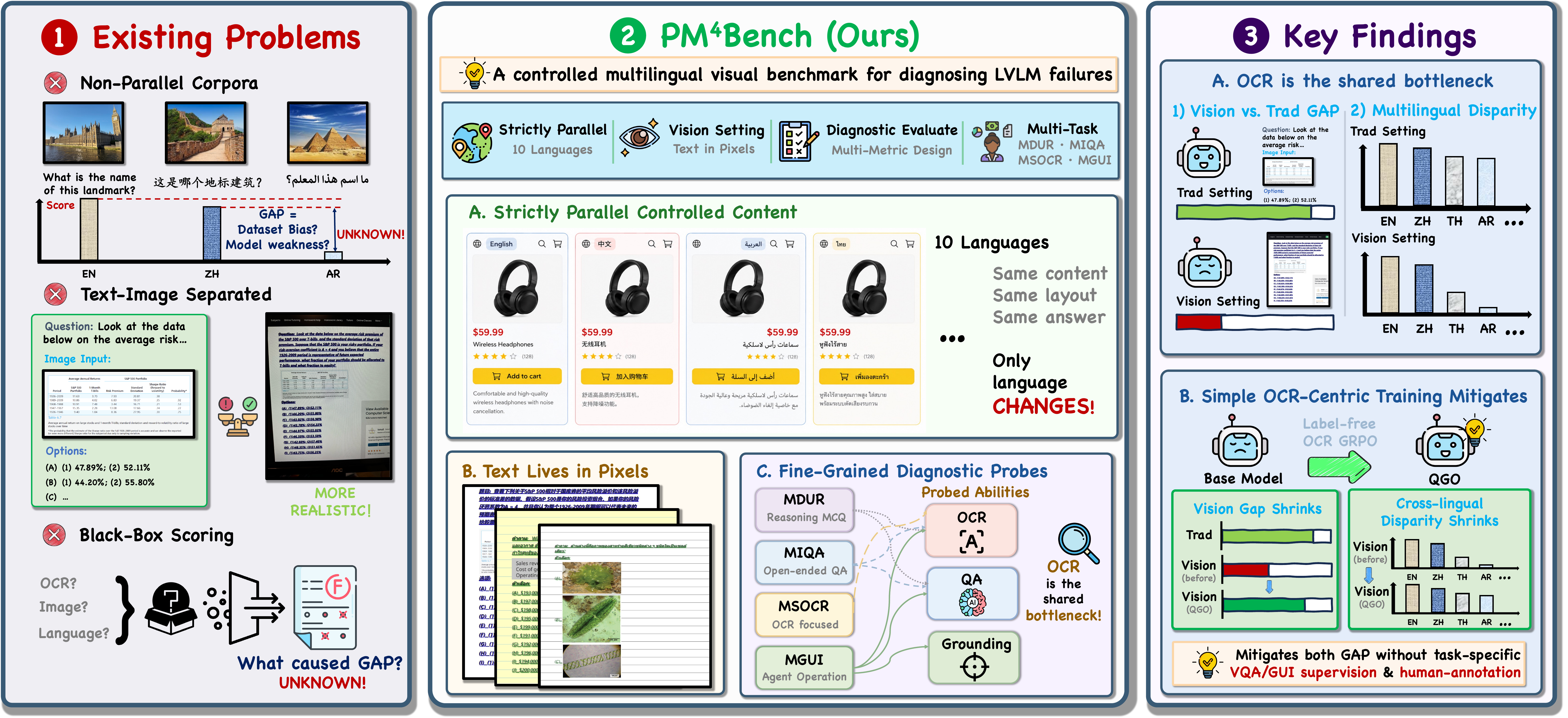}
  \caption{Overview of PM\textsuperscript{4}Bench. \textbf{(1) Existing problems} with current multilingual LVLM benchmarks: \emph{non-parallel corpora} confound model capability with content variance across languages; \emph{text-image separation} diverges from realistic deployment, where agents observe unified visual interfaces; and \emph{lack of cause analysis} leaves the drivers of cross-lingual disparity and \texttt{vision}-setting performance drops undiagnosed. \textbf{(2) PM\textsuperscript{4}Bench (ours)}: (A) strictly parallel content fixes semantics, layout, and answer across 10 languages so only the language varies; (B) a \texttt{vision} setting embeds text inside the image, matching how agents perceive real interfaces; (C) four complementary tasks jointly measure cross-lingual consistency from multiple dimensions. \textbf{(3) Key findings}: OCR is a shared cross-lingual bottleneck under \texttt{vision}, motivating our label-free OCR-centric GRPO recipe whose gains transfer to external benchmarks (CC-OCR, MTVQA, OCRBench).}
  \label{fig:fig1}
\end{figure*}

Large Language Models (LLMs) excel at QA, reasoning, and instruction following, yet performance gaps persist across languages even on language-agnostic tasks like math and code. Prior work attacks the problem from multiple angles: model mechanisms~\cite{wendler2024llamas, tang2024language, zhao2024large}, multilingual corpora~\cite{xue2021mt5massivelymultilingualpretrained, yu2025wanjuansiluhighqualityopensourcewebtext}, training and inference techniques~\cite{zhu2024power,she2024mapo,zhu2024question,shi2022language, huang2023not}, and benchmarks~\cite{sun2024benchmarking, zhang2024p, huang2025benchmax}.

Large Vision Language Models (LVLMs) extend this picture to multimodal input and now serve as the perceptual backbone of \textbf{multimodal AI agents} that read smartphone, web, and desktop interfaces as images. LVLMs inherit cross-lingual disparities from LLMs and add new ones, notably imbalanced text recognition across scripts, which results in uneven user experience once these agents are deployed globally.

A comprehensive evaluation of LVLMs in multilingual scenarios is essential for identifying shortcomings and guiding optimization, yet most existing benchmarks fall short on two axes. (1) \textbf{Non-parallel corpora}: which introduces uncontrolled variance across languages, making it difficult to isolate whether performance gaps among languages stem from differences in corpus content or fundamental model capabilities. (2) \textbf{Text and images are processed as separate streams}, which differs from realistic deployment scenarios where AI agents must interpret unified visual observations while interacting with virtual or physical environments.

To address these gaps, we introduce \textbf{PM\textsuperscript{4}Bench}, to our knowledge the first multilingual, multimodal, multi-task benchmark that simultaneously satisfies: (i) a \textbf{strictly parallel 10-language corpus}, where semantically identical content is rendered in every language, enabling apples-to-apples cross-lingual comparison; (ii) a \textbf{\texttt{vision} setting} that fuses text and images into a single visual input, better approximating real-world usage scenarios; and (iii) a \textbf{multi-perspective multi-task evaluation suite} that probes diverse capability dimensions, spanning reasoning VQA, open-ended generation, multi-scale OCR, and coordinate-based MGUI grounding. Taken together, these properties let us probe per-language performance on identical content and reason about cross-lingual gaps on equal footing.

Using PM\textsuperscript{4}Bench, we evaluate 10 LVLMs covering leading commercial APIs and open-source models, and observe pronounced performance drops alongside sharper cross-lingual disparities under the \texttt{vision} setting. Since \texttt{vision} differs from \texttt{traditional} only in that text now lives \emph{inside} the image, OCR is the natural candidate for explaining the gap. Further analysis supports this in \S\ref{sec:causes_of_gap}: a controlled experiment that supplies ground-truth text alongside the vision input visibly narrows the gap, and per-language OCR task scores correlate strongly with VQA performance. Together they identify \textbf{OCR as a key bottleneck} behind cross-lingual disparity and performance drop. Base-model capability remains the upper bound, but OCR is a high-leverage axis we can intervene on cheaply.

Motivated by this, we propose an OCR-centric reinforcement learning recipe using Group Relative Policy Optimization (GRPO)~\cite{deepseek-math} on \texttt{Qwen3-VL-8B-Thinking}. The core design is a \textbf{dual reward} that strengthens OCR while preserving the model's general capability: an OCR accuracy term sharpens recognition, and a length/format term keeps the chain of thought intact, since pure-accuracy optimization otherwise collapses CoT and degrades reasoning-heavy tasks (\cref{tab:QGO_detail}). \textbf{Trained on fully synthesized, label-free OCR data, the resulting model improves performance across tasks, settings, and languages on PM\textsuperscript{4}Bench, and the gains transfer to multilingual benchmarks beyond PM\textsuperscript{4}Bench, evidencing genuine generalization.} Upgrading fundamental OCR skills is thus a scalable, low-cost pathway to boost the model's multilingual performance prior to release.

Overall, our work closes the loop between evaluation and improvement for more equitable global deployment of LVLMs: we provide a fair evaluation yardstick, identify a high-leverage optimization direction, and demonstrate a low-cost mitigation strategy.
\begin{itemize}
\item \textbf{From evaluation gap to benchmark construction.} We introduce \textbf{PM\textsuperscript{4}Bench}, the first 10-language strictly parallel-corpus benchmark with a \textbf{\texttt{vision} setting} that fuses text and images into a single visual input, enabling apples-to-apples cross-lingual evaluation from diverse capability dimensions.

\item \textbf{From benchmark evaluation to bottleneck diagnosis.} We evaluate 10 LVLMs and find that the \texttt{vision} setting is harder than the \texttt{traditional} interleaved-input, exhibits sharper cross-lingual disparities on all tasks and capability dimensions. Further analysis identifies \textbf{OCR capability} as a key bottleneck behind these disparities.

\item \textbf{From bottleneck diagnosis to targeted mitigation.} We propose an OCR-centric GRPO training strategy using \textbf{label-free synthetic OCR data}, improving performance across tasks, settings, and languages on PM\textsuperscript{4}Bench while yielding transferable gains in general multilingual capabilities beyond PM\textsuperscript{4}Bench.
\end{itemize}

%% file: Chapter/2_Related_Work.tex
\section{Related Work}

\textbf{LVLM Benchmark} Early benchmarks~\cite{MMBench,fu2023mme,li2023seed} relied on MCQ or short VQA; recent work shifts to open-ended evaluation (MMDU~\cite{liu2024mmdu}) and pure visual input (MMMU-pro~\cite{yue2024mmmu}).

\textbf{Multilingual Benchmark} Many multilingual LLM benchmarks are built by translating English evaluation items, as in MGSM~\cite{shi2022language}; naturally occurring multilingual resources such as XL-Sum~\cite{hasan-etal-2021-xl} avoid this route, while recent efforts like P-MMEval~\cite{zhang2024p} and BenchMAX~\cite{huang2025benchmax} use parallel corpora for fair cross-lingual comparison.

\textbf{Multilingual LVLM Benchmark} Existing multilingual LVLM benchmarks evaluate visual question answering and reasoning~\cite{pfeiffer2021xgqa,liu2021marvl,changpinyo2022maxm,bugliarello2022iglue}, text-centric VQA~\cite{tang2024mtvqa}, exam reasoning (M3Exam~\cite{zhang2023m3exam}, EXAMS-V~\cite{das2024exams}), and multilingual or culturally grounded content (CVQA~\cite{romero2024cvqa}, M5-VLOD~\cite{schneider2024m5}, ALM-bench~\cite{vayani2024all}). Non-parallel corpora, however, conflate language ability with data artifacts, while parallel-corpus benchmarks (M4U~\cite{wang2024m4u}, MMMB~\cite{sun2025parrot}, XT-VQA~\cite{yu2024cross}) retain mixed text-image inputs and lack matched pure-vision evaluation.

\textbf{Multilingual OCR as a Training Driver} From an orthogonal model-development perspective, Centurio~\cite{geigle-etal-2025-centurio} studies multilingual training mixtures across 43 languages and finds that non-English OCR data in both pre-training and instruction tuning are critical for multilingual text-in-image understanding. This complements our evaluation-side evidence: Centurio asks which training data create the capability, whereas PM\textsuperscript{4}Bench holds content parallel and uses matched input settings, ground-truth-text injection, and OCR-centric post-training to diagnose recognition as the bottleneck behind cross-lingual downstream gaps. The convergence connects a training-data driver with a controlled behavioral mechanism.

\textbf{GUI-Agent Benchmarks and the Multilingual Gap} GUI grounding and interaction have become core capabilities for multimodal agents, evaluated across ScreenSpot~\cite{cheng2024seeclick}, ScreenSpot-Pro~\cite{li2025screenspotpro}, Mind2Web~\cite{deng2023mind2web}, VisualWebArena~\cite{koh2024visualwebarena}, and OSWorld~\cite{xie2024osworld}; these resources are overwhelmingly English-only. MPR-GUI~\cite{chen2025mprgui} contributes parallel multilingual GUI \emph{perception and reasoning} but no coordinate grounding. macOSWorld~\cite{yang2025macosworld} provides a comprehensive 5-language interactive macOS environment with coordinate-based GUI interaction. However, studying cross-lingual capability \emph{imbalance} requires a corpus that is parallel both visually and linguistically across runs, which an interactive multi-step environment cannot maintain because the UI state diverges across trajectories.

As summarized in~\cref{tab:related_benchmarks}, to our knowledge no prior
benchmark combines a strictly parallel multilingual corpus, pure-vision input,
open-ended generation, OCR diagnostics, and coordinate-based GUI grounding in
one controlled suite. Pooling non-parallel benchmarks would reintroduce the
content and source confounds that the verified parallel construction is
designed to remove.

\input{Table/related_benchmark}

%% file: Table/related_benchmark.tex
\begin{table*}[t]
\centering
\scriptsize
\setlength{\tabcolsep}{3pt}
\renewcommand{\arraystretch}{1.05}
\begin{tabular*}{\textwidth}{l @{\extracolsep{\fill}} c c c c c c c}
\toprule
\textbf{Benchmark} &
\textbf{\#Lang.} &
\textbf{\makecell{Parallel\\Corpus}} &
\textbf{\makecell{Pure-Vision\\Input}} &
\textbf{\makecell{Knowledge/\\Reason.}} &
\textbf{\makecell{Open-end.\\Gen.}} &
\textbf{\makecell{OCR\\Challenge}} &
\textbf{\makecell{Coord.\\GUI}} \\
\midrule
xGQA~\cite{pfeiffer2021xgqa}        & 8   & Q \checkmark, A \textcolor{red}{\ding{55}}\textsuperscript{*} & \texttimes & \checkmark & \texttimes & \texttimes & \texttimes \\
MaRVL~\cite{liu2021marvl}           & 5   & \texttimes & \texttimes & \checkmark & \texttimes & \texttimes & \texttimes \\
XVNLI~\cite{bugliarello2022iglue}   & 5   & \checkmark & \texttimes & \checkmark & \texttimes & \texttimes & \texttimes \\
xFlickrCO~\cite{bugliarello2022iglue} & 8 & \texttimes & \texttimes & \texttimes & \texttimes & \texttimes & \texttimes \\
MaXM~\cite{changpinyo2022maxm}      & 7   & \texttimes & \texttimes & \checkmark & \checkmark & \texttimes & \texttimes \\
M3Exam~\cite{zhang2023m3exam}       & 9   & \texttimes & \texttimes & \checkmark & \texttimes & \texttimes & \texttimes \\
EXAMS-V~\cite{das2024exams}         & 11  & \texttimes & \checkmark & \checkmark & \texttimes & \texttimes & \texttimes \\
MTVQA~\cite{tang2024mtvqa}          & 9   & \texttimes & \texttimes & \checkmark & \checkmark & \checkmark & \texttimes \\
CVQA~\cite{romero2024cvqa}          & 31  & \texttimes & \texttimes & \checkmark & \texttimes & \texttimes & \texttimes \\
M4U~\cite{wang2024m4u}              & 6   & T \checkmark, I \textcolor{red}{\ding{55}} & \texttimes & \checkmark & \texttimes & \texttimes & \texttimes \\
MMMB~\cite{sun2025parrot}           & 6   & \checkmark & \texttimes & \checkmark & \texttimes & \texttimes & \texttimes \\
M5-VLOD~\cite{schneider2024m5}      & 12  & \texttimes & \texttimes & \checkmark & \texttimes & \texttimes & \texttimes \\
ALM-bench~\cite{vayani2024all}      & 100 & \checkmark & \texttimes & \checkmark & \checkmark & \texttimes & \texttimes \\
XT-VQA~\cite{yu2024cross}           & 3   & T \checkmark, I \textcolor{red}{\ding{55}} & \texttimes & \checkmark & \checkmark & \checkmark & \texttimes \\
macOSWorld~\cite{yang2025macosworld} & 5  & \checkmark\textsuperscript{$\ddagger$} & \texttimes & \checkmark & \texttimes & \texttimes & \checkmark\textsuperscript{$\ddagger$} \\
MPR-GUI~\cite{chen2025mprgui}       & 6   & \checkmark & \texttimes & \checkmark & \texttimes & \texttimes & \texttimes \\
\midrule
\textbf{PM\textsuperscript{4}Bench (ours)} & \textbf{10} & \ding{51} & \ding{51} & \ding{51} & \ding{51} & \ding{51} & \ding{51} \\
\bottomrule
\end{tabular*}
\caption{Comparison of PM\textsuperscript{4}Bench with related multilingual multimodal benchmarks. Knowledge/Reason.: \checkmark if any included task uses knowledge or reasoning. \textsuperscript{*}: translated questions, English answers. \textsuperscript{$\ddagger$}: macOSWorld has parallel instructions and coordinate-based actions; multi-step runs can diverge in UI state.}
\label{tab:related_benchmarks}
\end{table*}

%% file: Chapter/3_Benchmark.tex
\section{\texorpdfstring{PM\textsuperscript{4}Bench}{PM4Bench}}

\subsection{Design Principles}
\label{sec: 3__design_principles}

PM\textsuperscript{4}Bench is built on four principles: (1) \textbf{Targeted Language Selection}, covering diverse families and writing scripts; (2) \textbf{Parallel Corpus}, semantically identical content across languages so any score gap reflects model capability rather than corpus variance; (3) \textbf{Vision Setting}, with text and queries ``printed'' onto images to simulate real applications; (4) \textbf{Task Diversity}, spanning competencies relevant to LVLM and agent deployment. Rendering-side confounds are controlled by sharing identical templates, font families, decorations, and backgrounds across languages; inherent script-level differences (length, density, direction) are deliberately preserved so scores reflect real-world deployment challenges.

\subsection{Language Selection}
PM\textsuperscript{4}Bench supports 10 carefully selected languages: en, zh, ko, th, vi, ru, hu, sr, cs, ar. We quantify their graphic complexity following GraphCom~\cite{chang2018graphcom}; details are provided in \cref{sec: appendix_c}.

\subsection{Task Introduction}

Following~\cref{sec: 3__design_principles}, PM\textsuperscript{4}Bench includes a parallel 10-language corpus across four tasks: Multi-Discipline Understanding and Reasoning (MDUR), Multi-Image Question Answering (MIQA), Multi-Scale OCR Challenge (MSOCR), and Multilingual GUI Grounding (MGUI). Task statistics are summarized in~\cref{tab:task_cmp}; per-task examples are in \cref{sec: appendix_a}.

\input{Table/task_compare}

\textbf{(1) Multi-Discipline Understanding and Reasoning (MDUR)} assesses college-level multimodal reasoning. We source 1,730 English multiple-choice questions from open-sourced MMMU-pro~\cite{yue2024mmmu} and translate them into 9 additional languages (details in~\cref{sec: 3__benchmark_construction}). Since MMMU-pro's native vision-only variant is English-restricted, we built an HTML-to-screenshot pipeline that embeds text and images into webpages with randomized fonts and backgrounds, guaranteeing identical layouts across languages. MDUR yields 1,730 cases per language under both \texttt{traditional} (interleaved) and \texttt{vision} (single rendered image) input formats.

\textbf{(2) Multi-Image Question Answering (MIQA)} targets open-ended QA over multi-image inputs. We sampled 218 QA pairs from MMDU~\cite{liu2024mmdu}, a Wikipedia-sourced multi-turn multi-image benchmark, and translated them into 9 languages. Both \texttt{traditional} and \texttt{vision} settings are provided; for \texttt{vision}, question text and reference images are placed on a 1280-pixel-wide canvas.

\textbf{(3) Multi-Scale OCR Challenge (MSOCR)} probes multilingual text-recognition limits. Each image contains 20 lines of semantically meaningless words from Wikipedia parallel corpora rendered with decreasing font sizes (40 down to 2); the recognition threshold is the first line where errors occur. We provide 100 images per language, semantically identical across languages. MSOCR offers a strictly parallel visual corpus and a fully automated, easily extensible construction; in our cross-lingual-balance focus, this lets us evaluate the OCR dimension cheaply and apples-to-apples across languages.

\textbf{(4) Multilingual GUI Grounding (MGUI)} evaluates an LVLM as a multilingual GUI agent, following the coordinate-based grounding paradigm of ScreenSpot~\cite{cheng2024seeclick}. We create 100 reusable HTML/Jinja webpage templates spanning 10 common product surfaces (\eg, flight search, seat map, payment; full list in~\cref{tab:gui_topics}) and programmatically render each in all 10 languages on a fixed $1280\times800$ layout. Each image is paired with two instructions, so each language has 200 test items in total. Every natural-language instruction is indirect (\eg, ``\textit{I'm emigrating abroad and won't need a return ticket, switch the trip type so I'm only booking the outbound leg.}''), without leaking the text label of the target button, and the model must emit a single click coordinate $(x, y)$ inside the target element's bounding box. Once the templates are specified, rendering and language instantiation are automated and human-translation-free, making MGUI inexpensive to scale.

\subsection{Settings}
\label{sec: task settings}

To better reveal the capabilities of LVLMs and help in-depth analysis, we adopt diverse settings across tasks (see~\cref{tab:task_cmp} for a summary). \textbf{(1) \texttt{Vision} setting.} The LVLM receives a single image containing all the information necessary to complete the task; the text prompt is minimal, typically a concise instruction such as ``Answer the question in the image.'' This setup encompasses all four tasks. \textbf{(2) \texttt{Traditional} setting.} The textual content of the questions and the embedded images are provided as separate inputs (MSOCR and MGUI are excluded by construction, as they only meaningfully exist as a single image).

\subsection{Translation Pipeline}
\label{sec: 3__benchmark_construction}

\begin{figure}[t]
  \centering
  \includegraphics[width=\linewidth]{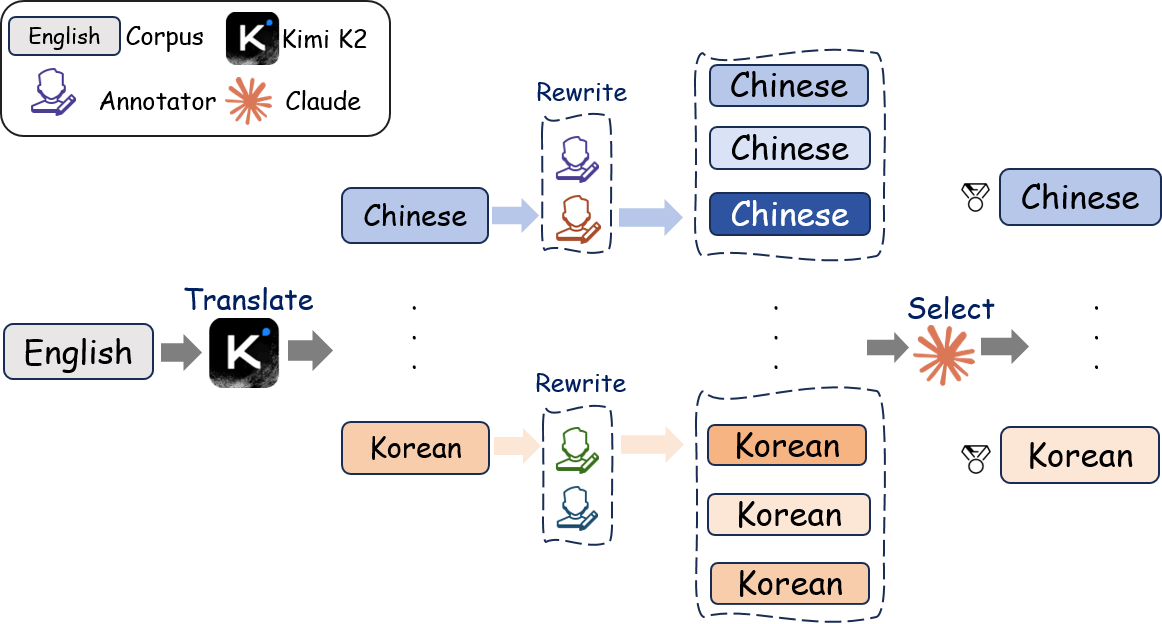}
  \caption{The translation of PM\textsuperscript{4}Bench's parallel texts involves three steps: 1) Kimi K2 Reference Generation; 2) Human Rewriting; 3) Claude Selection.}
  \label{fig:translation_process}
\end{figure}

To ensure data quality, we adopt an LLM-and-human-expert in-loop pipeline for the parallel corpus of MDUR and MIQA. As shown in~\cref{fig:translation_process}, it has 3 stages: reference translation, manual translation, and selection.

We first use Kimi K2~\cite{kimiteam2025kimik2openagentic}, which is not in our evaluation set, to generate a reference translation. The original English plus the reference are then given to two independent groups of native-speaker annotators (also proficient in English); each group internally verifies and submits its own translation. Finally, the original English plus the 3 candidates (1 machine + 2 human) are passed to Claude Sonnet 4.5~\cite{anthropic2025claudesonnet45} (also outside our evaluation set) to pick the best. For MIQA, 64\% of selected translations were human; for MDUR, this exceeded 99\%. The gap reflects task content: MDUR's items lean on rigorous numerical and domain-specific phrasing where minor mistranslations can flip correctness, so human candidates dominate; MIQA's plainer descriptive prose lets machine and human candidates compete on near-equal footing.

%% file: Table/task_compare.tex
\begin{table}[ht]
\centering
\small
\setlength{\tabcolsep}{4pt}
\renewcommand{\arraystretch}{1.1}

\begin{tabular*}{\columnwidth}{@{\extracolsep{\fill}} c c c c @{}}
\toprule

\textbf{Task} & \textbf{\makecell{\#sample\\(per lang.)}} & \textbf{QA Type} & \textbf{Settings} \\

\midrule

\textbf{MDUR} & 1730 & MCQ & trad.\//vis.  \\
\textbf{MIQA} & 218 & OE & trad.\//vis.  \\
\textbf{MSOCR} & 100 & OCR & vis. \\
\textbf{MGUI} & 200 & Coord. Grounding & vis. \\

\bottomrule
\end{tabular*}

\caption{The 4 tasks of PM\textsuperscript{4}Bench. Detailed explanation of trad.\//vis. settings is in~\cref{sec: task settings}.}
\label{tab:task_cmp}
\end{table}

%% file: Chapter/4_Training.tex
\section{OCR-centric GRPO Reinforcement Learning}
\label{sec: ocr_centric_rl}

Results in~\cref{sec:causes_of_gap} indicate that multilingual OCR robustness is a key factor behind cross-lingual disparity in LVLMs. To directly target this factor, we design a simple yet effective OCR-centric reinforcement learning strategy for \texttt{Qwen3-VL-8B-Thinking} based on Group Relative Policy Optimization (GRPO)~\cite{deepseek-math}, which boosts other non-OCR tasks and mitigates cross-lingual imbalance. We choose GRPO for this setting because we aim to improve OCR while preserving the base model's general-task reasoning. In our diagnostic comparison using the same synthetic OCR corpus, full-parameter and LoRA SFT degrade MDUR from their earliest audited checkpoints, whereas reward-constrained GRPO improves it (\cref{tab:post_training_comparison}).

\textbf{Synthetic multilingual OCR training data.} We automatically construct a synthetic OCR dataset without any human annotation. For each sample, we first choose a random background image and resolution/aspect ratio, then render multiple lines of text, one per language. Each line is composed of semantically meaningless noun combinations from Wikipedia, so that no language receives an advantage from specific semantics. Font style and font size are also randomly perturbed. Refer to \cref{sec: appendix_a_synth} for examples.

\textbf{GRPO reward design.} Our reward function is designed to enhance multilingual OCR capabilities without compromising general abilities. Let $s$ denote the full generated sequence (chain-of-thought plus final answer) with length $l = |s|$, and let $y$ be the answer extracted from the \verb|\boxed{}| span within $s$. The ground-truth is denoted by $y^*$. Our total reward consists of an accuracy reward $R_{\text{acc}}$ and a formatting reward $R_{\text{fmt}}$ that regulates output length.

\textbf{1) Format and Length Constraint $R_{\text{fmt}}$}
Two issues, identified through case studies on early training rollouts, motivate this term: overly long or under-structured outputs degrade general task performance, and pure OCR accuracy optimization causes the model to abandon chain-of-thought reasoning, degrading general performance. We define a target length interval $[L_{\min}, L_{\max}]$, write $\Delta_l=l-L_{\max}$ for excess length, and use $g(d)=\min(1,\log(1+d/\lambda))$, with scale parameter $\lambda$, for logarithmic growth. The penalty $P_{\text{len}}(l)$ is
\begin{equation}
\label{eq:length_penalty}
P_{\text{len}}(l) =
\begin{cases}
1 - l/L_{\min}, & l < L_{\min}, \\[2pt]
0, & L_{\min} \le l \le L_{\max}, \\[2pt]
g(\Delta_l), & l > L_{\max}.
\end{cases}
\end{equation}
Additionally, to prevent degenerate repetition, we compute the unique $4$-gram ratio $\rho(s) = |\mathcal{N}_4^{\text{uniq}}| / |\mathcal{N}_4^{\text{all}}|$. A penalty $P_{\text{rep}}(s)$ is applied if this diversity metric falls below a threshold $\tau$:
\begin{align}
P_{\text{rep}}(s) = \max\left(0, \; 1 - \frac{\rho(s)}{\tau}\right).
\label{eq:rep_penalty}
\end{align}
The combined formatting reward $R_{\text{fmt}}(s)$ encourages sequences within the target length while penalizing undesirable behaviors:
\begin{equation}
\label{eq:len_reward}
\begin{aligned}
R_{\text{fmt}}(s) ={}& \alpha_1 \mathbb{I}_{[L_{\min}, L_{\max}]}(l)
 - \alpha_2 P_{\text{len}}(l) \\
&{} - \alpha_3 P_{\text{rep}}(s).
\end{aligned}
\end{equation}
Here $\mathbb{I}_{[L_{\min}, L_{\max}]}(l)$ equals $1$ when $l$ lies in the target interval and $0$ otherwise. Thus, $\alpha_1$ supplies a constant positive reward within the interval, while $\alpha_2$ and $\alpha_3$ weight the length and repetition penalties.

\textbf{2) OCR Accuracy Reward $R_{\text{acc}}$}
We measure the correctness of the extracted answer $y$ using the normalized Levenshtein distance, denoted as $\mathcal{D}_{\text{lev}}(\cdot, \cdot)$:
\begin{align}
R_{\text{acc}}(y, y^*) = 1 - \frac{\mathcal{D}_{\text{lev}}(y, y^*)}{\max(|y|, |y^*|)}.
\label{eq:acc_reward}
\end{align}

\textbf{3) Total Reward.}
The final reward is a weighted sum emphasizing accuracy while maintaining structural constraints:
\begin{align}
R(s, y^*) = \omega_1 \cdot R_{\text{acc}}(y, y^*) + \omega_2 \cdot R_{\text{fmt}}(s),
\label{eq:final_reward}
\end{align}
where $\omega_1$ and $\omega_2$ are weighting hyper-parameters.


%% file: Chapter/5_Experiment.tex
\section{Experiment Settings}
\subsection{Evaluation Protocol}
To comprehensively compare the performance of various kinds of LVLMs, we include the following 10 models in our experiment. Leading commercial APIs: Gemini-3.1-Pro-Preview~\cite{google2025gemini3propreview}, GPT-5~\cite{openai2025gpt5systemcard}, GPT-5-mini~\cite{openai2025gpt5systemcard}, Doubao-Seed-1.6~\cite{bytedance2025doubaoseed16}; and open-source models: Llama-4-Maverick~\cite{meta2025llama4maverick}, GLM-4.5v~\cite{vteam2025glm45vglm41vthinkingversatilemultimodal}, and Qwen3-VL series~\cite{Qwen3-VL}. All models are evaluated with \textbf{greedy decoding} under default chat template. QGO-8B uses AdamW at $1\times10^{-6}$ with $32$ prompts and $8$ rollouts per prompt, yielding $256$ sampled trajectories per optimization step; rewards and group-relative advantages are computed per rollout. We select the step-$200$ checkpoint, reached in approximately $9$ hours on $8$ NVIDIA H200 GPUs. Detailed inference and training configurations are provided in \cref{sec: appendix_h_training_config}; per-task prompts are listed in \cref{sec: appendix_b_task_prompts}.

\subsection{Metrics}
For the performance score of MDUR, we evaluate the correct ratio $S^{\text{MDUR}} = N_{\text{cor}} / N_{\text{total}}$. For MIQA, we adopt the LLM-as-judge approach from MMDU, using DeepSeek-v4-flash~\cite{deepseekai2026deepseekv4} (outside our evaluation set and model families) to rate responses across six dimensions: \textit{Creativity}, \textit{Richness}, \textit{Visual Perception}, \textit{Logical Coherence}, \textit{Answer Accuracy}, \textit{Image Relationship Understanding}, against human-provided references. The final score averages these six dimensions:
$
S^{\text{MIQA}} = \frac{1}{N} \sum_{i=1}^{N} \left( \frac{1}{6} \sum_{d \in D} S_i^{(d)} \right)
$.
\cref{sec: appendix_e_translate_consistency} confirms the reliability of this judge against the disparity of language capability of judge model; \cref{sec: appendix_e_alt_judge} additionally reports a cross-validation with Claude-Opus-4.7~\cite{anthropic2026claudeopus47} as judge: per-language scores agree with DeepSeek-v4-flash judge with minor variance. The score for MSOCR is
$
S^{\text{MSOCR}} = \frac{1}{N} \sum_{i=1}^{N} (40 - s_i)
$,
where $s_i$ is the font size of the line where image $i$ first errs, and the constant $40$ equals the maximum font size, ensuring scores start at $0$. For MGUI, we measure click accuracy: each item asks the model to emit a coordinate $(x, y)$, and the prediction is correct if it lies inside the ground-truth bounding box,
$
S^{\text{MGUI}} = \frac{1}{N}\sum_{i=1}^{N}\mathbf{1}\big[(x_i, y_i) \in \mathrm{bbox}_i\big] \times 100
$,
following ScreenSpot~\cite{cheng2024seeclick}. Coordinates emitted in the conventional $1000\times1000$ normalized space are rescaled to the $1280\times800$ canvas before the bbox check.



%% file: Chapter/6_Results.tex
\section{Results \& Findings}

\subsection{Overall performance \& cross-lingual disparity}
\input{Table/overall_results}
\cref{tab:overall_res} reports overall results, with average score (\avg) and coefficient of variation (\cv = $\sigma/\mu \times 100\%$, measures cross-lingual variance) computed across 10 languages. Per-language results: \cref{sec: appendix_d}.

\begin{figure*}[ht]
  \centering
  \includegraphics[width=1\textwidth]{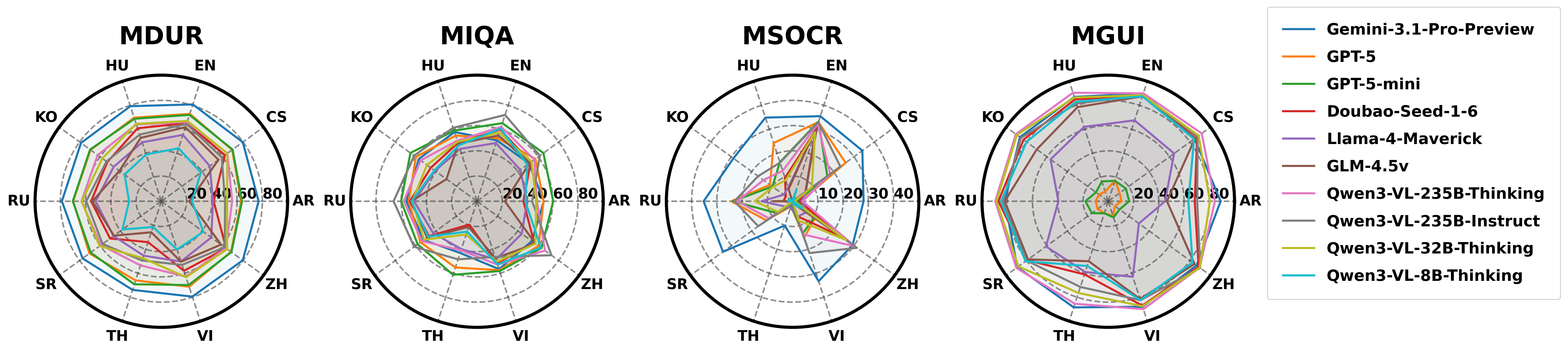}
  \caption{Scores (\texttt{vision} setting) of each model and each language across all four tasks of PM\textsuperscript{4}Bench.}
  \label{fig:compound_radar_chart}
\end{figure*}

\begin{figure*}[!htb]
  \centering
  \begin{subfigure}[t]{0.48\textwidth}
    \centering
    \includegraphics[width=\linewidth]{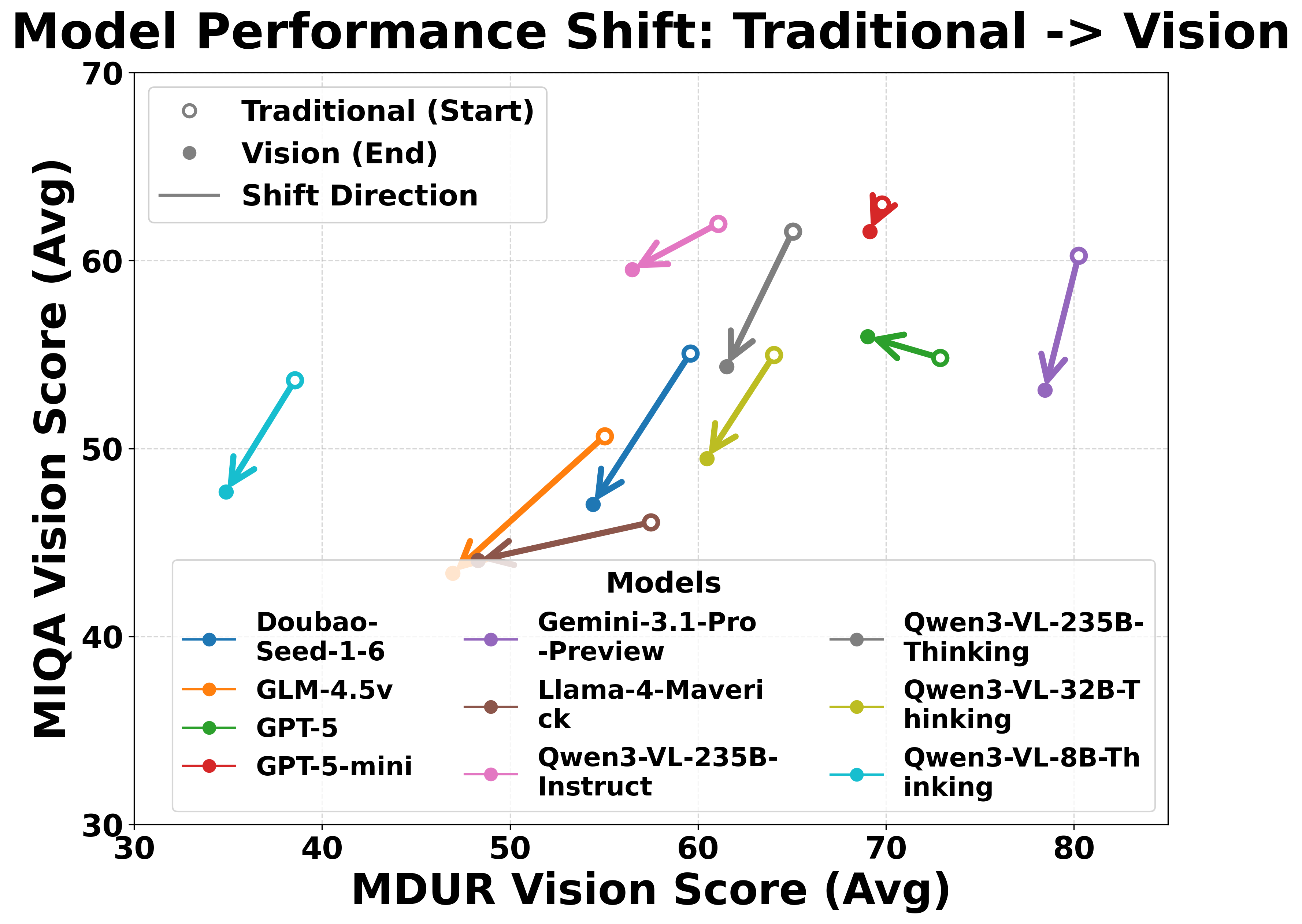}
    \caption{Performance shift from \texttt{traditional} to \texttt{vision} setting.}
    \label{fig:model_performance_shift}
  \end{subfigure}
  \hfill
  \begin{subfigure}[t]{0.48\textwidth}
    \centering
    \includegraphics[width=\linewidth]{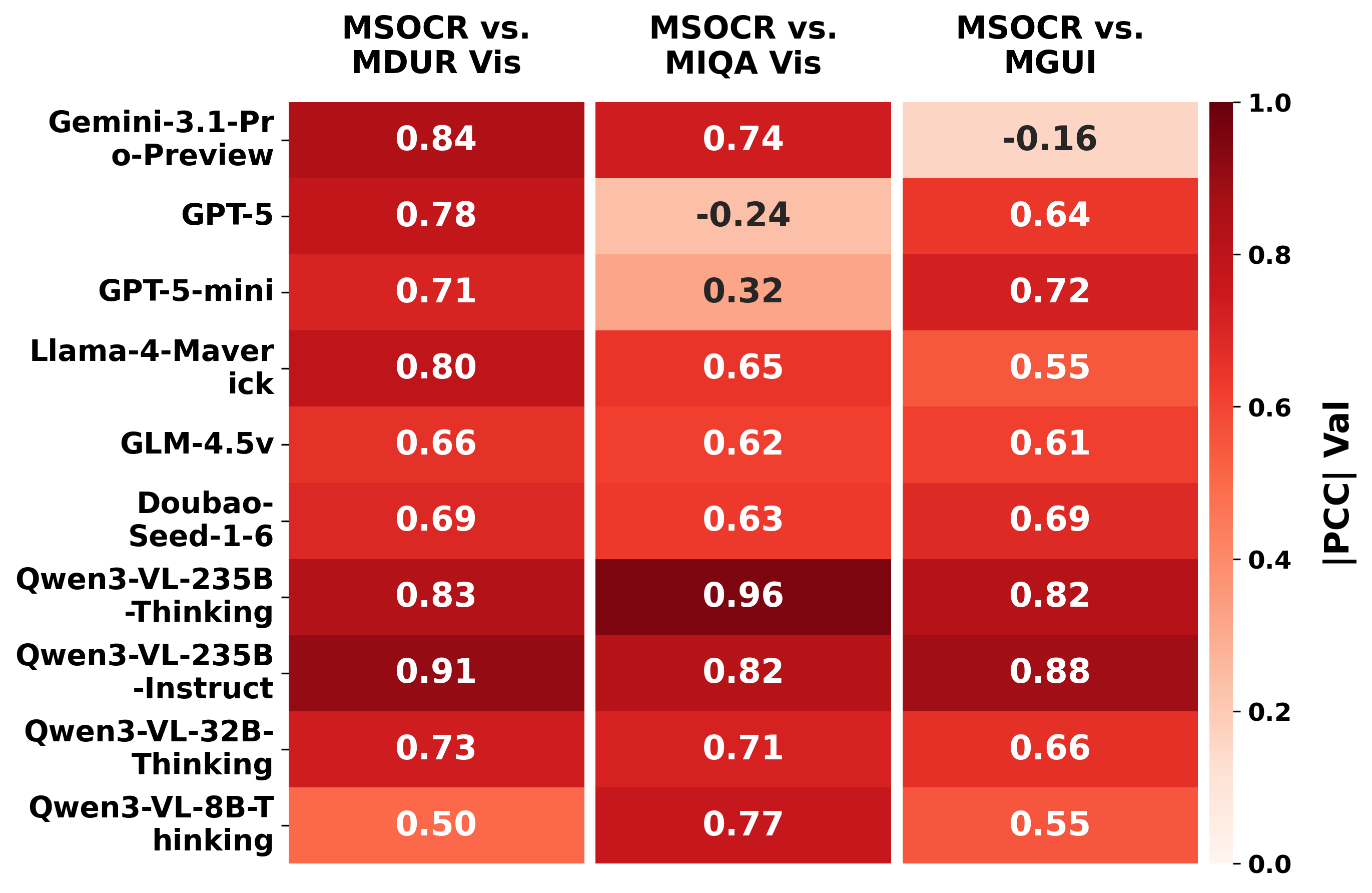}
    \caption{Per-model correlations between MSOCR and three downstream tasks across 10 languages; see \cref{sec: appendix_g_pcc}.}
    \label{fig:pcc_heatmap_main}
  \end{subfigure}
  \caption{Cross-setting performance shifts and the relationship between OCR
  and downstream multilingual performance.}
\end{figure*}

Gemini-3.1-Pro-Preview and the Qwen3-VL-235B series lead on most tasks. Yet \cref{fig:compound_radar_chart} shows \textbf{cross-lingual consistency remains a challenge}, with imbalance especially pronounced on MSOCR: every model exhibits a wide score spread across languages, and low-resource, visually dense scripts (AR, TH) consistently lag high-resource ones (EN, ZH) for both commercial APIs and open-source models. Under \texttt{vision} (\cref{fig:model_performance_shift}), \avg on MDUR/MIQA drops and 90\%/80\% of models show higher \cv than under \texttt{traditional}, so the \textbf{\texttt{vision} setting compromises both performance and cross-lingual balance}.

We transparently report an anomaly: GPT-5 and GPT-5-mini lead on MDUR/MIQA yet their MGUI accuracy collapses to $10$-$15\%$. We rule out base-capability deficit, evaluation bug, and data issue: a \textbf{MGUI\textsubscript{text}} sanity check (\cref{sec: appendix_d_gui_text}) shows GPT-5 reaches $94.0$ when asked for the visible text of the click target instead of a coordinate, and GPT-5's median bbox-edge distance is only $33$~px ($\sim$2.2\% of diagonal). These signals confirm GPT-5 \emph{understands} the MGUI task and \emph{correctly locates} the target; the failure is solely in emitting precise pixel coordinates, consistent with its strong MDUR/MIQA results.

\subsection{OCR is a key bottleneck causing performance gaps}
\label{sec:causes_of_gap}

Since \texttt{vision} differs from \texttt{traditional} mainly in that text now lives inside the image, OCR is the natural candidate for explaining the gap. We test this hypothesis with three converging signals.

\input{Table/QGO_detail}
\input{Table/vision_w_ocr_MDUR}

\textbf{(i) Controlled experiment.} Supplying ground-truth texts alongside \texttt{vision} input (\cref{tab:vis_w_ocr}) lifts \avg and lowers \cv for both commercial APIs and the Qwen3-VL series, with larger effects on smaller models.

\textbf{(ii) Per-language correlation.} Across the 10 languages, \textbf{most evaluated models exhibit $|\text{PCC}| > 0.5$} between MSOCR and other-task score vectors, indicating a strong correlation between text recognition and downstream VQA (\cref{fig:pcc_heatmap_main}).
\textbf{(iii) Causal intervention.} OCR-centric RL (\S\ref{sec: ocr_centric_rl}) lifts non-OCR tasks, confirming \textbf{OCR as a driver of cross-lingual performance gaps}.

\subsection{Effect of OCR-centric GRPO reinforcement learning}
\begin{figure*}[!tbp]
  \centering
  \includegraphics[width=\textwidth]{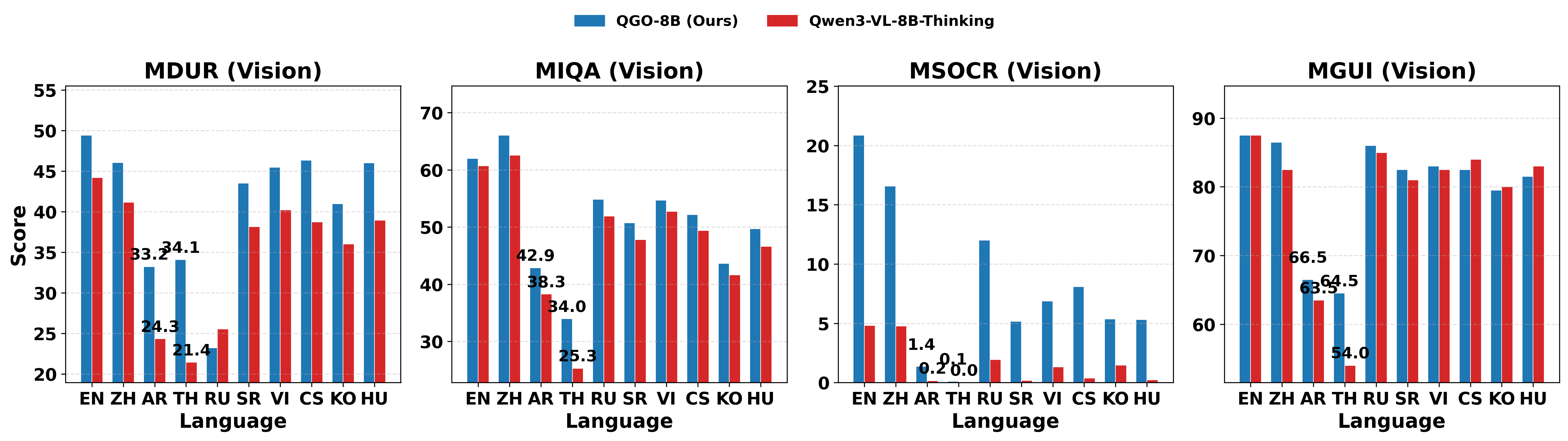}
  \caption{Per-language QGO-8B and base-model scores on four \texttt{vision}-setting tasks. Values are annotated for AR and TH, the baseline's weakest languages.}
  \label{fig:QGO_per_language_main}
\end{figure*}

Against Qwen3-VL-8B-Thinking, the trained model \textbf{QGO-8B improves \avg on every task and setting} (\cref{tab:overall_res}). It lowers \cv on all four \texttt{vision}-setting tasks and on MIQA under \texttt{traditional}; MDUR under \texttt{traditional} is the sole exception, where \cv rises from $0.061$ to $0.080$. The per-language comparison across all four \texttt{vision}-setting tasks (\cref{fig:QGO_per_language_main}) shows that the largest gains concentrate on the baseline's weakest languages AR and TH. Our synthetic training corpus contains no PM\textsuperscript{4}Bench evaluation examples. Gains on MDUR, MIQA, and MGUI therefore reflect cross-task transfer, while MSOCR shares only the underlying recognition objective.

\textbf{External-benchmark transfer.} On three external suites (\cref{sec: appendix_f}), QGO-8B improves \textbf{CC-OCR}~\cite{liu2024ccocr} by $+4.7$ NED overall ($+4.0$ over six unseen languages), matches the \textbf{OCRBench}~\cite{liu2024ocrbench} baseline ($791$ vs $789$), and gains $+0.5$ \avg on \textbf{MTVQA}~\cite{tang2024mtvqa} trained languages. CC-OCR improves in all 10 languages; OCRBench shows no broad regression; and MTVQA's unseen-language scores are essentially preserved ($-0.4$). Transfer is strongest for multilingual reading and more limited on text-centric QA.

\subsection{Ablation Study}

We isolate $R_{\text{fmt}}$ and $R_{\text{acc}}$ to verify both are essential. Their effects follow task structure (\cref{tab:QGO_detail}). \textbf{MDUR} requires both precise reading and sustained knowledge-intensive reasoning: \textit{only-fmt} gives a limited gain ($34.88\to36.81$), whereas \textit{only-ocr} truncates the chain of thought and collapses to $26.67$. \textbf{MIQA} contains shorter, less precision-sensitive visual text, so maintaining reasoning alone is more useful ($47.69\to49.48$), while the joint reward reaches $51.06$. \textbf{MSOCR} is pure recognition, making $R_{\text{acc}}$ the main driver ($1.53\to8.13$), with the full model at $8.17$. \textbf{MGUI} requires reasoning from indirect instructions to interface elements: \textit{only-ocr} falls from $78.30$ to $68.50$, while full QGO-8B reaches $80.00$. The joint reward therefore couples accurate perception with the reasoning behavior needed by downstream tasks.

\input{Table/post_training_comparison}

\paragraph{Comparison with supervised OCR tuning.}
\Cref{tab:post_training_comparison} compares post-training methods using the
same Qwen3-VL-8B-Thinking base and synthetic OCR data. Full-parameter SFT and
LoRA are already below the $34.9$ base score at their earliest audited
checkpoint and decline further, whereas QGO-8B reaches $40.8$. Thus, the shared
OCR data alone do not explain the gain; the post-training design is critical.
This comparison is diagnostic rather than fully compute-matched, because the
supervised baselines use their own tuning configurations. Together with the
accuracy-only reward ablation above, it shows that preserving the model's
reasoning behavior is essential when improving OCR.

\FloatBarrier

%% file: Table/overall_results.tex
\begin{table*}[!t]
\centering
\setlength{\tabcolsep}{3pt}
\renewcommand{\arraystretch}{1.1}
\resizebox{\textwidth}{!}{%
{\footnotesize
\begin{tabular}{@{}l *{4}{c!{\hspace{-4pt}}c} *{4}{c}@{}}
\toprule
\multicolumn{1}{c}{\multirow{4}{*}{\bfseries\footnotesize Model}} & 
\multicolumn{4}{c}{\bfseries MDUR} & 
\multicolumn{4}{c}{\bfseries MIQA} & 
\multicolumn{2}{c}{\bfseries MSOCR} & 
\multicolumn{2}{c}{\bfseries MGUI} \\
\cmidrule(lr){2-5} \cmidrule(lr){6-9} \cmidrule(lr){10-11} \cmidrule(lr){12-13}
 & 
\multicolumn{2}{c}{\bfseries \avg$\uparrow$} & 
\multicolumn{2}{c}{\bfseries \cv$\downarrow$} & 
\multicolumn{2}{c}{\bfseries \avg$\uparrow$} & 
\multicolumn{2}{c}{\bfseries \cv$\downarrow$} & 
\multicolumn{1}{c}{\bfseries \avg$\uparrow$} & 
\multicolumn{1}{c}{\bfseries \cv$\downarrow$} & 
\multicolumn{1}{c}{\bfseries \avg$\uparrow$} & 
\multicolumn{1}{c}{\bfseries \cv$\downarrow$} \\
\cmidrule(lr){2-3} \cmidrule(lr){4-5} \cmidrule(lr){6-7} \cmidrule(lr){8-9} \cmidrule(lr){10-10} \cmidrule(lr){11-11} \cmidrule(lr){12-12} \cmidrule(lr){13-13}
 & \texttt{trad.} & \texttt{vision} & \texttt{trad.} & \texttt{vision} & 
\texttt{trad.} & \texttt{vision} & \texttt{trad.} & \texttt{vision} & 
\texttt{vision} & \texttt{vision} & \texttt{vision} & \texttt{vision} \\
\midrule
\scriptsize Gemini-3.1-Pro-Preview & \textbf{80.24} & \textbf{78.45} & 0.018 & \textbf{0.024} & 60.27 & 53.11 & 0.097 & 0.077 & \textbf{30.38} & \textbf{0.238} & \underline{87.10} & \underline{0.026} \\
\scriptsize GPT-5 & \underline{72.89} & 69.03 & \textbf{0.010} & 0.037 & 54.82 & 55.95 & 0.057 & \underline{0.046} & 16.31 & 0.594 & 10.56 & 0.264 \\
\scriptsize GPT-5-mini & 69.78 & \underline{69.13} & \underline{0.013} & \underline{0.028} & \textbf{62.98} & \textbf{61.55} & \textbf{0.032} & \textbf{0.044} & 13.58 & 0.676 & 14.83 & 0.185 \\
\scriptsize Llama-4-Maverick & 57.50 & 48.29 & 0.026 & 0.077 & 46.07 & 44.05 & 0.071 & 0.080 & 8.27 & 0.943 & 54.80 & 0.214 \\
\scriptsize GLM-4.5v & 55.03 & 46.96 & 0.054 & 0.272 & 50.66 & 43.37 & 0.060 & 0.296 & 7.70 & 1.398 & 74.40 & 0.186 \\
\scriptsize Doubao-Seed-1.6 & 59.59 & 54.40 & 0.054 & 0.176 & 55.06 & 47.04 & 0.065 & 0.233 & 8.58 & 1.100 & 81.45 & 0.109 \\
\scriptsize Qwen3VL-235B-Thinking & 65.04 & 61.53 & 0.024 & 0.062 & 61.56 & 54.35 & \underline{0.041} & 0.126 & 15.99 & 0.605 & \textbf{89.15} & \textbf{0.024} \\
\scriptsize Qwen3VL-235B-Instruct & 61.07 & 56.49 & 0.037 & 0.083 & \underline{61.96} & \underline{59.52} & 0.131 & 0.162 & \underline{18.34} & \underline{0.569} & 82.10 & 0.077 \\
\scriptsize Qwen3VL-32B-Thinking & 64.04 & 60.46 & 0.017 & 0.098 & 54.98 & 49.46 & 0.062 & 0.160 & 11.37 & 0.794 & 86.35 & 0.048 \\
\midrule
\scriptsize Qwen3VL-8B-Thinking & 38.55 & 34.88 & 0.061 & 0.218 & 53.63 & 47.69 & 0.072 & 0.216 & 1.53 & 1.145 & 78.30 & 0.130 \\
\scriptsize \textbf{QGO-8B (ours)} & 46.82 & 40.84 & 0.080 & 0.190 & 55.24 & 51.06 & 0.060 & 0.174 & 8.17 & 0.759 & 80.00 & 0.095 \\
\bottomrule
\end{tabular}
}}
\caption{Overall model performance comparison on MDUR, MIQA, MSOCR and MGUI, where \avg is the average score across 10 languages, and \cv refers to the coefficient of variance among 10 languages. Best values in \textbf{bold}, second best \underline{underlined}. The Qwen3-VL-8B-Thinking model after OCR-centric reinforcement learning is denoted as QGO-8B. Qwen3-VL-235B-A22B-Thinking/Instruct is referred as Qwen3VL-235B-Thinking/Instruct for brevity.}
\label{tab:overall_res}
\end{table*}

%% file: Table/QGO_detail.tex
\begin{table*}[!t]
\centering
\setlength{\tabcolsep}{3pt}
\renewcommand{\arraystretch}{1.1}
\resizebox{\textwidth}{!}{%
{\footnotesize
\begin{tabular}{@{}l *{4}{cc} *{4}{c}@{}}
\toprule
\multicolumn{1}{c}{\multirow{4}{*}{\bfseries\footnotesize Model}} &
\multicolumn{4}{c}{\bfseries MDUR} &
\multicolumn{4}{c}{\bfseries MIQA} &
\multicolumn{2}{c}{\bfseries MSOCR} &
\multicolumn{2}{c}{\bfseries MGUI} \\
\cmidrule(lr){2-5} \cmidrule(lr){6-9} \cmidrule(lr){10-11} \cmidrule(lr){12-13}
&
\multicolumn{2}{c}{\bfseries \avg $\uparrow$} &
\multicolumn{2}{c}{\bfseries \cv $\downarrow$} &
\multicolumn{2}{c}{\bfseries \avg $\uparrow$} &
\multicolumn{2}{c}{\bfseries \cv $\downarrow$} &
\multicolumn{1}{c}{\bfseries \avg $\uparrow$} &
\multicolumn{1}{c}{\bfseries \cv $\downarrow$} &
\multicolumn{1}{c}{\bfseries \avg $\uparrow$} &
\multicolumn{1}{c}{\bfseries \cv $\downarrow$} \\
\cmidrule(lr){2-3} \cmidrule(lr){4-5} \cmidrule(lr){6-7} \cmidrule(lr){8-9} \cmidrule(lr){10-10} \cmidrule(lr){11-11} \cmidrule(lr){12-12} \cmidrule(lr){13-13}
& \texttt{trad.} & \texttt{vision} & \texttt{trad.} & \texttt{vision} &
\texttt{trad.} & \texttt{vision} & \texttt{trad.} & \texttt{vision} &
\texttt{vision} & \texttt{vision} & \texttt{vision} & \texttt{vision} \\
\midrule
\scriptsize Original Qwen3 & 38.55 & 34.88 & 0.061 & 0.218 & 53.63 & 47.69 & 0.072 & 0.216 & 1.53 & 1.145 & 78.30 & 0.130 \\
\midrule
\scriptsize QGO-8B (only-fmt) & 42.05 & 36.81 & 0.073 & 0.219 & \textbf{55.25} & 49.48 & 0.062 & 0.207 & 2.35 & 1.244 & 78.90 & 0.121 \\
\scriptsize QGO-8B (only-ocr) & 38.89 & 26.67 & \textbf{0.051} & 0.273 & 53.90 & 47.40 & 0.066 & 0.217 & 8.13 & \textbf{0.738} & 68.50 & 0.225 \\
\scriptsize \textbf{QGO-8B (full)} & \textbf{46.82} & \textbf{40.84} & 0.080 & \textbf{0.190} & 55.24 & \textbf{51.06} & \textbf{0.060} & \textbf{0.174} & \textbf{8.17} & 0.759 & \textbf{80.00} & \textbf{0.095} \\
\bottomrule
\end{tabular}
}}
\caption{Ablation study of the proposed reward functions on the MDUR, MIQA, MSOCR and MGUI tasks across 10 languages. ``only-fmt'' and ``only-ocr'' denote models trained exclusively with the Format and Length Constraint ($R_{\text{fmt}}$) or the OCR Accuracy Reward ($R_{\text{acc}}$), respectively.}
\label{tab:QGO_detail}
\end{table*}

%% file: Table/vision_w_ocr_MDUR.tex
\begin{table}[t]
\centering
\scriptsize
\renewcommand{\arraystretch}{1.1}
\setlength{\tabcolsep}{3pt}
\begin{tabular*}{\columnwidth}{@{\extracolsep{\fill}} l c c c c @{}}
\toprule
\textbf{Model} & \textbf{w\_o\_text} & \textbf{w\_text} & \textbf{$\Delta$ Avg} & \textbf{$\Delta$ CV} \\
\midrule
\textbf{GPT-5} & 69.03 & 71.10 & +2.07 & -2.46\% \\
\textbf{GPT-5-mini} & 69.13 & 70.71 & +1.58 & -1.39\% \\
\textbf{Gemini-3.1-Pro-Preview} & 78.45 & 80.45 & +1.99 & -0.48\% \\
\textbf{Qwen3-VL-235B-Thinking} & 61.53 & 66.97 & +5.44 & -4.42\% \\
\textbf{Qwen3-VL-8B-Thinking} & 34.88 & 41.75 & +6.87 & -17.64\% \\
\bottomrule
\end{tabular*}
\caption{MDUR vision results with and without a text reference. \textbf{w\_o\_text}: original vision input; \textbf{w\_text}: reference text provided. $\Delta$ Avg is the score gain, and $\Delta$ CV is the consistency change (lower is better).}
\label{tab:vis_w_ocr}
\end{table}

%% file: Table/post_training_comparison.tex
\begin{table}[!tbp]
\centering
\footnotesize
\setlength{\tabcolsep}{3.5pt}
\renewcommand{\arraystretch}{1.05}
\begin{tabular*}{\columnwidth}{@{\extracolsep{\fill}}lccc@{}}
\toprule
\textbf{Method} & \textbf{Early} & \textbf{Mid} & \textbf{Late} \\
\midrule
Base 8B & \multicolumn{3}{c}{34.9} \\
Full SFT & 26.1 (s10) & 23.4 (s30) & 22.4 (s50) \\
LoRA SFT & 26.7 (s10) & 25.9 (s30) & 24.8 (s70) \\
QGO-8B & \multicolumn{3}{c}{\textbf{40.8}} \\
\bottomrule
\end{tabular*}
\caption{MDUR \texttt{vision} average over 10 languages. All post-training
runs use the same base model and synthetic OCR data; parentheses give training
steps. Optimization configurations otherwise differ across methods.}
\label{tab:post_training_comparison}
\end{table}

%% file: Chapter/7_Conclusion.tex
\section{Conclusion}
We introduce PM\textsuperscript{4}Bench, a strictly parallel 10-language, four-task benchmark with controlled content and visual layout. Matched \texttt{traditional} and \texttt{vision} inputs isolate model behavior from dataset inconsistencies and expose text-in-image reading challenges. Across 10 LVLMs, vision input widens performance and cross-lingual gaps; ground-truth-text injection, per-language correlations, and a matched OCR intervention identify recognition as a key bottleneck.

Our label-free OCR-centric GRPO combines accuracy and format rewards to improve reading while preserving downstream reasoning. Trained without PM\textsuperscript{4}Bench evaluation examples, QGO-8B improves all four tasks across languages and transfers to external multilingual reading tasks.

%% file: Chapter/8_Limitations.tex
\section{Limitations}

\textbf{Scope of RL gains.} Our OCR-centric GRPO yields meaningful but bounded improvements: it lifts all four \texttt{vision}-setting tasks on PM\textsuperscript{4}Bench. It also transfers to CC-OCR, MTVQA, and OCRBench, although gains on the latter two are modest (matching the OCRBench baseline and gaining $+0.5$ on MTVQA trained-language \avg). When transferred directly to Qwen3-VL-32B-Thinking, the same recipe still improves OCR recognition, but downstream VQA performance declines modestly, suggesting that larger models may require scale-specific optimization. Genuine cross-lingual capability ultimately requires base-model improvements; OCR-centric RL is a high-leverage but partial intervention.

\textbf{Synthetic-data scope.} MSOCR and the MGUI task are intentionally small, fully synthetic decoupling probes designed to isolate cross-lingual capability differences from data artifacts (cultural variance, translation drift, content-coverage imbalance). They are not a replacement for specialized OCR or GUI benchmarks (\eg, dedicated text-spotting/KIE suites, interactive multi-step environments such as macOSWorld) and should not be read as such. PM\textsuperscript{4}Bench complements rather than supersedes those resources.

\textbf{Judge reliability.} MIQA evaluation uses an LLM-as-judge protocol with a single judge model. We provide both a translate-to-English consistency check (\cref{sec: appendix_e_translate_consistency}) and a Claude-Opus-4.7 cross-judge cross-validation (\cref{sec: appendix_e_alt_judge}); residual per-dimension and inter-judge variability is left for future work.

\textbf{Synthetic-to-real gap.} We acknowledge that MSOCR and MGUI inevitably differ from real-world OCR and GUI scenarios, and cannot fully simulate the complexity of real deployments. These synthetic tasks are designed primarily to fairly measure cross-lingual disparity under controlled conditions, rather than to precisely estimate absolute real-world performance. Our goal is therefore to make the performance patterns on synthetic data approximate real-world distributions as closely as possible, while leaving more faithful real-world evaluation to other benchmarks.

\textbf{Potential risks.} PM\textsuperscript{4}Bench is intended as a research artifact for diagnosing and mitigating cross-lingual capability disparities of LVLMs, but we acknowledge the existence of potential risks. \textbf{Data misuse.} Although our parallel corpus is sourced from publicly available datasets and rendered with synthetic templates, redistribution or downstream usage outside the intended evaluation context could amplify any residual biases in the source data. We therefore encourage users to audit cultural and demographic balance before any deployment-oriented use. \textbf{Evaluation misinterpretation.} Scores on PM\textsuperscript{4}Bench reflect controlled, parallel-corpus settings and should not be over-generalized as an absolute measure of real-world multilingual deployment readiness; we encourage users to pair our diagnostics with deployment-domain evaluation before drawing operational conclusions.

%% file: Chapter/Appendix_A.tex
\section{Vision-Setting Input Samples}
\label{sec: appendix_a}

Unlike traditional interleaved text-image input paradigms, the \textbf{Vision Setting} in PM\textsuperscript{4}Bench consolidates all textual information and reference images into a single, unified image as the input for Large Vision-Language Models (LVLMs). This design specifically emulates the authentic operational environments likely to be encountered by AI Agents and Embodied AI Robots.

During autonomous GUI navigation, these agents may receive smartphone or computer interfaces as pure pixel-level visual inputs (images or videos). Consequently, the agent is forced to rely exclusively on visual perception to parse interface semantics, particularly the embedded on-screen text. \textbf{This visual parsing becomes exceptionally challenging when the target interface operates in non-English or multilingual contexts.} The dedicated MGUI task in PM\textsuperscript{4}Bench (\cref{tab:gui_detail}) directly probes this regime through coordinate-based grounding on a parallel multilingual UI corpus.

As the pioneering benchmark evaluating multilingual capabilities under vision-only inputs, and, to our knowledge, the first to evaluate \emph{coordinate-based MGUI grounding on a strictly parallel multilingual corpus}, PM\textsuperscript{4}Bench bridges a critical gap in the field. Our data, findings and OCR-centric reinforcement learning method provide essential references for the future development and optimization of AI Agents/Embodied AI Robots, particularly those targeting global, multilingual user bases.

Input examples under both the \texttt{traditional} and \texttt{vision} settings, for each task and each language, are illustrated in~\cref{fig:Appendix_MDUR_vision_sample}--\cref{fig:Appendix_GUI_vision_sample}.

\subsection{MDUR Input Samples}
\label{sec: appendix_a_mdur}

\begin{figure*}[!tbp]
    \centering
    \includegraphics[width=\textwidth, height=0.78\textheight, keepaspectratio]{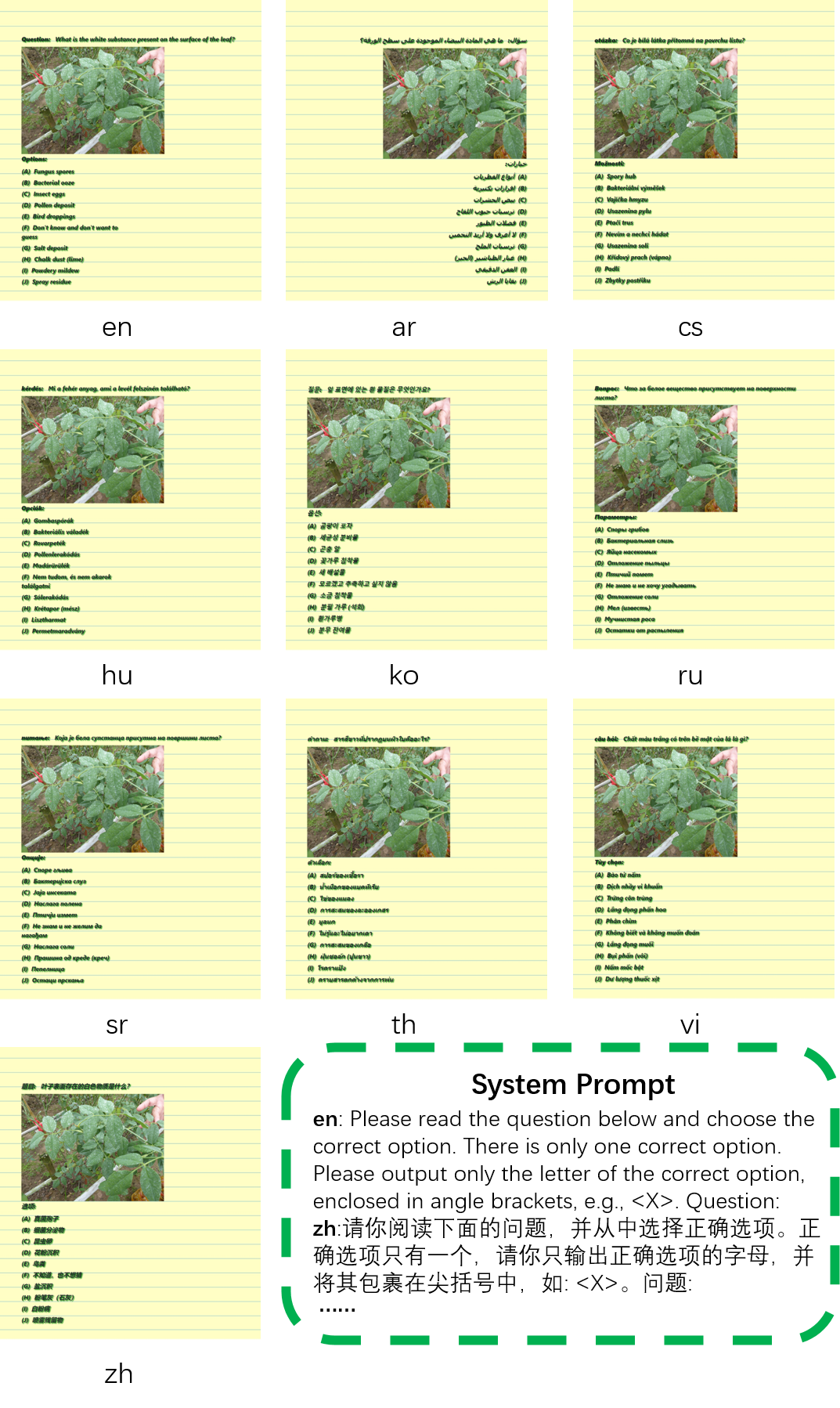}
    \caption{Examples of MDUR's \texttt{vision} setting input.}
    \label{fig:Appendix_MDUR_vision_sample}
\end{figure*}

\begin{figure*}[!tbp]
    \centering
    \includegraphics[width=\textwidth, height=0.78\textheight, keepaspectratio]{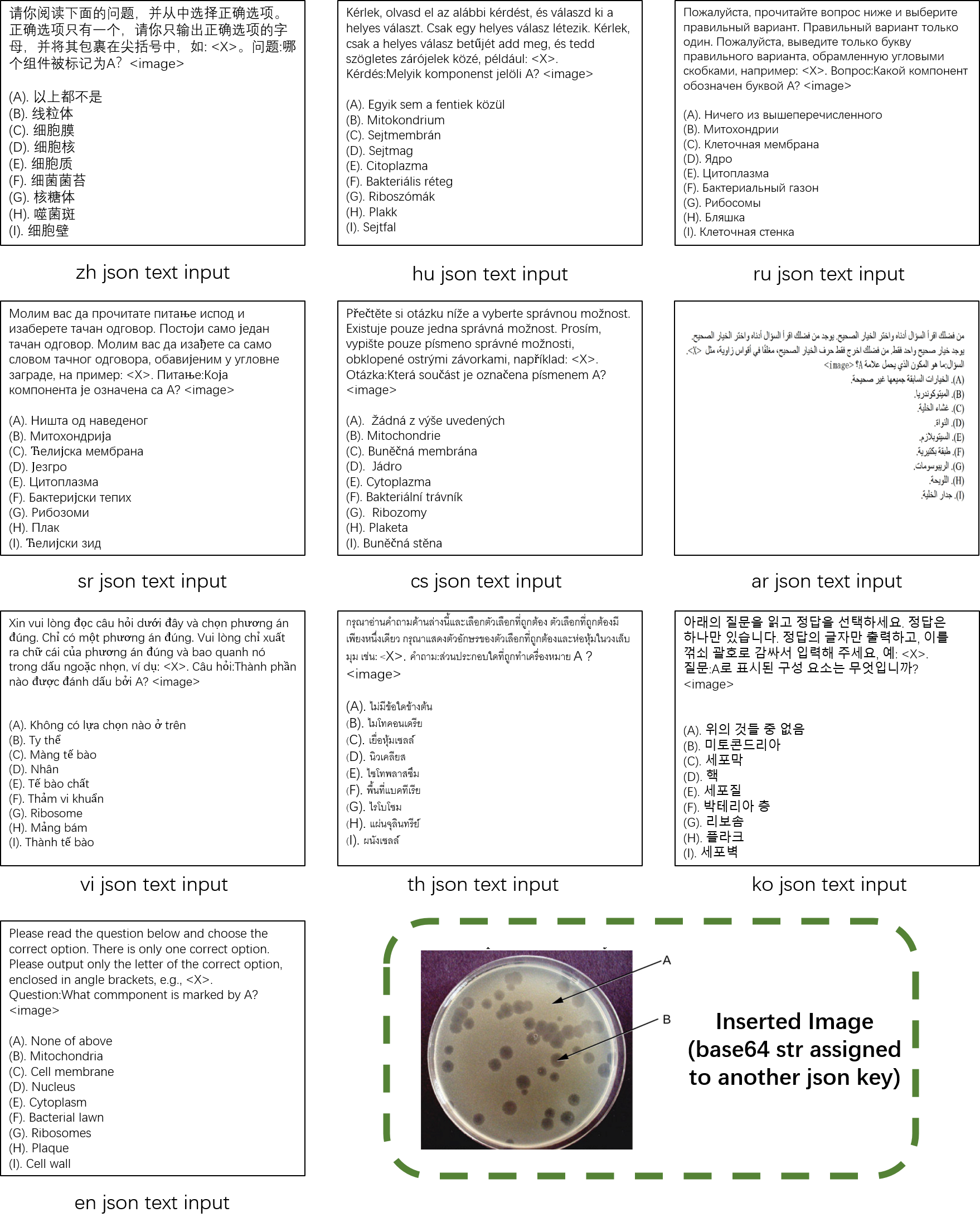}
    \caption{Examples of MDUR's \texttt{traditional} setting input.}
    \label{fig:Appendix_MDUR_standard_sample}
\end{figure*}

The \texttt{vision} setting (\cref{fig:Appendix_MDUR_vision_sample}) renders the question, options, and reference images into one page-level image; the \texttt{traditional} setting (\cref{fig:Appendix_MDUR_standard_sample}) supplies them as text plus separate image files. All 10 languages share the same underlying problem semantics and reference image; only the textual portion is localized.

\subsection{MIQA Input Samples}
\label{sec: appendix_a_miqa}

\begin{figure*}[!tbp]
    \centering
    \includegraphics[width=\textwidth, height=0.78\textheight, keepaspectratio]{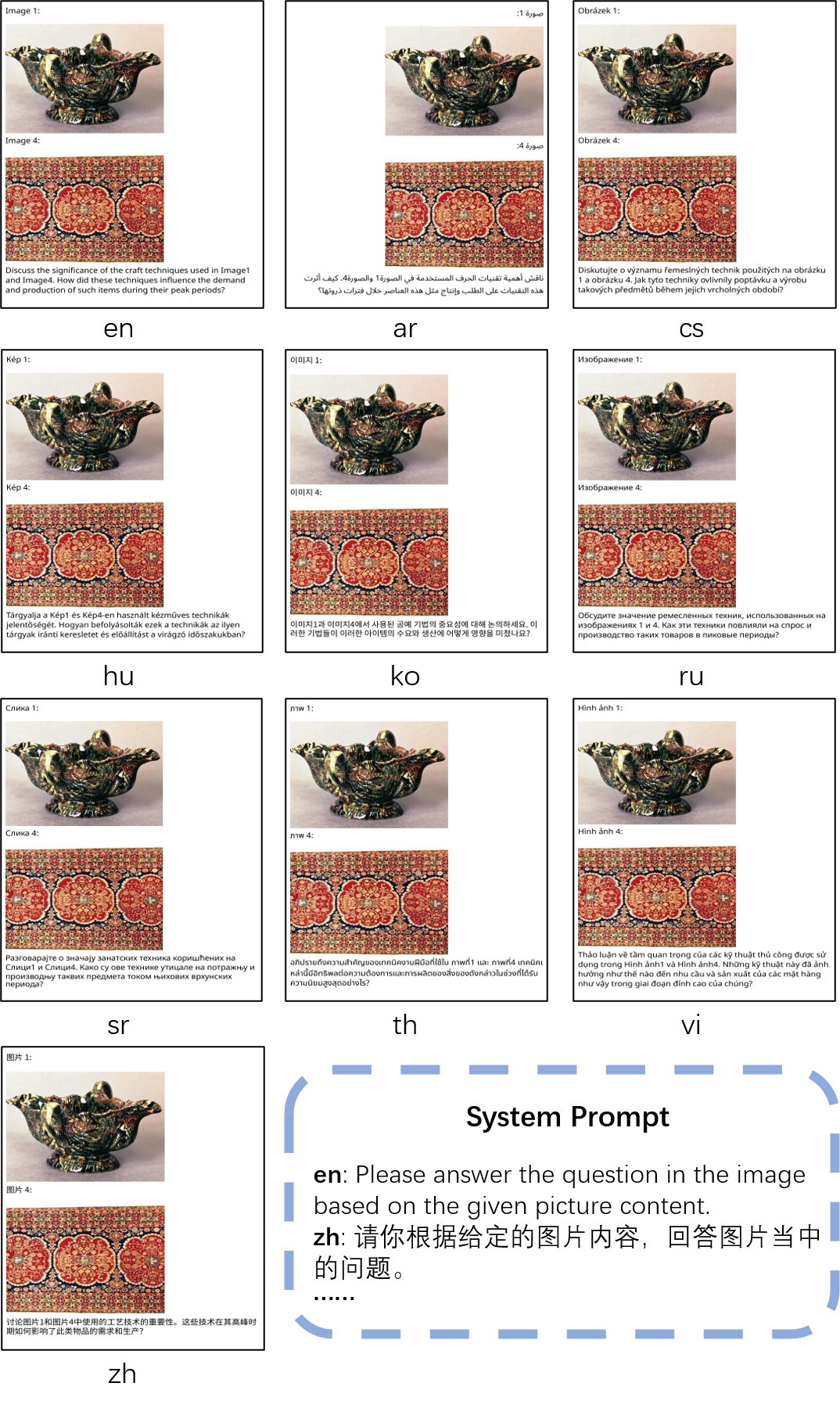}
    \caption{Examples of MIQA's \texttt{vision} setting input.}
    \label{fig:Appendix_MIQA_vision_sample}
\end{figure*}

\begin{figure*}[!tbp]
    \centering
    \includegraphics[width=\textwidth, height=0.78\textheight, keepaspectratio]{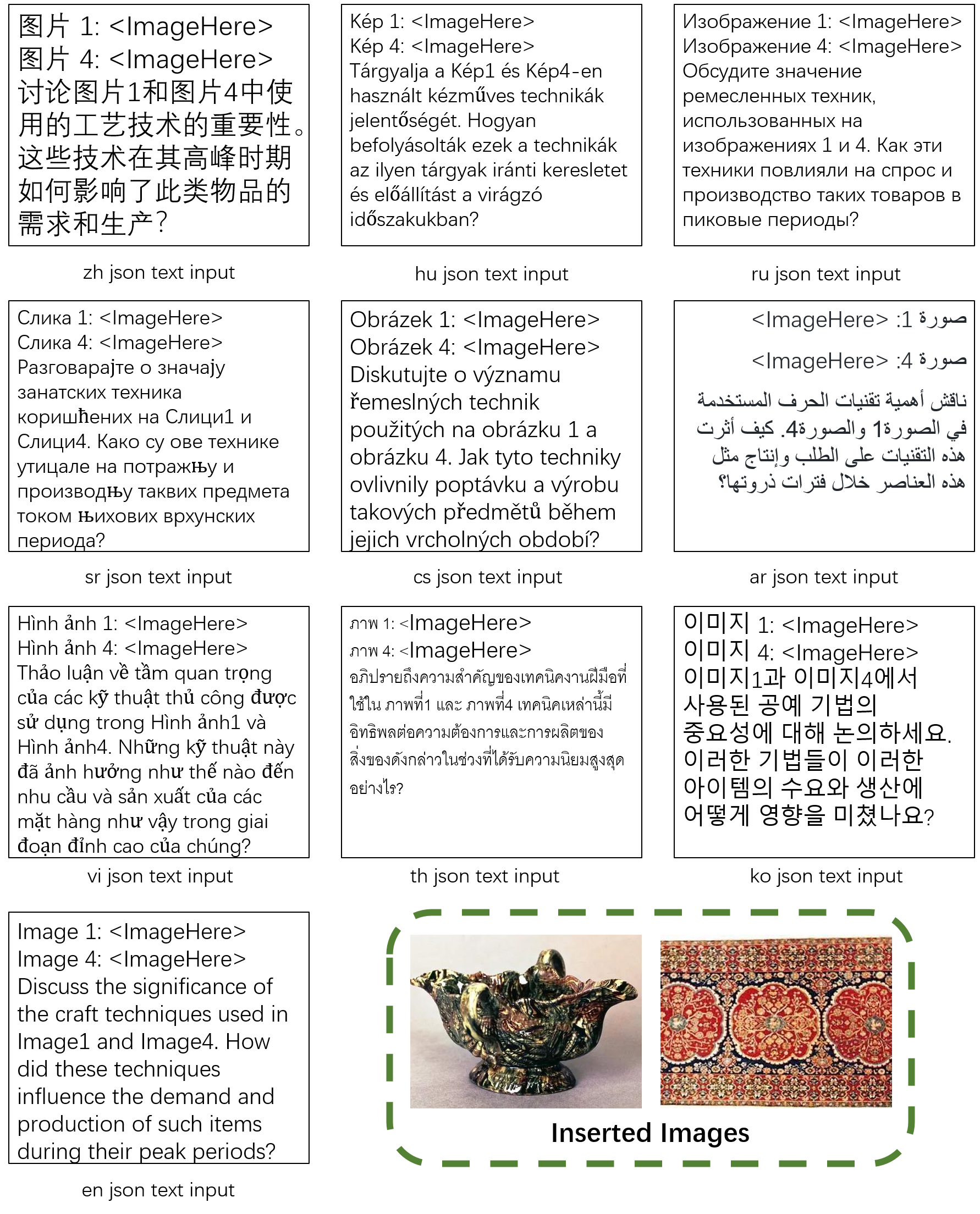}
    \caption{Examples of MIQA's \texttt{traditional} setting input.}
    \label{fig:Appendix_MIQA_standard_sample}
\end{figure*}

Each MIQA item asks a free-form question with multiple reference images. The \texttt{vision} setting (\cref{fig:Appendix_MIQA_vision_sample}) consolidates the question text and all reference images into a single composite image; the \texttt{traditional} setting (\cref{fig:Appendix_MIQA_standard_sample}) supplies the question as text and the images as separate inputs. The 10 language variants share the same reference images and only differ in question wording.

\subsection{MSOCR Input Samples}
\label{sec: appendix_a_msocr}

\begin{figure*}[!tbp]
    \centering
    \includegraphics[width=\textwidth, height=0.78\textheight, keepaspectratio]{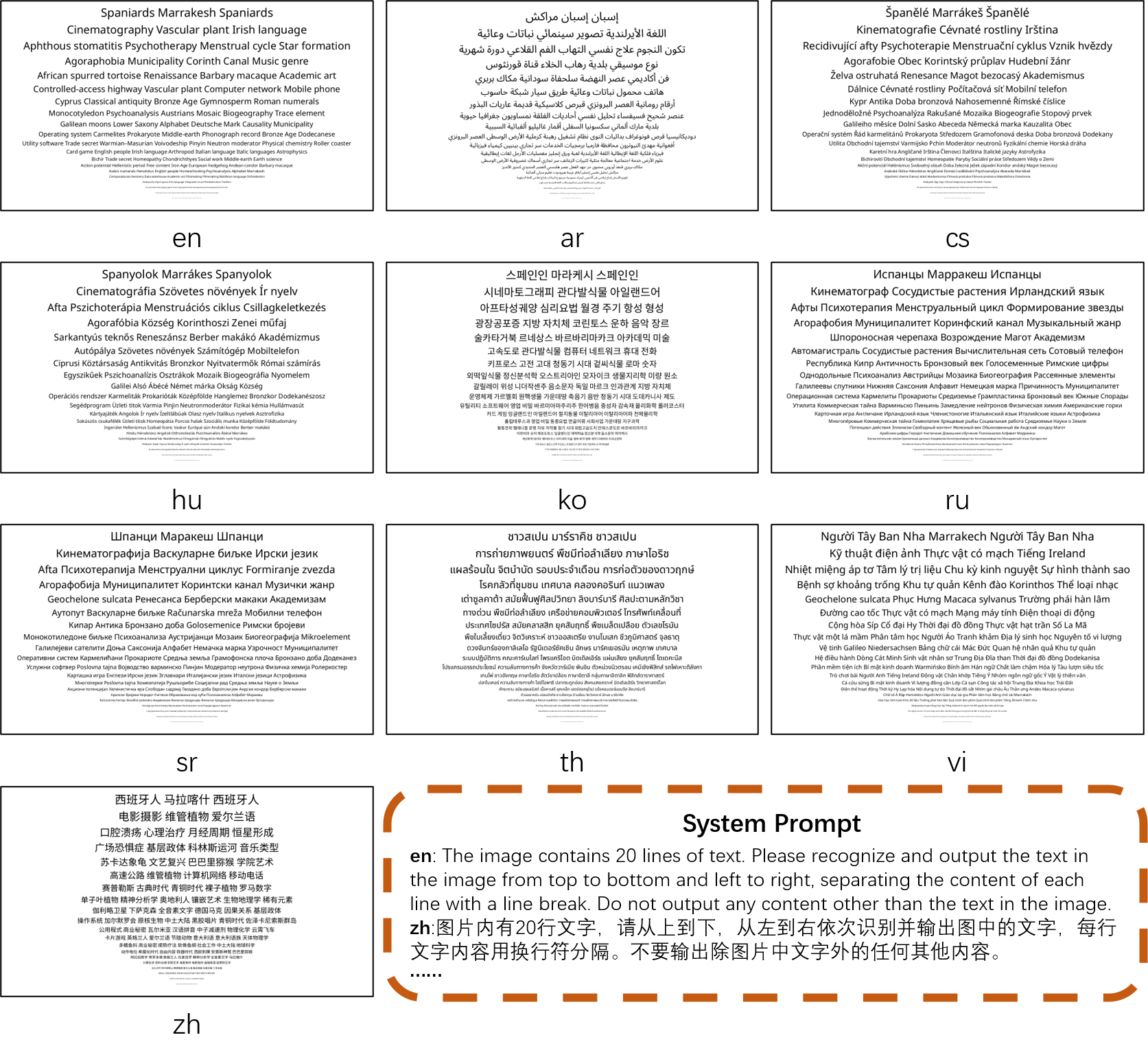}
    \caption{Examples of MSOCR input.}
    \label{fig:Appendix_MSOCR_vision_sample}
\end{figure*}

Each MSOCR sample is a single rendered image containing 20 lines of text in a single target language. The model is asked to transcribe the full content top-to-bottom; we score by exact-match accuracy at the line level. \cref{fig:Appendix_MSOCR_vision_sample} shows representative examples spanning the 10 supported languages.

\subsection{MGUI Input Samples}
\label{sec: appendix_a_gui}

\begin{figure*}[!tbp]
    \centering
    \includegraphics[width=\textwidth, height=0.78\textheight, keepaspectratio]{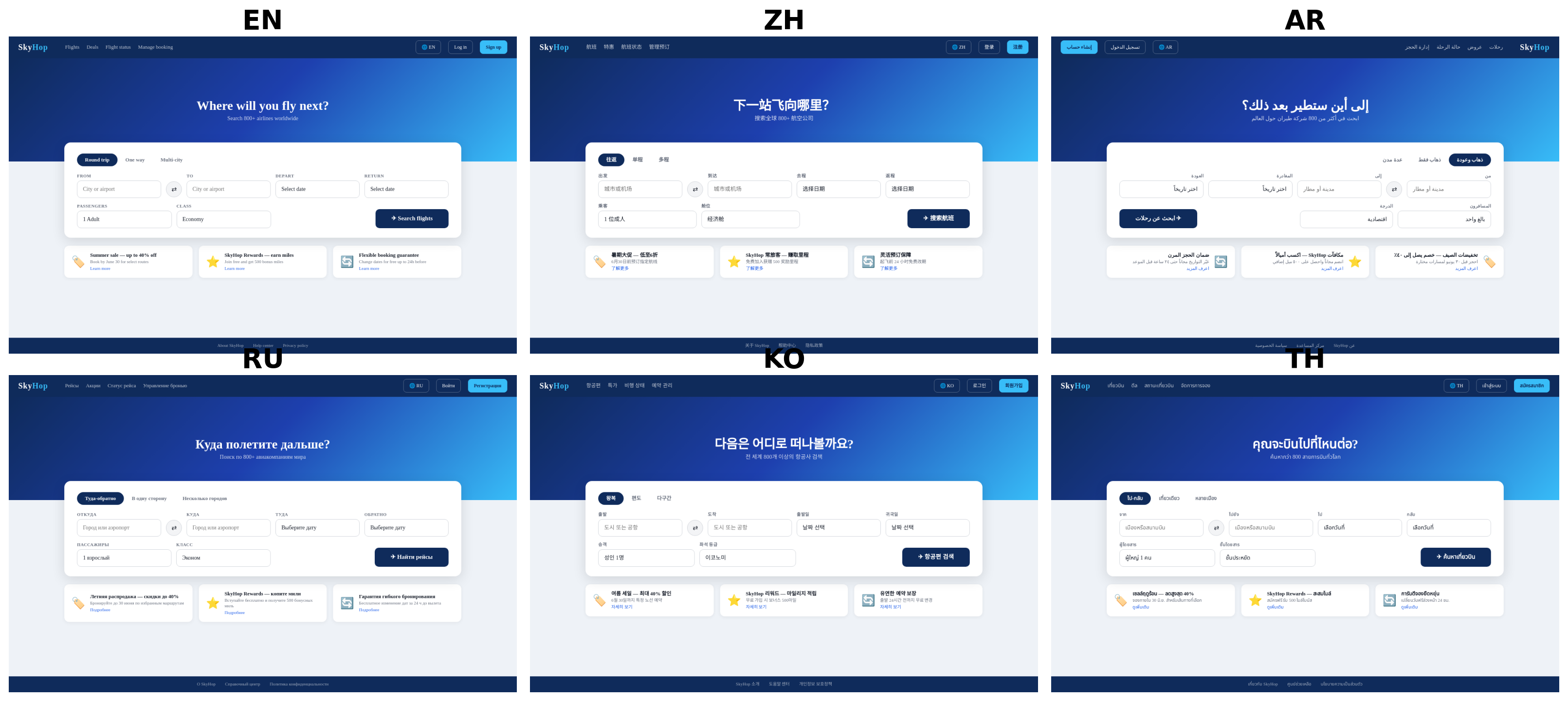}
    \caption{Examples of the MGUI task. The same flight-search webpage template (\texttt{a01\_search}) is rendered identically in 6 of the 10 supported languages (EN, ZH, AR, RU, KO, TH); only the visible text changes.}
    \label{fig:Appendix_GUI_vision_sample}
\end{figure*}

The MGUI task is constructed by programmatically rendering a fixed set of HTML/Jinja webpage templates with the same layout in all 10 languages. As shown in~\cref{fig:Appendix_GUI_vision_sample}, the visual structure (positions and sizes of buttons, fields, banners, etc.) is byte-identical across languages: only the visible text is swapped, so any cross-lingual difference in click accuracy is attributable to the model's multilingual visual understanding rather than dataset bias. The model is asked to emit a single $(x, y)$ coordinate per natural-language instruction; the prediction is judged correct if the point falls inside the ground-truth bounding box of the target interface element.

The 100 templates are organized into 10 product-surface \emph{topics} (10 sub-page templates per topic), spanning the most common GUI flows that a multilingual visual agent would encounter in real deployment. \cref{tab:gui_topics} enumerates the full topic-to-sub-page mapping.

\input{Table/gui_topics}

\FloatBarrier

\subsection{Synthetic OCR Training Data}
\label{sec: appendix_a_synth}

\begin{figure}[!htbp]
    \centering
    \includegraphics[width=\columnwidth, keepaspectratio]{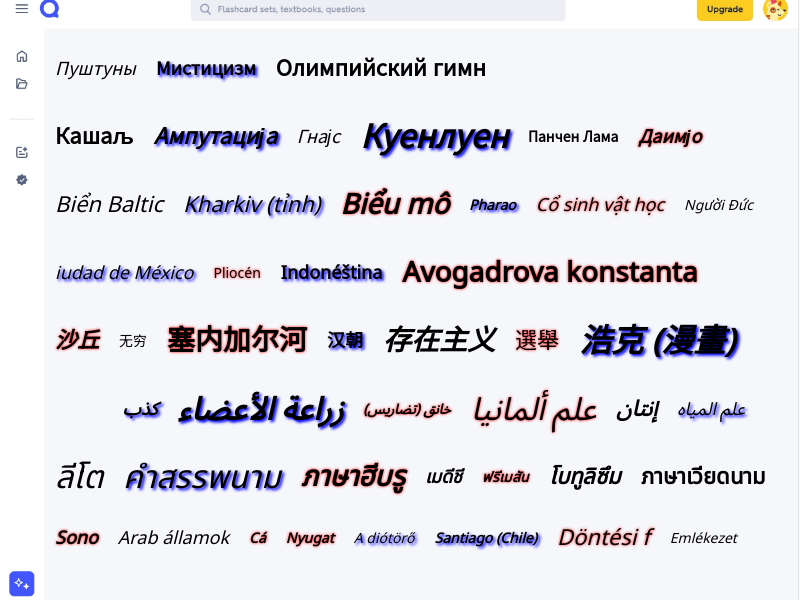}
    \caption{Example of synthesized OCR-centric GRPO training data.}
    \label{fig:train_sample}
\end{figure}

A representative example of the multilingual synthetic data used to train QGO-8B (\cref{sec: ocr_centric_rl}) is shown in~\cref{fig:train_sample}.

\FloatBarrier

%% file: Table/gui_topics.tex
\begin{table*}[!htbp]
\centering
\footnotesize
\renewcommand{\arraystretch}{1.15}
\setlength{\tabcolsep}{4pt}
\begin{tabular*}{\textwidth}{@{\extracolsep{\fill}} l p{0.78\textwidth}}
\toprule
\textbf{Topic (prefix)} & \textbf{10 sub-page templates} \\
\midrule
Flight booking (\texttt{a}) & Flight Search; Search Results; Flight Detail; Seat Selection; Booking Form; Payment Page; Boarding Pass; Flight Status; Manage Booking; Loyalty Program. \\
Online banking (\texttt{b}) & Dashboard; Money Transfer; Transaction History; Card Management; Bill Payments; Investment Portfolio; Loan Overview; Budget Planner; Security Center; Support \& Help. \\
Code editor / IDE (\texttt{c}) & Main Editor; Integrated Terminal; Git Panel; Extension Marketplace; Debug View; Global Search; Settings; Merge Conflict Resolver; Refactor Tools; Project Dashboard. \\
Admin dashboard (\texttt{d}) & Overview; Orders; Customers; Products; Inventory; Analytics; Marketing; Team; Settings; Reports. \\
E-commerce (\texttt{e}) & Product Detail; Shopping Cart; Checkout; Search Results; Order History; Wishlist; Product Comparison; Reviews; Return/Refund Flow; Account Settings. \\
Food delivery (\texttt{f}) & Home; Restaurant Menu; Cart; Checkout; Order Tracking; Search Results; Reviews; User Profile; Deals \& Coupons; App Settings. \\
Maps (\texttt{g}) & Explore; Route Planner; Place Detail Card; Public Transit; Saved Places; Traffic \& Incidents; Street View; Nearby Businesses; Offline Maps; Map Settings. \\
Email client (\texttt{m}) & Inbox; Compose; Read View; Folder Management; Advanced Search; Settings; Contacts; Calendar Invite; Label Management; Spam Folder. \\
Social media (\texttt{s}) & Feed; Profile; Notifications; Messages; Compose; Trending; Groups; Story Viewer; Privacy Settings; Analytics. \\
Weather (\texttt{w}) & Current Dashboard; 7-Day Forecast; Radar Map; Alerts; Saved Locations; Astronomy; Air Quality; Precipitation Detail; App Settings; Historical Comparison. \\
\bottomrule
\end{tabular*}
\caption{The 10 product-surface \emph{topics} that anchor PM\textsuperscript{4}Bench's MGUI task, with 10 HTML/Jinja sub-page templates per topic ($10 \times 10 = 100$ templates total). Each template is rendered with byte-identical layout in all 10 languages and paired with two indirect instructions, yielding $100 \times 10 \times 2 = 2000$ language-balanced (200 per language) coordinate-grounding test items.}
\label{tab:gui_topics}
\end{table*}

%% file: Chapter/Appendix_B.tex
\section{Prompts}
\label{sec: appendix_b}
\label{sec: appendix_prompts}

This section collects the prompts we used to query the evaluated LVLMs (one figure per task) and the prompts used by the LLM judge for MIQA. The translation prompt used to localise the corpus into the 9 non-English languages is given in~\cref{table:Translation_Prompt}.

\subsection{Task Prompts}
\label{sec: appendix_b_task_prompts}

\begin{figure}[!htbp]
    \centering
    \includegraphics[width=\columnwidth, keepaspectratio]{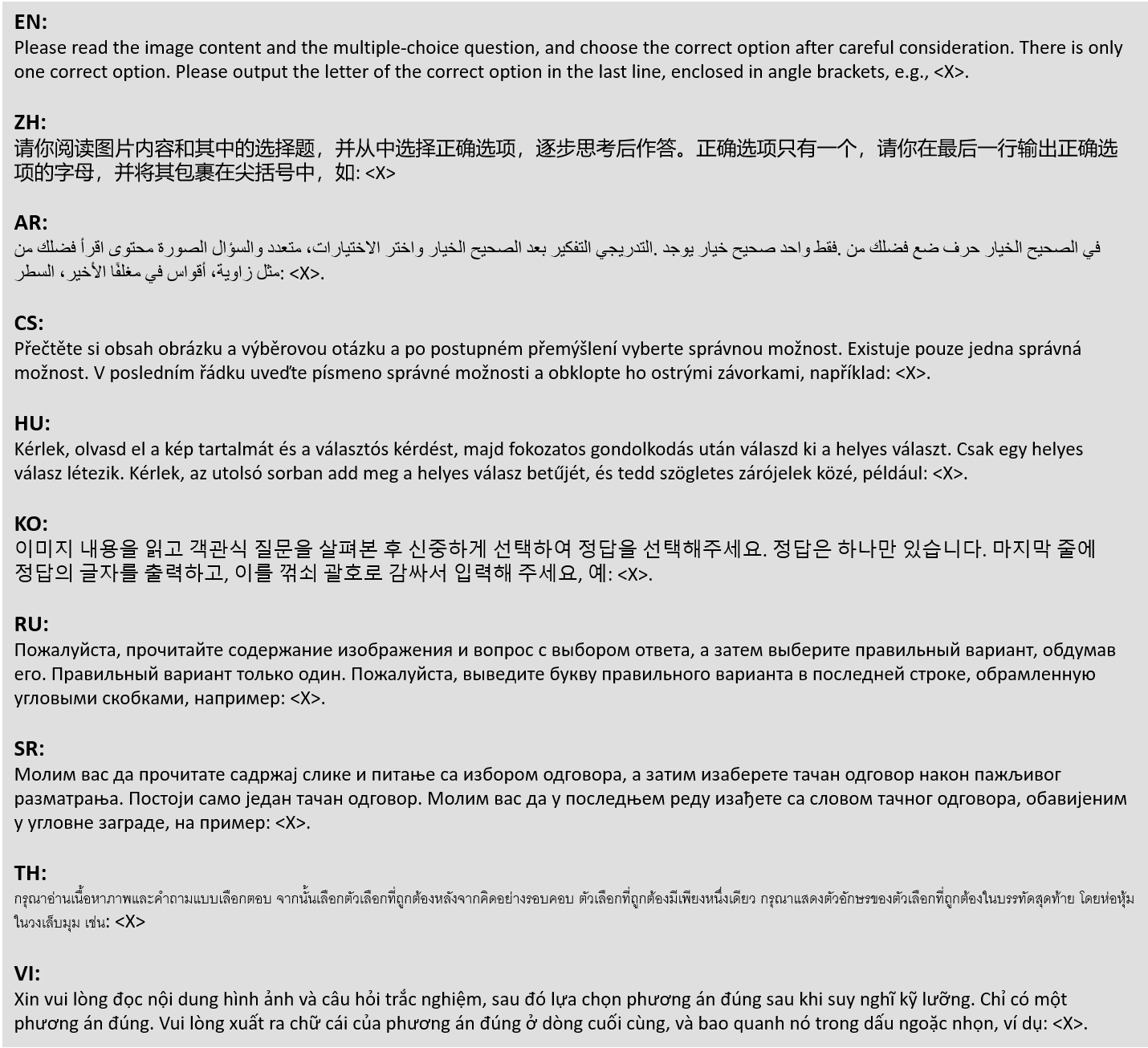}
    \caption{Prompts used for MDUR.}
    \label{fig:MDUR_prompt}
\end{figure}

\begin{figure}[!htbp]
    \centering
    \includegraphics[width=\columnwidth, keepaspectratio]{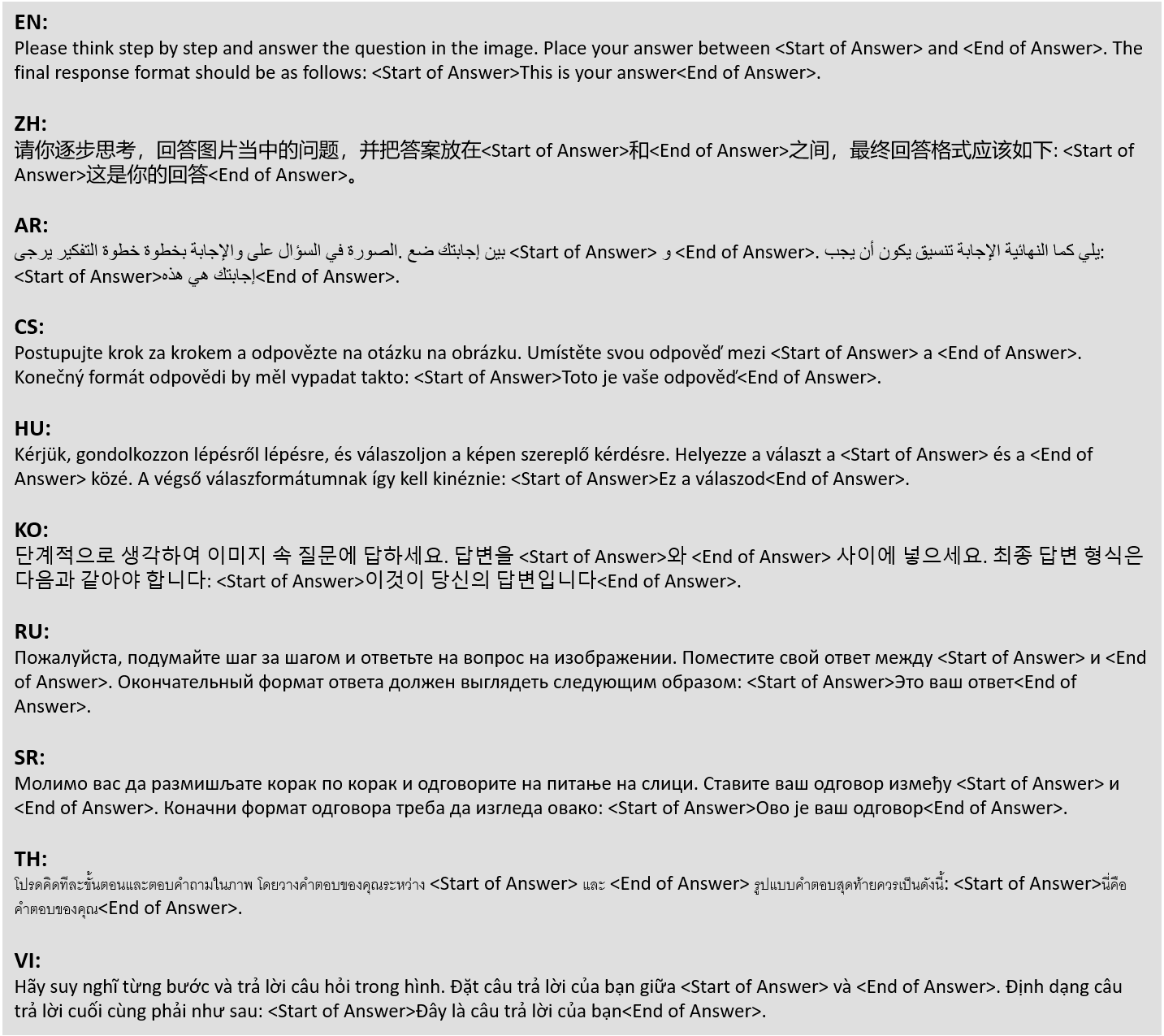}
    \caption{Prompts used for MIQA.}
    \label{fig:MIQA_prompt}
\end{figure}

\begin{figure}[!htbp]
    \centering
    \includegraphics[width=\columnwidth, keepaspectratio]{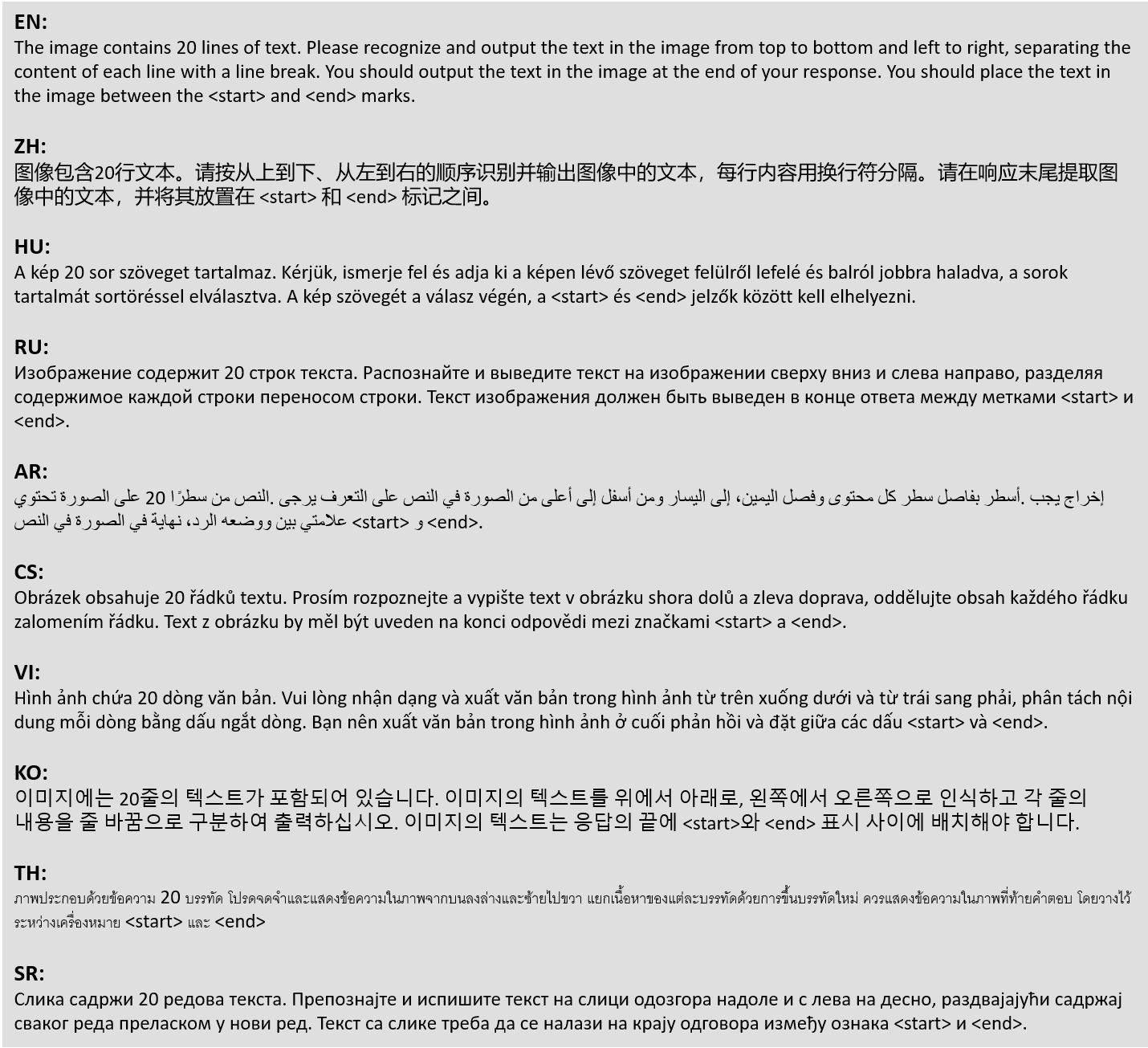}
    \caption{Prompts used for MSOCR.}
    \label{fig:MSOCR_prompt}
\end{figure}

\FloatBarrier

\subsection{LLM-as-Judge Prompt for MIQA}
\label{sec: appendix_b_judge_prompt}

\begin{figure}[!htbp]
    \centering
    \includegraphics[width=\columnwidth, keepaspectratio]{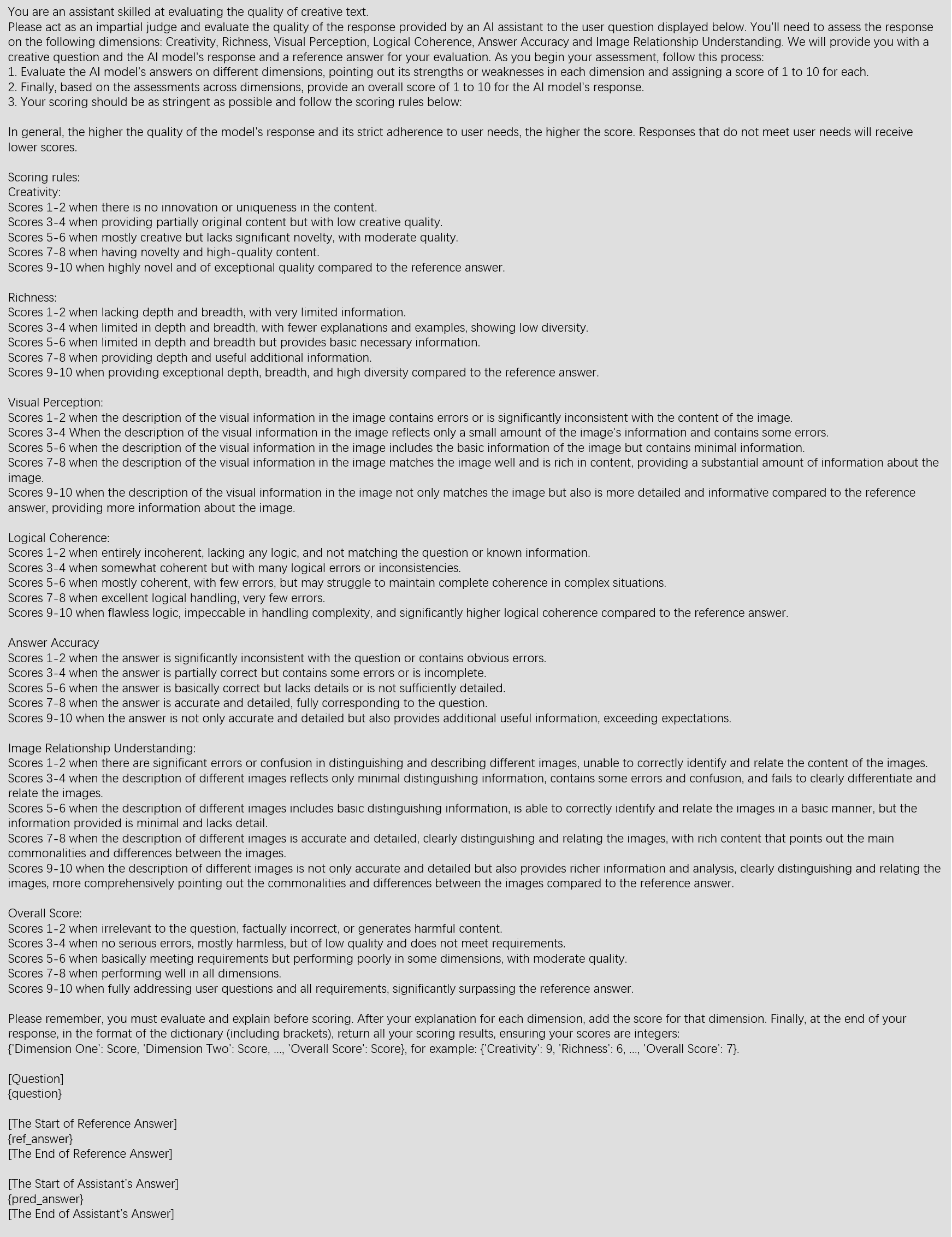}
    \caption{Prompt used by the LLM judge to score MIQA responses.}
    \label{fig:MIQA_eval_prompt}
\end{figure}

\FloatBarrier

\subsection{Translation Prompt}
\label{sec: appendix_b_translation_prompt}

\cref{table:Translation_Prompt} shows the two prompts used by our human-expert-in-the-loop translation pipeline (\S3.5). The \emph{Translation Prompt} asks an LLM to render the source content into the target language while preserving variables, numerals, mathematical notation, code, and image / option placeholders verbatim. For each target-language item, three independent translations are produced; the \emph{Claude Selection Prompt} then asks a second LLM to choose the best of the three by index. The selected candidate is finally cross-checked by a native-speaker annotator before entering the benchmark.

\input{Table/translation_prompt}

\FloatBarrier

%% file: Table/translation_prompt.tex
\begin{table}[!ht]
\centering
\begin{tcolorbox}[title={Translation Prompt}, colback=white, coltitle=black, colbacktitle=white!0]
You are a language expert specialized in \{\textit{lang}\}. Please translate the following content \{\textit{lang}\}. Please organize your output according to the format of the input strictly. Do not output anything other than translation. Do not translate any variables, Arabic numerals, mathematical symbols Python code or people's names. Only translate the textual description. Do not translate image marks or option indexes included by '$\langle  \rangle$', '()', for instance, '$\langle ImageHere \rangle$', '(b)'. Input: \{\textit{input}\}\\Your translation:
\end{tcolorbox}

\begin{tcolorbox}[title={Claude Selection Prompt}, colback=white, coltitle=black, colbacktitle=white!0]
You are a language expert specialized in \{\textit{lang}\}. Given 3 translations of an English text, you are to choose one translation which has the best quality. A good translation should accurately preserve the numbers, variables or Python code in the English text, and correctly translate other content. You can only output the index of translation, '1', '2' or '3', do not output anything else. \\
$\langle start of English text \rangle$ \\
\{\textit{Original English text}\}\\
$\langle End of English text \rangle$ \\
$\langle start of translation 1 \rangle$ \\
\{\textit{translation\_1}\}\\
$\langle end of translation 1 \rangle$ \\
$\langle start of translation 2 \rangle$ \\
\{\textit{translation\_2}\}\\
$\langle end of translation 2 \rangle$\\
$\langle start of translation 3 \rangle$\\
\{\textit{translation\_3}\}\\
$\langle end of translation 3 \rangle$\\
Your selection:
\end{tcolorbox}
\caption{Prompt templates for human-expert in loop translation. The Italic \{\textit{text}\} in curly braces represents variables that need to be replaced.}
\label{table:Translation_Prompt}
\end{table}

%% file: Chapter/Appendix_C.tex
\section{Language Selection}
\label{sec: appendix_c}

\input{Table/language_selection}

The selected languages cover multiple language families and script systems. We additionally report GraphCom's four source-defined dimensions as descriptive attributes of the corresponding writing systems. These measurements provide context for visual-form variation but are not treated as causal predictors of benchmark performance. Among the selected languages for which GraphCom reports a matching script, simplified Chinese has the largest mean value on each of the four dimensions, yet Chinese remains relatively strong in the \texttt{vision} setting, consistent with the importance of training-resource coverage beyond graphic form alone.

As shown in~\cref{tab:language_selection} and~\cref{tab:GraphCom_score}, we report the following metrics:

\input{Table/GCscore}

\begin{itemize}
    \item \textbf{Perimetric Complexity (PC)}:
    \begin{itemize}
        \item Formula: \( PC = P^2 / (4 \pi A) \).
        \item \( P \): total perimeter of the shape (in pixels).
        \item \( A \): pixel area of the foreground (the shape itself).
        \item Reflects the spatial density of the shape, independently of size.
    \end{itemize}

    \item \textbf{Disconnected Components (DC)}:
    \begin{itemize}
        \item Counts the independent, non-connected parts of the shape.
        \item For example, the letter ``i'' has two disconnected components (the dot and the vertical stroke), while ``T'' has one continuous part.
        \item Reflects the discontinuity of the shape, indicating how fragmented it appears visually.
    \end{itemize}

    \item \textbf{Number of Connected Points (CP)}:
    \begin{itemize}
        \item Counts the intersection points where multiple segments meet.
        \item For instance, the letter ``T'' has one connected point, while ``F'' has two.
        \item Reflects the cohesion of the shape, indicating how well its strokes are interconnected.
    \end{itemize}

    \item \textbf{Number of Simple Features (SF)}:
    \begin{itemize}
        \item Counts the basic elements that make up the shape, such as strokes, lines, dots or circles.
        \item For example, the letter ``L'' consists of two simple features (a vertical and a horizontal line).
        \item Closely related to stroke count; especially useful for evaluating complex writing systems such as Chinese characters or Japanese kana.
    \end{itemize}

    \item \textbf{Graph Inventory (GI)}:
    \begin{itemize}
        \item GI is the number of characters in the script's character set.
    \end{itemize}

\end{itemize}

\FloatBarrier

%% file: Table/language_selection.tex
\begin{table*}[!htbp]
\centering
\begin{tabular*}{\textwidth}{l @{\extracolsep{\fill}} c l l}
\toprule
\textbf{Language} & \textbf{ISO} & \textbf{Language Family} & \textbf{Script System} \\
\midrule
\textbf{Chinese} & zh & Sino-Tibetan & Simplified Chinese characters \\
\textbf{Thai} & th & Kra--Dai & Thai \\
\textbf{Korean} & ko & Koreanic & Hangul / Chos\v{o}n'g\u{u}l \\
\textbf{Arabic} & ar & Afro-Asiatic & Arabic alphabet \\
\textbf{Hungarian} & hu & Uralic & Latin \\
\textbf{Czech} & cs & Indo-European & Latin \\
\textbf{Russian} & ru & Indo-European & Cyrillic \\
\textbf{Serbian} & sr & Indo-European & Serbian Cyrillic \\
\textbf{Vietnamese} & vi & Austroasiatic & Latin \\
\textbf{English} & en & Indo-European & Latin \\
\bottomrule
\end{tabular*}
\caption{Language and writing-system information for the 10 languages in PM\textsuperscript{4}Bench. GraphCom measurements are reported separately in \cref{tab:GraphCom_score}.}
\label{tab:language_selection}
\end{table*}

%% file: Table/GCscore.tex
\begin{table*}[!htbp]
\centering
\begin{tabular*}{\textwidth}{l @{\extracolsep{\fill}} ccccc}
\hline
\multicolumn{1}{c}{\textbf{Language}} & \textbf{GI} & \textbf{PC Mean} & \textbf{DC Mean} & \textbf{CP Mean} & \textbf{SF Mean} \\ \hline
\textbf{Chinese (Simplified)}          & 6097 & 29.47 & 4.01 & 9.54 & 10.60 \\
\textbf{Thai}                          & 102  & 14.88 & 1.68 & 4.54 & 6.24  \\
\textbf{Korean}                        & 40   & 14.71 & 1.38 & 2.15 & 3.40  \\
\textbf{Arabic}                        & 28   & 8.78  & 1.82 & 1.36 & 3.07  \\
\textbf{Hungarian}                     & ---  & ---   & ---  & ---  & ---   \\
\textbf{Czech}                         & ---  & ---   & ---  & ---  & ---   \\
\textbf{Russian}                       & 33   & 7.51  & 1.12 & 2.05 & 2.89  \\
\textbf{Serbian}                       & 30   & 7.34  & 1.02 & 2.02 & 2.83  \\
\textbf{Vietnamese}                    & ---  & ---   & ---  & ---  & ---   \\
\textbf{English}                       & 26   & 6.85  & 1.04 & 1.44 & 2.25  \\ \hline
\end{tabular*}
\caption{Source-reported GraphCom~\cite{chang2018graphcom} measurements for matching scripts. Means average all characters in the graph inventory (GI). --- indicates that GraphCom does not report the modern script used by our benchmark; in particular, its ``Hungarian Runes'' row does not represent modern Hungarian written in Latin script.}
\label{tab:GraphCom_score}
\end{table*}

%% file: Chapter/Appendix_D.tex
\section{Detailed Per-Language Results}
\label{sec: appendix_d}

This section reports the per-language scores that lie behind the aggregated \avg and \cv values shown in~\cref{tab:overall_res}. Tables~\ref{tab:msocr_detail}, \ref{tab:mdur_detail}, \ref{tab:miqa_detail} and~\ref{tab:gui_detail} give the breakdown for MSOCR, MDUR, MIQA and MGUI, respectively.

\subsection{\texorpdfstring{Sanity-check task: MGUI\textsubscript{text}}{Sanity-check task: MGUI-text}}
\label{sec: appendix_d_gui_text}

Because GPT-5 and GPT-5-mini both score unusually low on the main MGUI grounding task (\cref{tab:gui_detail}), in contrast to their strong performance on every other benchmark, we ran a controlled companion task to verify that this is a genuine model property rather than an experimental artifact. We call this companion task MGUI\textsubscript{text}.

\textbf{Setup.}\; The MGUI\textsubscript{text} task uses the exact same screenshots and the exact same natural-language instructions as the main MGUI grounding task. The only difference is the answer format: instead of emitting a click coordinate $(x, y)$, the model is asked to output the visible \emph{text} of the target UI element (\eg{}, ``One way'', ``Search flights''). A prediction is judged correct if its normalized text matches the target element's visible text. This isolates the recognition + understanding capability (``which element should be clicked'') from the spatial grounding capability (``where exactly is that element'').

\textbf{Results.}\; \cref{tab:gui_text_detail} reports the per-language results. GPT-5 reaches $94.04$ averaged over 10 languages, second only to Gemini-3.1-Pro-Preview ($97.15$) and clearly above all open-source models in our suite. GPT-5-mini reaches $91.91$. Across all 10 languages the two GPT-5 variants stay above $85$ on every language, including the visually hard AR ($88.00$ / $87.94$) and TH ($93.50$ / $85.00$), demonstrating that their recognition pipeline copes with non-English on-screen text just fine.

\textbf{Implication.}\; Combined with the median click-to-bbox-edge distance of only $\sim 33$~px reported in~\cref{sec:causes_of_gap}, this confirms that GPT-5 cleanly identifies the correct UI element and emits well-formed coordinates aimed at roughly the right neighborhood; the failure is restricted to fine-grained pixel-level grounding. We surface this anomaly to forestall the natural reviewer concern that our pipeline mishandled GPT-5 outputs, and leave a deeper investigation of why a strong general VLM exhibits such a narrow precision-grounding gap to future work. The gap is also model-specific rather than vendor-specific: other commercial systems (e.g. Gemini-3.1-Pro-Preview) do not exhibit this asymmetry, suggesting it reflects a particular decoding-time bias of the GPT-5 family rather than a fundamental limitation of the screenshot-grounding paradigm.

\input{Table/MSOCR}
\input{Table/MDUR}
\input{Table/MIQA}
\input{Table/GUI}
\input{Table/GUI_text}

\FloatBarrier

%% file: Table/MSOCR.tex
\begin{table*}[!htbp]
\centering
\resizebox{1\textwidth}{!}{
\small
\begin{tabular}{lcccccccccc}
\hline
\textbf{Model}    & EN    & ZH    & HU    & RU    & SR    & CS    & AR    & VI    & TH    & KO    \\ \hline
\textbf{Gemini-3.1-Pro-Preview}             & 35.47 & 29.40 & 34.88 & 35.26 & 34.08 & 34.14 & 28.10 & 33.36 & 10.14 & 28.96 \\
\textbf{GPT-5}             & 33.02 & 11.08 & 24.28 & 24.30 & 12.90 & 25.96 & 3.26 & 15.06 & 2.12 & 11.08 \\
\textbf{GPT-5-mini}             & 33.30 & 10.70 & 16.30 & 23.74 & 7.92 & 16.96 & 1.74 & 13.32 & 1.90 & 9.90 \\
\textbf{Llama-4-Maverick}             & 30.38 & 7.08 & 5.66 & 12.34 & 3.58 & 6.48 & 3.62 & 6.46 & 2.74 & 4.40 \\
\textbf{GLM-4.5v}             & 31.88 & 24.68 & 5.10 & 7.82 & 0.12 & 6.24 & 0 & 0.92 & 0 & 0.26 \\
\textbf{Doubao-Seed-1.6}             & 29.42 & 23.50 & 9.78 & 3.66 & 0.70 & 6.30 & 2.78 & 6.28 & 0 & 3.34 \\
\textbf{Qwen3-VL-235B-A22B-Thinking}             & 32.29 & 30.39 & 12.69 & 21.96 & 11.55 & 18.36 & 3.98 & 13.86 & 0.52 & 14.29 \\
\textbf{Qwen3-VL-235B-A22B-Instruct}             & 32.96 & 31.44 & 16.12 & 23.04 & 16.84 & 23.38 & 0.04 & 21.60 & 1.02 & 16.92 \\
\textbf{Qwen3-VL-32B-Thinking}             & 28.76 & 26.09 & 8.81 & 14.95 & 7.20 & 9.43 & 0.72 & 8.12 & 0 & 9.64 \\
\midrule
\textbf{Qwen3-VL-8B-Thinking}             & 4.82 & 4.76 & 0.22 & 1.94 & 0.18 & 0.38 & 0.16 & 1.32 & 0 & 1.48 \\
\textbf{QGO-8B (ours)}             & 20.86 & 16.58 & 5.32 & 12.02 & 5.16 & 8.10 & 1.36 & 6.88 & 0.10 & 5.36 \\
\hline
\end{tabular}
}
\caption{MSOCR detailed results.}
\label{tab:msocr_detail}
\end{table*}

%% file: Table/MDUR.tex
\begin{table*}[!htbp]
\centering
\resizebox{1\textwidth}{!}{
{\small
\begin{tabular}{lcccccccccc}
\hline
\multirow{2}{*}{\textbf{Model}}    & \multicolumn{10}{c}{\textbf{\texttt{Traditional} setting}}                                             \\ \cline{2-11}
                                       & EN    & ZH    & HU    & RU    & SR    & CS    & AR    & VI    & TH    & KO    \\ \hline
\textbf{Gemini-3.1-Pro-Preview}             & 81.49 & 80.87 & 79.83 & 80.92 & 78.96 & 81.04 & 81.10 & 81.16 & 76.39 & 80.69 \\
\textbf{GPT-5}             & 74.28 & 72.72 & 73.82 & 73.01 & 72.43 & 73.64 & 71.79 & 72.49 & 72.72 & 72.02 \\
\textbf{GPT-5-mini}             & 72.02 & 69.65 & 69.65 & 69.65 & 69.88 & 69.48 & 68.90 & 70.40 & 68.27 & 69.88 \\
\textbf{Llama-4-Maverick}             & 60.87 & 57.46 & 58.15 & 58.21 & 56.76 & 58.55 & 55.20 & 57.57 & 56.01 & 56.18 \\
\textbf{GLM-4.5v}             & 59.69 & 57.94 & 56.22 & 57.79 & 54.24 & 54.53 & 48.78 & 53.90 & 52.24 & 54.94 \\
\textbf{Doubao-Seed-1.6}             & 65.41 & 63.16 & 62.41 & 57.72 & 53.79 & 60.61 & 58.07 & 59.11 & 59.05 & 56.54 \\
\textbf{Qwen3-VL-235B-A22B-Thinking}             & 68.14 & 64.92 & 66.33 & 62.54 & 62.70 & 65.58 & 65.17 & 65.64 & 64.54 & 64.83 \\
\textbf{Qwen3-VL-235B-A22B-Instruct}             & 65.17 & 64.57 & 61.03 & 62.44 & 60.50 & 60.31 & 61.13 & 57.97 & 58.84 & 58.71 \\
\textbf{Qwen3-VL-32B-Thinking}             & 66.10 & 64.40 & 64.27 & 64.54 & 61.67 & 63.13 & 63.90 & 64.51 & 64.42 & 63.49 \\
\midrule
\textbf{Qwen3-VL-8B-Thinking}             & 40.12 & 38.27 & 40.35 & 33.06 & 38.67 & 41.62 & 36.18 & 40.58 & 38.09 & 38.61 \\
\textbf{QGO-8B (ours)}             & 51.04 & 42.89 & 47.86 & 37.80 & 46.24 & 51.04 & 47.57 & 48.84 & 46.94 & 47.92 \\
\hline
\multirow{2}{*}{\textbf{Model}}    & \multicolumn{10}{c}{\textbf{\texttt{Vision} setting}}                                             \\ \cline{2-11}
                                       & EN    & ZH    & HU    & RU    & SR    & CS    & AR    & VI    & TH    & KO    \\ \hline
\textbf{Gemini-3.1-Pro-Preview}             & 80.75 & 79.71 & 79.29 & 78.79 & 77.11 & 79.71 & 77.05 & 79.54 & 73.90 & 78.66 \\
\textbf{GPT-5}             & 72.89 & 68.21 & 69.71 & 69.88 & 70.40 & 69.36 & 63.24 & 71.27 & 66.01 & 69.31 \\
\textbf{GPT-5-mini}             & 72.02 & 68.61 & 68.79 & 70.12 & 69.19 & 69.77 & 63.91 & 69.88 & 69.25 & 69.77 \\
\textbf{Llama-4-Maverick}             & 55.38 & 48.90 & 49.13 & 52.08 & 46.82 & 47.46 & 40.92 & 50.29 & 45.38 & 46.59 \\
\textbf{GLM-4.5v}             & 61.75 & 58.44 & 52.87 & 54.58 & 46.43 & 55.95 & 22.17 & 50.44 & 25.89 & 41.06 \\
\textbf{Doubao-Seed-1.6}             & 65.26 & 63.47 & 60.87 & 55.61 & 50.23 & 62.20 & 41.45 & 57.98 & 34.10 & 52.89 \\
\textbf{Qwen3-VL-235B-A22B-Thinking}             & 66.18 & 63.25 & 64.04 & 63.21 & 59.20 & 64.62 & 56.09 & 62.79 & 53.52 & 62.35 \\
\textbf{Qwen3-VL-235B-A22B-Instruct}             & 64.45 & 62.09 & 55.55 & 59.48 & 57.57 & 59.13 & 50.46 & 53.06 & 49.42 & 53.70 \\
\textbf{Qwen3-VL-32B-Thinking}             & 66.94 & 64.44 & 64.57 & 62.97 & 60.43 & 64.57 & 52.13 & 63.35 & 47.43 & 57.81 \\
\midrule
\textbf{Qwen3-VL-8B-Thinking}             & 44.22 & 41.16 & 38.96 & 25.55 & 38.15 & 38.73 & 24.34 & 40.23 & 21.45 & 36.01 \\
\textbf{QGO-8B (ours)}             & 49.42 & 46.07 & 46.01 & 23.24 & 43.53 & 46.36 & 33.24 & 45.49 & 34.10 & 40.98 \\
\hline
\end{tabular}
}
}
\caption{MDUR detailed results.}
\label{tab:mdur_detail}
\end{table*}

%% file: Table/MIQA.tex
\begin{table*}[!htbp]
\centering
\resizebox{1\textwidth}{!}{
{\small
\begin{tabular}{lcccccccccc}
\hline
\multirow{2}{*}{\textbf{Model}}    & \multicolumn{10}{c}{\textbf{\texttt{Traditional} setting}}                                             \\ \cline{2-11}
                                       & EN    & ZH    & HU    & RU    & SR    & CS    & AR    & VI    & TH    & KO    \\ \hline
\textbf{Gemini-3.1-Pro-Preview}             & 52.37 & 64.30 & 65.16 & 62.40 & 60.79 & 58.11 & 65.68 & 64.52 & 46.90 & 62.45 \\
\textbf{GPT-5}             & 50.44 & 61.65 & 53.51 & 52.29 & 52.89 & 57.12 & 54.36 & 55.21 & 52.64 & 58.05 \\
\textbf{GPT-5-mini}             & 64.01 & 65.44 & 61.09 & 61.89 & 61.22 & 64.75 & 63.46 & 60.03 & 61.44 & 66.51 \\
\textbf{Llama-4-Maverick}             & 49.54 & 47.14 & 41.58 & 51.83 & 46.23 & 45.85 & 43.70 & 49.52 & 41.92 & 43.42 \\
\textbf{GLM-4.5v}             & 53.59 & 56.31 & 49.16 & 51.77 & 51.78 & 51.68 & 45.83 & 50.01 & 46.08 & 50.35 \\
\textbf{Doubao-Seed-1.6}             & 62.32 & 60.36 & 51.99 & 56.05 & 52.74 & 54.13 & 51.37 & 54.65 & 51.08 & 55.91 \\
\textbf{Qwen3-VL-235B-A22B-Thinking}             & 64.47 & 67.35 & 59.67 & 61.91 & 60.26 & 62.64 & 61.16 & 59.91 & 58.41 & 59.78 \\
\textbf{Qwen3-VL-235B-A22B-Instruct}             & 72.54 & 73.33 & 63.25 & 64.19 & 62.44 & 63.33 & 55.93 & 42.81 & 59.92 & 61.88 \\
\textbf{Qwen3-VL-32B-Thinking}             & 60.77 & 60.92 & 50.77 & 50.95 & 53.93 & 56.77 & 54.89 & 52.11 & 53.50 & 55.21 \\
\midrule
\textbf{Qwen3-VL-8B-Thinking}             & 59.46 & 61.13 & 47.64 & 53.31 & 50.28 & 52.75 & 52.39 & 54.95 & 53.17 & 51.23 \\
\textbf{QGO-8B (ours)}             & 59.14 & 62.10 & 50.34 & 54.25 & 52.05 & 53.64 & 53.47 & 57.70 & 55.77 & 53.91 \\
\hline
\multirow{2}{*}{\textbf{Model}}    & \multicolumn{10}{c}{\textbf{\texttt{Vision} setting}}                                             \\ \cline{2-11}
                                       & EN    & ZH    & HU    & RU    & SR    & CS    & AR    & VI    & TH    & KO    \\ \hline
\textbf{Gemini-3.1-Pro-Preview}             & 50.79 & 53.77 & 57.75 & 55.89 & 52.29 & 51.00 & 52.95 & 56.63 & 43.12 & 56.96 \\
\textbf{GPT-5}             & 52.59 & 60.30 & 54.76 & 54.09 & 55.31 & 57.20 & 52.69 & 57.51 & 55.25 & 59.78 \\
\textbf{GPT-5-mini}             & 65.14 & 63.13 & 58.89 & 60.08 & 58.09 & 65.05 & 60.38 & 58.24 & 61.32 & 65.15 \\
\textbf{Llama-4-Maverick}             & 48.54 & 43.87 & 43.65 & 49.84 & 42.94 & 42.39 & 39.66 & 48.61 & 39.90 & 41.09 \\
\textbf{GLM-4.5v}             & 54.85 & 57.21 & 47.93 & 52.03 & 48.90 & 51.79 & 22.11 & 47.51 & 21.46 & 29.88 \\
\textbf{Doubao-Seed-1.6}             & 59.79 & 55.98 & 48.68 & 53.13 & 47.06 & 52.53 & 36.81 & 51.72 & 19.47 & 45.23 \\
\textbf{Qwen3-VL-235B-A22B-Thinking}             & 62.06 & 66.57 & 50.52 & 57.97 & 53.87 & 56.17 & 47.45 & 52.87 & 40.93 & 55.15 \\
\textbf{Qwen3-VL-235B-A22B-Instruct}             & 71.98 & 72.83 & 61.25 & 66.06 & 61.47 & 61.54 & 43.72 & 46.39 & 48.46 & 61.54 \\
\textbf{Qwen3-VL-32B-Thinking}             & 57.22 & 57.17 & 49.68 & 51.20 & 51.84 & 53.37 & 46.95 & 50.49 & 27.71 & 49.00 \\
\midrule
\textbf{Qwen3-VL-8B-Thinking}             & 60.68 & 62.55 & 46.61 & 51.92 & 47.81 & 49.38 & 38.29 & 52.74 & 25.31 & 41.63 \\
\textbf{QGO-8B (ours)}             & 61.97 & 66.03 & 49.68 & 54.86 & 50.70 & 52.14 & 42.88 & 54.68 & 33.98 & 43.65 \\
\hline
\end{tabular}
}
}
\caption{MIQA detailed results.}
\label{tab:miqa_detail}
\end{table*}

%% file: Table/GUI.tex
\begin{table*}[!htbp]
\centering
\resizebox{1\textwidth}{!}{
\small
\begin{tabular}{lcccccccccc}
\hline
\textbf{Model}    & EN    & ZH    & HU    & RU    & SR    & CS    & AR    & VI    & TH    & KO    \\ \hline
\textbf{Gemini-3.1-Pro-Preview}             & 90.00 & 86.50 & 87.00 & 82.00 & 89.00 & 84.50 & 89.00 & 88.00 & 88.50 & 86.50 \\
\textbf{GPT-5}             & 17.50 & 6.53 & 8.00 & 10.00 & 10.50 & 11.06 & 10.50 & 12.50 & 10.00 & 9.05 \\
\textbf{GPT-5-mini}             & 17.09 & 11.06 & 16.50 & 18.09 & 16.08 & 17.26 & 16.58 & 13.50 & 10.05 & 12.06 \\
\textbf{Llama-4-Maverick}             & 67.50 & 30.00 & 62.00 & 39.50 & 61.00 & 64.50 & 45.00 & 63.00 & 59.00 & 56.50 \\
\textbf{GLM-4.5v}             & 87.50 & 84.50 & 78.50 & 82.00 & 78.50 & 85.00 & 46.50 & 81.50 & 50.00 & 70.00 \\
\textbf{Doubao-Seed-1.6}             & 88.00 & 88.50 & 85.00 & 87.00 & 79.50 & 87.00 & 68.50 & 87.00 & 61.00 & 83.00 \\
\textbf{Qwen3-VL-235B-A22B-Thinking}             & 90.00 & 89.50 & 90.50 & 89.50 & 90.00 & 91.50 & 84.50 & 90.00 & 85.50 & 90.50 \\
\textbf{Qwen3-VL-235B-A22B-Instruct}             & 88.50 & 90.00 & 81.50 & 83.50 & 80.00 & 87.50 & 70.50 & 82.50 & 71.50 & 85.50 \\
\textbf{Qwen3-VL-32B-Thinking}             & 88.50 & 89.50 & 86.50 & 88.50 & 88.00 & 88.50 & 80.50 & 87.50 & 76.50 & 89.50 \\
\midrule
\textbf{Qwen3-VL-8B-Thinking}             & 87.50 & 82.50 & 83.00 & 85.00 & 81.00 & 84.00 & 63.50 & 82.50 & 54.00 & 80.00 \\
\textbf{QGO-8B (ours)}             & 87.50 & 86.50 & 81.50 & 86.00 & 82.50 & 82.50 & 66.50 & 83.00 & 64.50 & 79.50 \\
\hline
\end{tabular}
}
\caption{MGUI detailed per-language results (\texttt{vision} setting). Each cell reports the click-grounding accuracy: a prediction $(x, y)$ is correct iff it falls inside the target element's bbox.}
\label{tab:gui_detail}
\end{table*}

%% file: Table/GUI_text.tex
\begin{table*}[!htbp]
\centering
\resizebox{1\textwidth}{!}{
\small
\begin{tabular}{lcccccccccc}
\hline
\textbf{Model}    & EN    & ZH    & HU    & RU    & SR    & CS    & AR    & VI    & TH    & KO    \\ \hline
\textbf{Gemini-3.1-Pro-Preview}             & 98.00 & 96.00 & 97.00 & 98.00 & 97.50 & 97.00 & 94.50 & 97.00 & 98.50 & 98.00 \\
\textbf{GPT-5}             & 94.50 & 94.97 & 96.00 & 95.48 & 93.00 & 95.94 & 88.00 & 95.00 & 93.50 & 93.97 \\
\textbf{GPT-5-mini}             & 93.97 & 93.47 & 94.00 & 93.94 & 92.50 & 93.85 & 87.94 & 94.00 & 85.00 & 90.45 \\
\textbf{Llama-4-Maverick}             & 88.50 & 89.50 & 84.00 & 88.00 & 78.50 & 83.50 & 64.00 & 72.50 & 66.00 & 62.50 \\
\textbf{GLM-4.5v}             & 92.00 & 92.50 & 75.00 & 89.50 & 74.00 & 91.50 & 27.00 & 68.50 & 35.00 & 67.50 \\
\textbf{Doubao-Seed-1.6}             & 94.00 & 93.00 & 77.50 & 85.50 & 56.50 & 86.50 & 60.00 & 86.00 & 41.50 & 77.50 \\
\textbf{Qwen3-VL-235B-A22B-Thinking}             & 94.50 & 96.00 & 92.00 & 94.50 & 94.00 & 94.00 & 79.50 & 89.50 & 74.00 & 90.00 \\
\textbf{Qwen3-VL-235B-A22B-Instruct}             & 93.00 & 93.50 & 85.50 & 93.00 & 90.50 & 92.50 & 79.00 & 87.00 & 68.00 & 87.50 \\
\textbf{Qwen3-VL-32B-Thinking}             & 91.50 & 93.50 & 90.00 & 93.50 & 87.50 & 93.00 & 69.00 & 92.00 & 57.50 & 82.00 \\
\textbf{Qwen3-VL-8B-Thinking}             & 94.00 & 93.50 & 84.50 & 89.50 & 84.00 & 88.50 & 55.00 & 86.00 & 38.50 & 68.50 \\
\hline
\end{tabular}
}
\caption{MGUI\textsubscript{text} per-language results (\texttt{vision} setting). The companion task probes \emph{element identification} on the same screenshots and instructions used by the main MGUI grounding task: instead of emitting click coordinates, the model is asked to output the visible text of the target UI element. A prediction is judged correct if its normalized text matches the target element's visible text. This isolates ``what to click'' (recognition + understanding) from ``where to click'' (spatial grounding) and serves as a sanity-check for the unusually low MGUI grounding scores of GPT-5 / GPT-5-mini reported in~\cref{tab:gui_detail}.}
\label{tab:gui_text_detail}
\end{table*}

%% file: Chapter/Appendix_E.tex
\section{Reliability of LLM-as-Judge on MIQA}
\label{sec: appendix_e}

\subsection{Translate-to-English consistency}
\label{sec: appendix_e_translate_consistency}

While our benchmark is designed to evaluate cross-lingual capability disparities in LVLMs, one might worry that using an LLM-as-judge introduces bias from the inherently uneven cross-lingual performance of the judge itself. To verify that this is not the case, we implemented a \textbf{translate-to-EN-before-judge} strategy: for the MIQA vision task, we translated LVLM responses into English before evaluation. The results for both judging strategies are summarized in~\cref{tab:trans2en_before_judge}.

The translate-to-EN strategy produces results \textbf{nearly identical to those obtained by judging the original (non-English) responses directly}. This indicates that our structured prompt design, which incorporates multiple evaluation dimensions and detailed scoring criteria, effectively mitigates linguistic bias in the LLM-as-judge assessment process. We therefore retain direct LLM-as-judge in all subsequent experiments.

\input{Table/trans2en_vs_direct}

\subsection{Cross-validation with Claude-Opus-4.7}
\label{sec: appendix_e_alt_judge}

\input{Table/claude_vs_deepseek_judge}

To address concerns about single-judge reliability, we re-evaluated MIQA-vision with an independent second judge: \textbf{Claude-Opus-4.7}, replacing our default DeepSeek-v4-flash judge while keeping every other element of the protocol (the same six scoring dimensions, the same human-written references, the same prompt template) fixed. The second judge was applied to the three models for which full re-evaluations were available: GPT-5 (commercial leader), Qwen3-VL-8B-Thinking (our open-source baseline), and QGO-8B (our trained model). \cref{tab:claude_vs_deepseek_judge} reports the per-language comparison.

Across all 10 languages and 3 models, the two judges agree closely: mean $|\Delta|$ is $1.24$, $2.27$, and $2.01$ points respectively. Crucially, the relative ordering of models is preserved on every language and the cross-lingual pattern (large drop on AR/TH, strong on EN/ZH) reproduces under the alternate judge. This confirms that the cross-lingual disparities we report are properties of the evaluated LVLMs rather than artifacts of our judge choice.

\FloatBarrier

%% file: Table/trans2en_vs_direct.tex
\begin{table*}[!htbp]
\centering
\begin{tabular*}{\textwidth}{l @{\extracolsep{\fill}} ccc}
\toprule
\textbf{Model} & \textbf{Direct Judge} & \textbf{Translate to EN before Judge} & \textbf{$\Delta$} \\
\midrule
\textbf{GPT-5-mini} & 61.15 & 58.74 & -2.41 \\
\textbf{Qwen3-VL-235B-A22B-Instruct} & 58.14 & 56.52 & -1.62 \\
\textbf{GPT-5} & 56.32 & 54.35 & -1.97 \\
\textbf{Qwen3-VL-235B-A22B-Thinking} & 53.50 & 52.00 & -1.50 \\
\textbf{Gemini-3.1-Pro-Preview} & 53.37 & 51.59 & -1.79 \\
\textbf{QGO-8B} & 49.85 & 46.82 & -3.03 \\
\textbf{Qwen3-VL-32B-Thinking} & 48.60 & 47.46 & -1.14 \\
\textbf{Qwen3-VL-8B-Thinking} & 46.25 & 43.70 & -2.55 \\
\textbf{Doubao-Seed-1.6} & 45.62 & 44.69 & -0.93 \\
\textbf{Llama-4-Maverick} & 43.55 & 42.99 & -0.56 \\
\textbf{GLM-4.5v} & 42.09 & 41.20 & -0.89 \\
\bottomrule
\end{tabular*}
\caption{The scores represent the average of 9 non-English language evaluations, with a maximum possible score of 100.}
\label{tab:trans2en_before_judge}
\end{table*}

%% file: Table/claude_vs_deepseek_judge.tex
\begin{table*}[!htbp]
\centering
\renewcommand{\arraystretch}{1.05}
\setlength{\tabcolsep}{4pt}
\resizebox{\textwidth}{!}{%
\begin{tabular}{l l c c c c c c c c c c c}
\toprule
\textbf{Model} & \textbf{Judge} & \textbf{AR} & \textbf{CS} & \textbf{EN} & \textbf{HU} & \textbf{KO} & \textbf{RU} & \textbf{SR} & \textbf{TH} & \textbf{VI} & \textbf{ZH} & \textbf{Avg.} \\
\midrule
\multirow{3}{*}{GPT-5} & DeepSeek-v4-flash & 52.69 & 57.20 & 52.59 & 54.76 & 59.78 & 54.09 & 55.31 & 55.25 & 57.51 & 60.30 & 55.95 \\
                       & Claude-Opus-4.7   & 54.20 & 60.30 & 54.80 & 55.80 & 61.10 & 54.30 & 55.70 & 56.70 & 58.20 & 60.80 & 57.19 \\
                       & $\Delta$          & $+1.51$ & $+3.10$ & $+2.21$ & $+1.04$ & $+1.32$ & $+0.21$ & $+0.39$ & $+1.45$ & $+0.69$ & $+0.50$ & $+1.24$ \\
\midrule
\multirow{3}{*}{Qwen3-VL-8B-Thinking} & DeepSeek-v4-flash & 38.29 & 49.38 & 60.68 & 46.61 & 41.63 & 51.92 & 47.81 & 25.31 & 52.74 & 62.55 & 47.69 \\
                                       & Claude-Opus-4.7   & 36.00 & 47.20 & 59.20 & 42.40 & 38.40 & 51.10 & 43.50 & 25.00 & 51.10 & 60.30 & 45.42 \\
                                       & $\Delta$          & $-2.29$ & $-2.18$ & $-1.48$ & $-4.21$ & $-3.23$ & $-0.82$ & $-4.31$ & $-0.31$ & $-1.64$ & $-2.25$ & $-2.27$ \\
\midrule
\multirow{3}{*}{QGO-8B (ours)} & DeepSeek-v4-flash & 42.88 & 52.14 & 61.97 & 49.68 & 43.65 & 54.86 & 50.70 & 33.98 & 54.68 & 66.03 & 51.06 \\
                                & Claude-Opus-4.7   & 44.80 & 53.90 & 64.20 & 51.20 & 46.90 & 57.60 & 51.80 & 37.60 & 57.90 & 64.80 & 53.07 \\
                                & $\Delta$          & $+1.92$ & $+1.76$ & $+2.23$ & $+1.52$ & $+3.25$ & $+2.74$ & $+1.10$ & $+3.62$ & $+3.22$ & $-1.23$ & $+2.01$ \\
\bottomrule
\end{tabular}%
}
\caption{MIQA-vision per-language scores under our default DeepSeek-v4-flash judge versus a Claude-Opus-4.7 second judge. Both judges score on a 0--100 scale (six dimensions averaged, then $\times 10$). $\Delta$ = Claude $-$ DeepSeek. Mean $|\Delta|$ across languages is $1.24$, $2.27$, $2.01$ for the three models respectively, showing no systematic per-language bias that would alter our cross-lingual conclusions.}
\label{tab:claude_vs_deepseek_judge}
\end{table*}

%% file: Chapter/Appendix_F.tex
\section{External-Benchmark Per-Language Results}
\label{sec: appendix_f}

\S6.3 reports aggregated results of QGO-8B on three external multilingual suites; this appendix lists the per-language breakdowns.

\textbf{CC-OCR}~\cite{liu2024ccocr} is a 10-language full-text OCR set (150 samples per language). We use a fixed evaluator of our own: after NFKC, case, punctuation, and whitespace normalization, we macro-average sample-level Normalized Edit Distance, $1-d(p,g)/\max(|p|,|g|)$, and order-insensitive character-multiset F1. These metrics are distinct from CC-OCR's official Eval-Trans implementation. \cref{tab:cc_ocr_detail} shows that QGO-8B improves on every one of the 10 languages, with a +5.7 trained-language average and a +4.0 average on six languages (FR, DE, IT, JA, PT, ES) that were \emph{never} seen during RL, direct evidence of cross-lingual transfer.

\textbf{MTVQA}~\cite{tang2024mtvqa} is a multilingual text-centric VQA benchmark containing both direct text-reading and reasoning-dependent questions. Five of its nine languages overlap with our training set. \cref{tab:mtvqa_detail} reports per-language accuracy. Trained-language scores improve modestly (+0.5 on average), while unseen-language scores are essentially preserved ($-0.4$ on average).

\textbf{OCRBench}~\cite{liu2024ocrbench} is a single-score benchmark whose released 1,000-example v1 test set provides a fixed 10-sub-task taxonomy but no language field or per-language split. As \cref{tab:ocrbench_detail} shows, QGO-8B matches the baseline overall (791 vs 789) and is within $\pm 6$ points on every sub-task, confirming no regression of base OCR capability.

\input{Table/external_bench}

\FloatBarrier

%% file: Table/external_bench.tex
\begin{table}[!t]
\centering
\footnotesize
\renewcommand{\arraystretch}{1.1}
\setlength{\tabcolsep}{4pt}
\begin{tabular*}{\columnwidth}{@{\extracolsep{\fill}} l c c c @{}}
\toprule
\textbf{Lang} & \textbf{Baseline} & \textbf{QGO-8B} & \textbf{$\Delta$} \\
\midrule
\multicolumn{4}{l}{\textit{Trained languages (overlap with our 10)}} \\
AR & 36.8 & 47.8 & \textbf{+11.0} \\
KR & 57.0 & 61.9 & +4.9 \\
RU & 73.8 & 77.5 & +3.7 \\
VI & 53.1 & 56.3 & +3.2 \\
\midrule
\multicolumn{4}{l}{\textit{Unseen languages}} \\
FR & 77.0 & 80.0 & +3.0 \\
DE & 69.5 & 74.6 & +5.1 \\
IT & 67.7 & 73.2 & +5.4 \\
JA & 46.7 & 51.6 & +4.9 \\
PT & 76.0 & 77.9 & +1.9 \\
ES & 67.6 & 71.5 & +3.9 \\
\midrule
\textbf{Trained avg} & 55.2 & 60.9 & +5.7 \\
\textbf{Unseen avg}  & 67.4 & 71.5 & +4.0 \\
\textbf{Overall}     & 62.5 & 67.2 & \textbf{+4.7} \\
\textbf{Char-F1 (ours)} & 80.2 & 87.0 & +6.8 \\
\bottomrule
\end{tabular*}
\caption{CC-OCR~\cite{liu2024ccocr} results under our fixed evaluator. Language rows and grouped averages report NED (\%); the final row reports macro-averaged character-multiset F1 (\%). Higher is better. All 10 languages improve; the largest gain is on Arabic (+11.0). Six unseen languages (FR/DE/IT/JA/PT/ES) gain +4.0 on average, evidencing cross-lingual transfer beyond the training set.}
\label{tab:cc_ocr_detail}
\end{table}

\begin{table}[!t]
\centering
\footnotesize
\renewcommand{\arraystretch}{1.1}
\setlength{\tabcolsep}{4pt}
\begin{tabular*}{\columnwidth}{@{\extracolsep{\fill}} l c c c @{}}
\toprule
\textbf{Lang} & \textbf{Baseline} & \textbf{QGO-8B} & \textbf{$\Delta$} \\
\midrule
\multicolumn{4}{l}{\textit{Trained languages}} \\
AR & 20.1 & 20.5 & +0.4 \\
KR & 42.3 & 42.5 & +0.2 \\
TH & 28.6 & 29.4 & +0.9 \\
VI & 44.8 & 46.2 & +1.4 \\
RU & 17.6 & 17.2 & $-0.4$ \\
\midrule
\multicolumn{4}{l}{\textit{Unseen languages}} \\
DE & 38.6 & 39.2 & +0.6 \\
FR & 48.3 & 47.9 & $-0.5$ \\
IT & 43.8 & 43.2 & $-0.6$ \\
JA & 26.1 & 24.8 & $-1.3$ \\
\midrule
\textbf{Trained avg} & 30.7 & 31.1 & +0.5 \\
\textbf{Unseen avg}  & 39.2 & 38.8 & $-0.4$ \\
\textbf{Overall}     & 34.5 & 34.5 & +0.1 \\
\bottomrule
\end{tabular*}
\caption{Per-language MTVQA~\cite{tang2024mtvqa} results. Accuracy (\%); higher is better. The trained-language gain (+0.5) is positive but modest on this benchmark, which contains both direct text-reading and reasoning-dependent questions; unseen-language scores are essentially preserved.}
\label{tab:mtvqa_detail}
\end{table}

\begin{table}[!t]
\centering
\footnotesize
\renewcommand{\arraystretch}{1.1}
\setlength{\tabcolsep}{4pt}
\begin{tabular*}{\columnwidth}{@{\extracolsep{\fill}} l c c c @{}}
\toprule
\textbf{Sub-task} & \textbf{Baseline} & \textbf{QGO-8B} & \textbf{$\Delta$} \\
\midrule
Regular Text Recog.            & 98.0 & 98.0 & $\phantom{+}0.0$ \\
Irregular Text Recog.          & 94.0 & 96.0 & $+2.0$ \\
Artistic Text Recog.           & 94.0 & 92.0 & $-2.0$ \\
Handwriting Recog.             & 72.0 & 66.0 & $-6.0$ \\
Digit String Recog.            & 70.0 & 72.0 & $+2.0$ \\
Non-Semantic Text Recog.       & 92.0 & 94.0 & $+2.0$ \\
Scene Text-centric VQA         & 85.0 & 85.5 & $+0.5$ \\
Doc-oriented VQA               & 92.5 & 92.0 & $-0.5$ \\
Key Information Extraction     & 87.0 & 87.0 & $\phantom{+}0.0$ \\
HMER\textsuperscript{$\dagger$}        & 0.0  & 0.0  & $\phantom{+}0.0$ \\
\midrule
\textbf{Total (out of 1000)}   & 789  & 791  & $+2$  \\
\bottomrule
\end{tabular*}
\caption{Per-sub-task results on the released 1,000-example OCRBench v1 test set~\cite{liu2024ocrbench}. The release provides a fixed 10-task taxonomy but no language field or per-language split, so this is the finest available breakdown. QGO-8B matches the baseline overall, confirming no degradation of base OCR ability despite our RL training using none of OCRBench's data distribution. \textsuperscript{$\dagger$}HMER = Handwritten Mathematical Expression Recognition; both models score $0$ under the benchmark's strict string-containment evaluator.}
\label{tab:ocrbench_detail}
\end{table}

%% file: Chapter/Appendix_G.tex
\section{Per-language PCC Analysis}
\label{sec: appendix_g_pcc}

To probe whether OCR capability is linked to downstream task performance \emph{across languages}, we compute Pearson Correlation Coefficients (PCCs) between three pairs of per-language score vectors (length 10, one entry per language) for each of the 10 evaluated LVLMs: \emph{MSOCR vs.\ MDUR\textsubscript{vision}}, \emph{MSOCR vs.\ MIQA\textsubscript{vision}}, and \emph{MSOCR vs.\ MGUI}. Given two such per-language score vectors $\mathbf{x}, \mathbf{y} \in \mathbb{R}^{10}$, PCC is defined as
\begin{align}
D_x &= \sqrt{\sum_{i=1}^{10} (x_i - \bar{x})^2}, \notag\\
D_y &= \sqrt{\sum_{i=1}^{10} (y_i - \bar{y})^2}, \notag\\
\mathrm{PCC}(\mathbf{x}, \mathbf{y})
&= \frac{\sum_{i=1}^{10} (x_i - \bar{x})(y_i - \bar{y})}{D_x D_y}.
\end{align}
where $\bar{x}$ and $\bar{y}$ denote the language-wise means; $|\mathrm{PCC}| > 0.5$ is conventionally read as a strong linear correlation. The first pair probes the link between multi-scale OCR and reasoning VQA on visually rendered text; the second probes the link to open-ended generation; the third probes the link to coordinate-based grounding on multilingual screenshots. The full per-model heatmap visualizing these correlations is shown in the main text (\cref{fig:pcc_heatmap_main}): MSOCR vs.\ MDUR-vision is consistently strong across all 10 evaluated models; MIQA-vision and MGUI also exhibit predominantly strong positive correlations, with a few model-specific exceptions (e.g.\ GPT-5 on MIQA, Gemini-3.1-Pro on MGUI). The pattern supports our claim that OCR is a key cross-lingual factor.

\FloatBarrier

%% file: Chapter/Appendix_H.tex
\section{Training and Inference Configuration}
\label{sec: appendix_h_training_config}

This appendix details the inference setup we apply to all 10 evaluated LVLMs and the training recipe used to obtain QGO-8B from the Qwen3-VL-8B-Thinking base model.

\paragraph{Inference setup.} For all 10 evaluated models on every PM\textsuperscript{4}Bench task, we use \textbf{greedy decoding} (temperature $0$) and apply each model's official default chat template. Greedy decoding removes deliberate sampling and minimizes decoding stochasticity, but it does not guarantee bitwise-identical outputs across repeated runs. We therefore use this uniform low-stochasticity configuration without treating it as proof of exact determinism. Per-task user prompts (one figure per task) and the LLM-as-judge prompt for MIQA are listed in \cref{sec: appendix_b}.

\paragraph{Training framework and optimizer.} The OCR-centric GRPO post-training of QGO-8B is implemented on top of VeRL~\cite{sheng2024hybridflow}, a flexible RLHF framework. We optimize the base Qwen3-VL-8B-Thinking with AdamW at learning rate $1\times10^{-6}$. Each optimization step uses $32$ prompts and generates $8$ rollouts per prompt, yielding a global rollout batch of $256$ sampled trajectories. Rewards are computed per rollout and the group-relative advantage is used to update the policy.

\paragraph{Supervised baselines.} The Full SFT and LoRA SFT baselines in \cref{tab:post_training_comparison} use the same Qwen3-VL-8B-Thinking initialization and the same synthetic OCR corpus as QGO-8B. Both use a maximum sequence length of $12{,}288$, global batch size $16$, and seed $42$. Full SFT updates all parameters at learning rate $1\times10^{-5}$. LoRA uses rank $16$, $\alpha=32$, all linear layers as targets, a frozen vision encoder, and learning rate $1\times10^{-4}$. These are representative method-specific configurations rather than a compute-matched hyperparameter sweep, which is why the main-text comparison is interpreted diagnostically.

\begin{figure}[!t]
    \centering
    \includegraphics[width=\columnwidth, keepaspectratio]{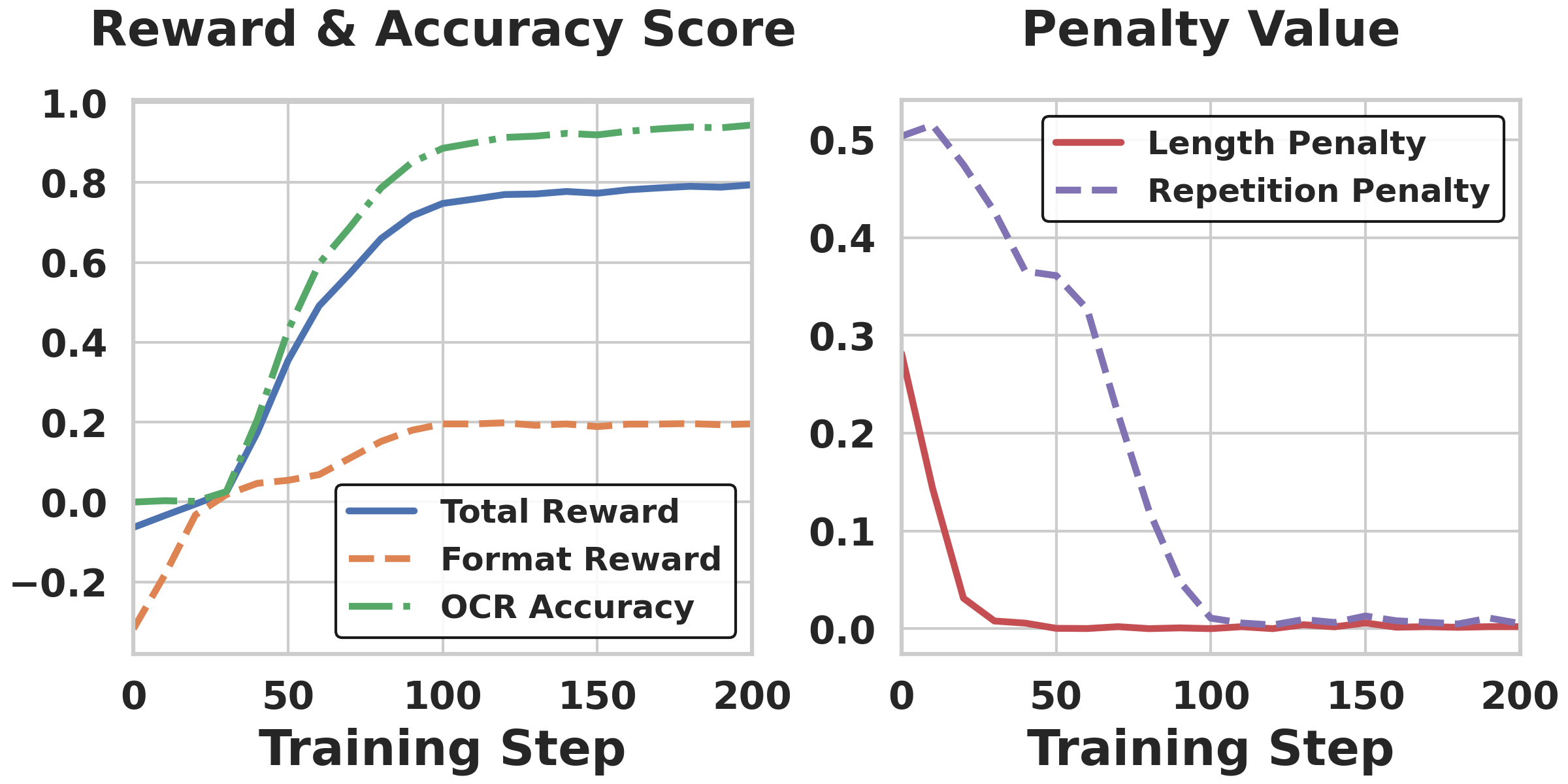}
    \caption{Training dynamics of OCR-centric GRPO. \textbf{Left:} The evolution of total reward, $R_{\text{fmt}}$, and $R_{\text{acc}}$ over training steps on validation split, calculated every 10 steps. \textbf{Right:} The reduction in $P_{\text{len}}$ and $P_{\text{rep}}$.}
    \label{fig:training_curves}
\end{figure}

\paragraph{Convergence and compute cost.} Empirically, the training reward stabilizes well within the first $200$ optimization steps (see~\cref{fig:training_curves}), and we use the step-$200$ checkpoint for QGO-8B. Reaching this checkpoint takes approximately $9$ hours on $8$ NVIDIA H200 GPUs. Crucially, our reward is computed entirely from synthetic OCR data without any human annotation or task-specific VQA supervision, and the base model is small (8B parameters), which makes our recipe a low-friction add-on that can be re-run after each base-model release.

\paragraph{Reward hyperparameters.} During training we empirically set $L_{\min} = 1000$, $L_{\max} = 10000$, $\lambda = 1200$, $\tau = 0.6$, $\alpha_1 = 0.2$, $\alpha_2 = 0.8$, and $\alpha_3 = 0.4$. Under the notation of \cref{eq:final_reward}, the accuracy and format weights are respectively $\omega_1 = 0.8$ and $\omega_2 = 0.2$. The format reward is intentionally weighted lower than the accuracy reward ($\omega_1 / \omega_2 = 4$): satisfying the format and length constraint should yield only a sustaining signal, while the OCR accuracy term drives the optimization trajectory. This weighting preserves the model's underlying chain-of-thought reasoning behavior, which is critical for the non-OCR transfer results reported in the main text.

\FloatBarrier

%% file: custom.bib
@article{deepseek-math,
  author    = {Zhihong Shao and Peiyi Wang and Qihao Zhu and Runxin Xu and Junxiao Song and Xiao Bi and Haowei Zhang and Mingchuan Zhang and Y. K. Li and Y. Wu and Daya Guo},
  title     = {DeepSeekMath: Pushing the Limits of Mathematical Reasoning in Open Language Models},
  journal   = {arXiv preprint arXiv:2402.03300},
  year      = {2024},
  url       = {https://arxiv.org/abs/2402.03300}
}

@misc{openai2025gpt5systemcard,
  title        = {{GPT}-5 System Card},
  author       = {{OpenAI}},
  year         = {2025},
  howpublished = {\url{https://openai.com/index/gpt-5-system-card/}},
  note         = {Accessed: 2026-08-22}
}

@misc{meta2025llama4maverick,
  title        = {The {Llama} 4 Herd: The Beginning of a New Era of Natively Multimodal {AI} Innovation},
  author       = {{Meta AI}},
  year         = {2025},
  howpublished = {\url{https://ai.meta.com/blog/llama-4-multimodal-intelligence/}},
  note         = {Accessed: 2026-08-22}
}

@misc{bytedance2025doubaoseed16,
  title        = {Introduction to Techniques Used in {Seed1.6}},
  author       = {{ByteDance Seed}},
  year         = {2025},
  howpublished = {\url{https://seed.bytedance.com/blog/introduction-to-techniques-used-in-seed1-6}},
  note         = {Accessed: 2026-08-22}
}

@misc{google2025gemini3propreview,
  title        = {{Gemini} 3.1 Pro Model Card},
  author       = {{Google DeepMind}},
  year         = {2026},
  howpublished = {\url{https://deepmind.google/models/model-cards/gemini-3-1-pro/}},
  note         = {Accessed: 2026-08-22}
}

@misc{anthropic2025claudesonnet45,
  title        = {Introducing {Claude Sonnet} 4.5},
  author       = {{Anthropic}},
  year         = {2025},
  howpublished = {\url{https://www.anthropic.com/news/claude-sonnet-4-5}},
  note         = {Accessed: 2026-08-22}
}

@misc{anthropic2026claudeopus47,
  title        = {Claude Opus 4.7 System Card},
  author       = {{Anthropic}},
  year         = {2026},
  howpublished = {\url{https://www.anthropic.com/system-cards}},
  note         = {Accessed: 2026-08-22}
}

@misc{vteam2025glm45vglm41vthinkingversatilemultimodal,
      title={GLM-4.5V and GLM-4.1V-Thinking: Towards Versatile Multimodal Reasoning with Scalable Reinforcement Learning},
      author={{GLM-V Team}},
      year={2025},
      eprint={2507.01006},
      archivePrefix={arXiv},
      primaryClass={cs.CV},
      url={https://arxiv.org/abs/2507.01006}
}

@article{Qwen3-VL,
      title={Qwen3-VL Technical Report}, 
      author={Shuai Bai and Yuxuan Cai and Ruizhe Chen and Keqin Chen and Xionghui Chen and Zesen Cheng and Lianghao Deng and Wei Ding and Chang Gao and Chunjiang Ge and Wenbin Ge and Zhifang Guo and Qidong Huang and Jie Huang and Fei Huang and Binyuan Hui and Shutong Jiang and Zhaohai Li and Mingsheng Li and Mei Li and Kaixin Li and Zicheng Lin and Junyang Lin and Xuejing Liu and Jiawei Liu and Chenglong Liu and Yang Liu and Dayiheng Liu and Shixuan Liu and Dunjie Lu and Ruilin Luo and Chenxu Lv and Rui Men and Lingchen Meng and Xuancheng Ren and Xingzhang Ren and Sibo Song and Yuchong Sun and Jun Tang and Jianhong Tu and Jianqiang Wan and Peng Wang and Pengfei Wang and Qiuyue Wang and Yuxuan Wang and Tianbao Xie and Yiheng Xu and Haiyang Xu and Jin Xu and Zhibo Yang and Mingkun Yang and Jianxin Yang and An Yang and Bowen Yu and Fei Zhang and Hang Zhang and Xi Zhang and Bo Zheng and Humen Zhong and Jingren Zhou and Fan Zhou and Jing Zhou and Yuanzhi Zhu and Ke Zhu},
	  journal={arXiv preprint arXiv:2511.21631},
      year={2025},
      url={https://arxiv.org/abs/2511.21631}
}

@misc{deepseekai2026deepseekv4,
      title={DeepSeek-V4: Towards Highly Efficient Million-Token Context Intelligence},
      author={{DeepSeek-AI}},
      year={2026},
      eprint={2606.19348},
      archivePrefix={arXiv},
      primaryClass={cs.CL},
      url={https://arxiv.org/abs/2606.19348}
}

@inproceedings{sheng2024hybridflow,
  title   = {HybridFlow: A Flexible and Efficient RLHF Framework},
  author  = {Guangming Sheng and Chi Zhang and Zilingfeng Ye and Xibin Wu and Wang Zhang and Ru Zhang and Yanghua Peng and Haibin Lin and Chuan Wu},
  year    = {2025},
  booktitle = {Proceedings of the Twentieth European Conference on Computer Systems},
  pages   = {1279--1297},
  doi     = {10.1145/3689031.3696075},
  url     = {https://doi.org/10.1145/3689031.3696075}
}

@misc{kimiteam2025kimik2openagentic,
      title={Kimi K2: Open Agentic Intelligence}, 
      author={{Kimi Team}},
      year={2025},
      eprint={2507.20534},
      archivePrefix={arXiv},
      primaryClass={cs.LG},
      url={https://arxiv.org/abs/2507.20534}
}

@inproceedings{sun2025parrot,
  title={Parrot: Multilingual Visual Instruction Tuning},
  author={Sun, Hai-Long and Zhou, Da-Wei and Li, Yang and Lu, Shiyin and Yi, Chao and Chen, Qing-Guo and Xu, Zhao and Luo, Weihua and Zhang, Kaifu and Zhan, De-Chuan and Ye, Han-Jia},
  booktitle={Proceedings of the 42nd International Conference on Machine Learning},
  series={Proceedings of Machine Learning Research},
  volume={267},
  pages={57984--58007},
  year={2025},
  publisher={PMLR},
  url={https://proceedings.mlr.press/v267/sun25ad.html}
}

@inproceedings{wendler2024llamas,
  title={Do Llamas Work in {E}nglish? On the Latent Language of Multilingual Transformers},
  author={Wendler, Chris and Veselovsky, Veniamin and Monea, Giovanni and West, Robert},
  booktitle={Proceedings of the 62nd Annual Meeting of the Association for Computational Linguistics (Volume 1: Long Papers)},
  pages={15366--15394},
  year={2024},
  publisher={Association for Computational Linguistics},
  doi={10.18653/v1/2024.acl-long.820},
  url={https://aclanthology.org/2024.acl-long.820/}
}

@inproceedings{tang2024language,
  title={Language-specific neurons: The key to multilingual capabilities in large language models},
  author={Tang, Tianyi and Luo, Wenyang and Huang, Haoyang and Zhang, Dongdong and Wang, Xiaolei and Zhao, Xin and Wei, Furu and Wen, Ji-Rong},
  booktitle={Proceedings of the 62nd Annual Meeting of the Association for Computational Linguistics (Volume 1: Long Papers)},
  pages={5701--5715},
  year={2024},
  publisher={Association for Computational Linguistics},
  doi={10.18653/v1/2024.acl-long.309},
  url={https://aclanthology.org/2024.acl-long.309/}
}

@inproceedings{zhao2024large,
  title={How do Large Language Models Handle Multilingualism?},
  author={Zhao, Yiran and Zhang, Wenxuan and Chen, Guizhen and Kawaguchi, Kenji and Bing, Lidong},
  booktitle={Advances in Neural Information Processing Systems},
  volume={37},
  pages={15296--15319},
  year={2024},
  doi={10.52202/079017-0489},
  url={https://doi.org/10.52202/079017-0489}
}

@inproceedings{xue2021mt5massivelymultilingualpretrained,
  title={mT5: A Massively Multilingual Pre-trained Text-to-Text Transformer},
  author={Linting Xue and Noah Constant and Adam Roberts and Mihir Kale and Rami Al-Rfou and Aditya Siddhant and Aditya Barua and Colin Raffel},
  booktitle={Proceedings of the 2021 Conference of the North American Chapter of the Association for Computational Linguistics: Human Language Technologies},
  pages={483--498},
  year={2021},
  publisher={Association for Computational Linguistics},
  doi={10.18653/v1/2021.naacl-main.41},
  url={https://aclanthology.org/2021.naacl-main.41/}
}

@misc{yu2025wanjuansiluhighqualityopensourcewebtext,
      title={WanJuanSiLu: A High-Quality Open-Source Webtext Dataset for Low-Resource Languages}, 
      author={Jia Yu and Fei Yuan and Rui Min and Jing Yu and Pei Chu and Jiayang Li and Wei Li and Ruijie Zhang and Zhenxiang Li and Zhifei Ren and Dong Zheng and Wenjian Zhang and Yan Teng and Lingyu Meng and ZhenJiang Jin and Jiantao Qiu and ShaSha Wang and Zhongying Tu and Dahua Lin and Yu Wang and Yu Qiao and Yanfeng Wang and Conghui He},
      year={2025},
      eprint={2501.14506},
      archivePrefix={arXiv},
      primaryClass={cs.CL},
      url={https://arxiv.org/abs/2501.14506}
}

@article{zhu2024power,
  title={The Power of Question Translation Training in Multilingual Reasoning: Broadened Scope and Deepened Insights},
  author={Zhu, Wenhao and Huang, Shujian and Yuan, Fei and Chen, Cheng and Chen, Jiajun and Birch, Alexandra},
  journal={arXiv preprint arXiv:2405.01345},
  year={2024},
  url={https://arxiv.org/abs/2405.01345}
}

@inproceedings{she2024mapo,
  title={Mapo: Advancing multilingual reasoning through multilingual alignment-as-preference optimization},
  author={She, Shuaijie and Zou, Wei and Huang, Shujian and Zhu, Wenhao and Liu, Xiang and Geng, Xiang and Chen, Jiajun},
  booktitle={Proceedings of the 62nd Annual Meeting of the Association for Computational Linguistics (Volume 1: Long Papers)},
  pages={10015--10027},
  year={2024},
  publisher={Association for Computational Linguistics},
  doi={10.18653/v1/2024.acl-long.539},
  url={https://aclanthology.org/2024.acl-long.539/}
}

@inproceedings{zhu2024question,
  title={Question translation training for better multilingual reasoning},
  author={Zhu, Wenhao and Huang, Shujian and Yuan, Fei and She, Shuaijie and Chen, Jiajun and Birch, Alexandra},
  booktitle={Findings of the Association for Computational Linguistics: ACL 2024},
  pages={8411--8423},
  year={2024},
  publisher={Association for Computational Linguistics},
  doi={10.18653/v1/2024.findings-acl.498},
  url={https://aclanthology.org/2024.findings-acl.498/}
}

@inproceedings{huang2023not,
  title={Not all languages are created equal in llms: Improving multilingual capability by cross-lingual-thought prompting},
  author={Huang, Haoyang and Tang, Tianyi and Zhang, Dongdong and Zhao, Xin and Song, Ting and Xia, Yan and Wei, Furu},
  booktitle={Findings of the Association for Computational Linguistics: EMNLP 2023},
  pages={12365--12394},
  year={2023},
  publisher={Association for Computational Linguistics},
  doi={10.18653/v1/2023.findings-emnlp.826},
  url={https://aclanthology.org/2023.findings-emnlp.826/}
}

@inproceedings{sun2024benchmarking,
  title={Benchmarking Chinese Commonsense Reasoning of LLMs: From Chinese-Specifics to Reasoning-Memorization Correlations},
  author={Sun, Jiaxing and Huang, Weiquan and Wu, Jiang and Gu, Chenya and Li, Wei and Zhang, Songyang and Yan, Hang and He, Conghui},
  booktitle={Proceedings of the 62nd Annual Meeting of the Association for Computational Linguistics (Volume 1: Long Papers)},
  pages={11205--11228},
  year={2024},
  address={Bangkok, Thailand},
  publisher={Association for Computational Linguistics},
  doi={10.18653/v1/2024.acl-long.604},
  url={https://aclanthology.org/2024.acl-long.604/}
}

@inproceedings{geigle-etal-2025-centurio,
  title={Centurio: On Drivers of Multilingual Ability of Large Vision-Language Model},
  author={Geigle, Gregor and Schneider, Florian and Holtermann, Carolin and Biemann, Chris and Timofte, Radu and Lauscher, Anne and Glava{\v{s}}, Goran},
  booktitle={Proceedings of the 63rd Annual Meeting of the Association for Computational Linguistics (Volume 1: Long Papers)},
  pages={2831--2881},
  year={2025},
  address={Vienna, Austria},
  publisher={Association for Computational Linguistics},
  doi={10.18653/v1/2025.acl-long.143},
  url={https://aclanthology.org/2025.acl-long.143/}
}

@inproceedings{zhang2024p,
  title={P-MMEval: A Parallel Multilingual Multitask Benchmark for Consistent Evaluation of LLMs},
  author={Zhang, Yidan and Wan, Yu and Deng, Boyi and Yang, Baosong and Wei, Hao-Ran and Huang, Fei and Yu, Bowen and Liu, Dayiheng and Lin, Junyang and Huang, Fei and Zhou, Jingren},
  booktitle={Proceedings of the 2025 Conference on Empirical Methods in Natural Language Processing},
  pages={4809--4836},
  year={2025},
  publisher={Association for Computational Linguistics},
  doi={10.18653/v1/2025.emnlp-main.242},
  url={https://aclanthology.org/2025.emnlp-main.242/}
}

@inproceedings{huang2025benchmax,
  title={BenchMAX: A Comprehensive Multilingual Evaluation Suite for Large Language Models},
  author={Huang, Xu and Zhu, Wenhao and Hu, Hanxu and He, Conghui and Li, Lei and Huang, Shujian and Yuan, Fei},
  booktitle={Findings of the Association for Computational Linguistics: EMNLP 2025},
  pages={16751--16774},
  year={2025},
  publisher={Association for Computational Linguistics},
  doi={10.18653/v1/2025.findings-emnlp.909},
  url={https://aclanthology.org/2025.findings-emnlp.909/}
}

@inproceedings{yue2024mmmu,
  title={Mmmu-pro: A more robust multi-discipline multimodal understanding benchmark},
  author={Yue, Xiang and Zheng, Tianyu and Ni, Yuansheng and Wang, Yubo and Zhang, Kai and Tong, Shengbang and Sun, Yuxuan and Yu, Botao and Zhang, Ge and Sun, Huan and Su, Yu and Chen, Wenhu and Neubig, Graham},
  booktitle={Proceedings of the 63rd Annual Meeting of the Association for Computational Linguistics (Volume 1: Long Papers)},
  pages={15134--15186},
  year={2025},
  publisher={Association for Computational Linguistics},
  doi={10.18653/v1/2025.acl-long.736},
  url={https://aclanthology.org/2025.acl-long.736/}
}

@inproceedings{das2024exams,
  title={EXAMS-V: A Multi-Discipline Multilingual Multimodal Exam Benchmark for Evaluating Vision Language Models},
  author={Das, Rocktim and Hristov, Simeon and Li, Haonan and Dimitrov, Dimitar and Koychev, Ivan and Nakov, Preslav},
  booktitle={Proceedings of the 62nd Annual Meeting of the Association for Computational Linguistics (Volume 1: Long Papers)},
  pages={7768--7791},
  year={2024},
  publisher={Association for Computational Linguistics},
  doi={10.18653/v1/2024.acl-long.420},
  url={https://aclanthology.org/2024.acl-long.420/}
}

@inproceedings{liu2024mmdu,
  title={MMDU: A Multi-Turn Multi-Image Dialog Understanding Benchmark and Instruction-Tuning Dataset for LVLMs},
  author={Liu, Ziyu and Chu, Tao and Zang, Yuhang and Wei, Xilin and Dong, Xiaoyi and Zhang, Pan and Liang, Zijian and Xiong, Yuanjun and Qiao, Yu and Lin, Dahua and Wang, Jiaqi},
  booktitle={Advances in Neural Information Processing Systems},
  volume={37},
  pages={8698--8733},
  year={2024},
  doi={10.52202/079017-0278},
  url={https://doi.org/10.52202/079017-0278}
}

@inproceedings{pfeiffer2021xgqa,
  title={xGQA: Cross-lingual visual question answering},
  author={Pfeiffer, Jonas and Geigle, Gregor and Kamath, Aishwarya and Steitz, Jan-Martin O and Roth, Stefan and Vuli{\'c}, Ivan and Gurevych, Iryna},
  booktitle={Findings of the Association for Computational Linguistics: ACL 2022},
  pages={2497--2511},
  year={2022},
  publisher={Association for Computational Linguistics},
  doi={10.18653/v1/2022.findings-acl.196},
  url={https://aclanthology.org/2022.findings-acl.196/}
}

@inproceedings{changpinyo2022maxm,
  title={Maxm: Towards multilingual visual question answering},
  author={Changpinyo, Soravit and Xue, Linting and Yarom, Michal and Thapliyal, Ashish V and Szpektor, Idan and Amelot, Julien and Chen, Xi and Soricut, Radu},
  booktitle={Findings of the Association for Computational Linguistics: EMNLP 2023},
  pages={2667--2682},
  year={2023},
  publisher={Association for Computational Linguistics},
  doi={10.18653/v1/2023.findings-emnlp.176},
  url={https://aclanthology.org/2023.findings-emnlp.176/}
}

@article{zhang2023m3exam,
  title={M3exam: A multilingual, multimodal, multilevel benchmark for examining large language models},
  author={Zhang, Wenxuan and Aljunied, Mahani and Gao, Chang and Chia, Yew Ken and Bing, Lidong},
  journal={Advances in Neural Information Processing Systems},
  volume={36},
  pages={5484--5505},
  year={2023},
  doi={10.52202/075280-0240},
  url={https://doi.org/10.52202/075280-0240}
}

@inproceedings{romero2024cvqa,
  title={Cvqa: Culturally-diverse multilingual visual question answering benchmark},
  author={Romero, David and Lyu, Chenyang and Wibowo, Haryo and Lynn, Teresa and Hamed, Injy and Kishore, Aditya and Mandal, Aishik and Dragonetti, Alina and Abzaliev, Artem and Tonja, Atnafu and Balcha, Bontu and Whitehouse, Chenxi and Salamea, Christian and Velasco, Dan and Adelani, David and Le Meur, David and Villa-Cueva, Emilio and Koto, Fajri and Farooqui, Fauzan and Belcavello, Frederico and Batnasan, Ganzorig and Vallejo, Gisela and Caulfield, Grainne and Ivetta, Guido and Song, Haiyue and Ademtew, Henok and Maina, Hern{\'a}n and Lovenia, Holy and Azime, Israel and Cruz, Jan and Gala, Jay and Geng, Jiahui and Ortiz-Barajas, Jesus-German and Baek, Jinheon and Dunstan, Jocelyn and Alemany, Laura and Nagasinghe, Kumaranage and Benotti, Luciana and D'Haro, Luis and Viridiano, Marcelo and Estecha-Garitagoitia, Marcos and Cabrera, Maria and Rodr{\'i}guez-Cantelar, Mario and Jouitteau, M{\'e}lanie and Mihaylov, Mihail and Etori, Naome and Imam, Mohamed and Adilazuarda, Muhammad and Gochoo, Munkhjargal and Otgonbold, Munkh-Erdene and Niyomugisha, Olivier and Silva, Paula and Chitale, Pranjal and Dabre, Raj and Chevi, Rendi and Zhang, Ruochen and Diandaru, Ryandito and Cahyawijaya, Samuel and G{\'o}ngora, Santiago and Jeong, Soyeong and Purkayastha, Sukannya and Kuribayashi, Tatsuki and Clifford, Teresa and Jayakumar, Thanmay and Torrent, Tiago and Ehsan, Toqeer and Araujo, Vladimir and Kementchedjhieva, Yova and Burzo, Zara and Lim, Zheng and Yong, Zheng and Ignat, Oana and Nwatu, Joan and Mihalcea, Rada and Solorio, Thamar and Aji, Alham},
  booktitle={Advances in Neural Information Processing Systems},
  volume={37},
  pages={11479--11505},
  year={2024},
  doi={10.52202/079017-0366},
  url={https://doi.org/10.52202/079017-0366}
}

@inproceedings{schneider2024m5,
  title={M5--A Diverse Benchmark to Assess the Performance of Large Multimodal Models Across Multilingual and Multicultural Vision-Language Tasks},
  author={Schneider, Florian and Sitaram, Sunayana},
  booktitle={Findings of the Association for Computational Linguistics: EMNLP 2024},
  pages={4309--4345},
  year={2024},
  publisher={Association for Computational Linguistics},
  doi={10.18653/v1/2024.findings-emnlp.250},
  url={https://aclanthology.org/2024.findings-emnlp.250/}
}

@inproceedings{vayani2024all,
  title={All languages matter: Evaluating lmms on culturally diverse 100 languages},
  author={Vayani, Ashmal and Dissanayake, Dinura and Watawana, Hasindri and Ahsan, Noor and Sasikumar, Nevasini and Thawakar, Omkar and Ademtew, Henok Biadglign and Hmaiti, Yahya and Kumar, Amandeep and Kuckreja, Kartik and Maslych, Mykola and Al Ghallabi, Wafa and Mihaylov, Mihail and Qin, Chao and Shaker, Abdelrahman M and Zhang, Mike and Ihsani, Mahardika Krisna and Esplana, Amiel and Gokani, Monil and Mirkin, Shachar and Singh, Harsh and Srivastava, Ashay and Hamerlik, Endre and Izzati, Fathinah Asma and Maani, Fadillah Adamsyah and Cavada, Sebastian and Chim, Jenny and Gupta, Rohit and Manjunath, Sanjay and Zhumakhanova, Kamila and Rabevohitra, Feno Heriniaina and Amirudin, Azril and Ridzuan, Muhammad and Kareem, Daniya and More, Ketan and Li, Kunyang and Shakya, Pramesh and Saad, Muhammad and Ghasemaghaei, Amirpouya and Djanibekov, Amirbek and Azizov, Dilshod and Jankovic, Branislava and Bhatia, Naman and Cabrera, Alvaro and Obando-Ceron, Johan and Otieno, Olympiah and Farestam, Fabian and Rabbani, Muztoba and Baliah, Sanoojan and Sanjeev, Santosh and Shtanchaev, Abduragim and Fatima, Maheen and Nguyen, Thao and Kareem, Amrin and Aremu, Toluwani and Xavier, Nathan and Bhatkal, Amit and Toyin, Hawau and Chadha, Aman and Cholakkal, Hisham and Anwer, Rao Muhammad and Felsberg, Michael and Laaksonen, Jorma and Solorio, Thamar and Choudhury, Monojit and Laptev, Ivan and Shah, Mubarak and Khan, Salman and Khan, Fahad Shahbaz},
  booktitle={2025 IEEE/CVF Conference on Computer Vision and Pattern Recognition},
  pages={19565--19575},
  year={2025},
  publisher={IEEE},
  doi={10.1109/CVPR52734.2025.01822},
  url={https://doi.org/10.1109/CVPR52734.2025.01822}
}

@inproceedings{wang2024m4u,
  title={M4U: Evaluating Multilingual Understanding and Reasoning for Large Multimodal Models},
  author={Wang, Hongyu and Xu, Jiayu and Xie, Senwei and Wang, Ruiping and Li, Jialin and Xie, Zhaojie and Zhang, Bin and Xiong, Chuyan and Chen, Xilin},
  booktitle={2026 IEEE/CVF Winter Conference on Applications of Computer Vision (WACV)},
  pages={382--392},
  year={2026},
  publisher={IEEE},
  doi={10.1109/WACV61042.2026.00045},
  url={https://doi.org/10.1109/WACV61042.2026.00045}
}

@inproceedings{yu2024cross,
  title={Cross-Lingual Text-Rich Visual Comprehension: An Information Theory Perspective},
  author={Yu, Xinmiao and Feng, Xiaocheng and Li, Yun and Liao, Minghui and Yu, Ya-Qi and Feng, Xiachong and Zhong, Weihong and Chen, Ruihan and Hu, Mengkang and Wu, Jihao and Tang, Duyu and Tu, Dandan and Qin, Bing},
  booktitle={Proceedings of the AAAI Conference on Artificial Intelligence},
  volume={39},
  pages={9680--9688},
  year={2025},
  doi={10.1609/aaai.v39i9.33049},
  url={https://doi.org/10.1609/aaai.v39i9.33049}
}

@article{chang2018graphcom,
  title={GraphCom: A multidimensional measure of graphic complexity applied to 131 written languages},
  author={Chang, Li-Yun and Chen, Yen-Chi and Perfetti, Charles A},
  journal={Behavior research methods},
  volume={50},
  pages={427--449},
  year={2018},
  publisher={Springer},
  doi={10.3758/s13428-017-0881-y},
  url={https://doi.org/10.3758/s13428-017-0881-y}
}

@inproceedings{MMBench,
  title={MMBench: Is Your Multi-modal Model an All-Around Player?},
  author={Yuan Liu and Haodong Duan and Yuanhan Zhang and Bo Li and Songyang Zhang and Wangbo Zhao and Yike Yuan and Jiaqi Wang and Conghui He and Ziwei Liu and Kai Chen and Dahua Lin},
  booktitle={Computer Vision -- ECCV 2024},
  pages={216--233},
  year={2024},
  publisher={Springer Nature Switzerland},
  doi={10.1007/978-3-031-72658-3_13},
  url={https://doi.org/10.1007/978-3-031-72658-3_13}
}

@inproceedings{fu2023mme,
  title={MME: A Comprehensive Evaluation Benchmark for Multimodal Large Language Models},
  author={Fu, Chaoyou and Chen, Peixian and Shen, Yunhang and Qin, Yulei and Zhang, Mengdan and Lin, Xu and Yang, Jinrui and Zheng, Xiawu and Li, Ke and Sun, Xing and Wu, Yunsheng and Ji, Rongrong and Shan, Caifeng and He, Ran},
  booktitle={Advances in Neural Information Processing Systems},
  volume={38},
  pages={162549--162567},
  year={2025},
  doi={10.52202/085713-4899},
  url={https://doi.org/10.52202/085713-4899}
}

@inproceedings{li2023seed,
  title={SEED-Bench: Benchmarking Multimodal Large Language Models},
  author={Li, Bohao and Ge, Yuying and Ge, Yixiao and Wang, Guangzhi and Wang, Rui and Zhang, Ruimao and Shan, Ying},
  booktitle={2024 IEEE/CVF Conference on Computer Vision and Pattern Recognition},
  pages={13299--13308},
  year={2024},
  publisher={IEEE},
  doi={10.1109/CVPR52733.2024.01263},
  url={https://doi.org/10.1109/CVPR52733.2024.01263}
}

@inproceedings{shi2022language,
  title={Language Models are Multilingual Chain-of-Thought Reasoners},
  author={Shi, Freda and Suzgun, Mirac and Freitag, Markus and Wang, Xuezhi and Srivats, Suraj and Vosoughi, Soroush and Chung, Hyung Won and Tay, Yi and Ruder, Sebastian and Zhou, Denny and Das, Dipanjan and Wei, Jason},
  booktitle={The Eleventh International Conference on Learning Representations},
  year={2023},
  url={https://openreview.net/forum?id=fR3wGCk-IXp}
}

@inproceedings{hasan-etal-2021-xl,
    title = "{XL}-Sum: Large-Scale Multilingual Abstractive Summarization for 44 Languages",
    author = "Hasan, Tahmid  and
      Bhattacharjee, Abhik  and
      Islam, Md. Saiful  and
      Mubasshir, Kazi  and
      Li, Yuan-Fang  and
      Kang, Yong-Bin  and
      Rahman, M. Sohel  and
      Shahriyar, Rifat",
    booktitle = "Findings of the Association for Computational Linguistics: ACL-IJCNLP 2021",
    month = aug,
    year = "2021",
    address = "Online",
    publisher = "Association for Computational Linguistics",
    pages = "4693--4703",
    doi = "10.18653/v1/2021.findings-acl.413",
    url = "https://aclanthology.org/2021.findings-acl.413/"
}

@inproceedings{cheng2024seeclick,
  title={SeeClick: Harnessing GUI Grounding for Advanced Visual GUI Agents},
  author={Cheng, Kanzhi and Sun, Qiushi and Chu, Yougang and Xu, Fangzhi and Li, Yantao and Zhang, Jianbing and Wu, Zhiyong},
  booktitle={Proceedings of the 62nd Annual Meeting of the Association for Computational Linguistics (Volume 1: Long Papers)},
  pages={9313--9332},
  year={2024},
  publisher={Association for Computational Linguistics},
  doi={10.18653/v1/2024.acl-long.505},
  url={https://aclanthology.org/2024.acl-long.505/}
}

@inproceedings{li2025screenspotpro,
  title        = {{ScreenSpot-Pro}: GUI Grounding for Professional High-Resolution Computer Use},
  author       = {Li, Kaixin and Meng, Ziyang and Lin, Hongzhan and Luo, Ziyang and Tian, Yuchen and Ma, Jing and Huang, Zhiyong and Chua, Tat-Seng},
  booktitle    = {Proceedings of the 33rd ACM International Conference on Multimedia},
  pages        = {8778--8786},
  year         = {2025},
  publisher    = {ACM},
  doi          = {10.1145/3746027.3755688},
  url          = {https://doi.org/10.1145/3746027.3755688}
}

@inproceedings{deng2023mind2web,
  title={Mind2Web: Towards a Generalist Agent for the Web},
  author={Deng, Xiang and Gu, Yu and Zheng, Boyuan and Chen, Shijie and Stevens, Sam and Wang, Boshi and Sun, Huan and Su, Yu},
  booktitle={Advances in Neural Information Processing Systems},
  volume={36},
  pages={28091--28114},
  year={2023},
  doi={10.52202/075280-1220},
  url={https://doi.org/10.52202/075280-1220}
}

@inproceedings{koh2024visualwebarena,
  title={VisualWebArena: Evaluating Multimodal Agents on Realistic Visual Web Tasks},
  author={Koh, Jing Yu and Lo, Robert and Jang, Lawrence and Duvvur, Vikram and Lim, Ming and Huang, Po-Yu and Neubig, Graham and Zhou, Shuyan and Salakhutdinov, Russ and Fried, Daniel},
  booktitle={Proceedings of the 62nd Annual Meeting of the Association for Computational Linguistics (Volume 1: Long Papers)},
  pages={881--905},
  year={2024},
  publisher={Association for Computational Linguistics},
  doi={10.18653/v1/2024.acl-long.50},
  url={https://aclanthology.org/2024.acl-long.50/}
}

@inproceedings{xie2024osworld,
  title={{OSWorld}: Benchmarking Multimodal Agents for Open-Ended Tasks in Real Computer Environments},
  author={Xie, Tianbao and Zhang, Danyang and Chen, Jixuan and Li, Xiaochuan and Zhao, Siheng and Cao, Ruisheng and Hua, Toh Jing and Cheng, Zhoujun and Shin, Dongchan and Lei, Fangyu and Liu, Yitao and Xu, Yiheng and Zhou, Shuyan and Savarese, Silvio and Xiong, Caiming and Zhong, Victor and Yu, Tao},
  booktitle={Advances in Neural Information Processing Systems},
  volume={37},
  pages={52040--52094},
  year={2024},
  doi={10.52202/079017-1650},
  url={https://doi.org/10.52202/079017-1650}
}

@inproceedings{liu2021marvl,
  title={Visually Grounded Reasoning across Languages and Cultures},
  author={Liu, Fangyu and Bugliarello, Emanuele and Ponti, Edoardo Maria and Reddy, Siva and Collier, Nigel and Elliott, Desmond},
  booktitle={Proceedings of the 2021 Conference on Empirical Methods in Natural Language Processing},
  pages={10467--10485},
  year={2021},
  publisher={Association for Computational Linguistics},
  doi={10.18653/v1/2021.emnlp-main.818},
  url={https://aclanthology.org/2021.emnlp-main.818/}
}

@inproceedings{tang2024mtvqa,
  title={{MTVQA}: Benchmarking Multilingual Text-Centric Visual Question Answering},
  author={Tang, Jingqun and Liu, Qi and Ye, Yongjie and Lu, Jinghui and Wei, Shu and Lin, Chunhui and Li, Wanqing and Mahmood, Mohamad Fitri Faiz Bin and Feng, Hao and Zhao, Zhen and Wang, Yanjie and Liu, Yuliang and Liu, Hao and Bai, Xiang and Huang, Can},
  booktitle={Findings of the Association for Computational Linguistics: ACL 2025},
  pages={7748--7763},
  year={2025},
  publisher={Association for Computational Linguistics},
  doi={10.18653/v1/2025.findings-acl.404},
  url={https://aclanthology.org/2025.findings-acl.404/}
}

@inproceedings{bugliarello2022iglue,
  title={{IGLUE}: A Benchmark for Transfer Learning across Modalities, Tasks, and Languages},
  author={Bugliarello, Emanuele and Liu, Fangyu and Pfeiffer, Jonas and Reddy, Siva and Elliott, Desmond and Ponti, Edoardo Maria and Vuli\'c, Ivan},
  booktitle={Proceedings of the 39th International Conference on Machine Learning},
  series={Proceedings of Machine Learning Research},
  volume={162},
  pages={2370--2392},
  year={2022},
  publisher={PMLR},
  url={https://proceedings.mlr.press/v162/bugliarello22a.html}
}

@inproceedings{liu2024ccocr,
  title={{CC-OCR}: A Comprehensive and Challenging {OCR} Benchmark for Evaluating Large Multimodal Models in Literacy},
  author={Yang, Zhibo and Tang, Jun and Li, Zhaohai and Wang, Pengfei and Wan, Jianqiang and Zhong, Humen and Liu, Xuejing and Yang, Mingkun and Wang, Peng and Bai, Shuai and Jin, Lianwen and Lin, Junyang},
  booktitle={Proceedings of the IEEE/CVF International Conference on Computer Vision (ICCV)},
  pages={21744--21754},
  year={2025},
  url={https://openaccess.thecvf.com/content/ICCV2025/html/Yang_CC-OCR_A_Comprehensive_and_Challenging_OCR_Benchmark_for_Evaluating_Large_ICCV_2025_paper.html}
}

@article{liu2024ocrbench,
  title={{OCRBench}: On the Hidden Mystery of {OCR} in Large Multimodal Models},
  author={Liu, Yuliang and Li, Zhang and Huang, Mingxin and Yang, Biao and Yu, Wenwen and Li, Chunyuan and Yin, Xu-Cheng and Liu, Cheng-Lin and Jin, Lianwen and Bai, Xiang},
  journal={Science China Information Sciences},
  volume={67},
  number={12},
  pages={220102},
  year={2024},
  doi={10.1007/s11432-024-4235-6},
  url={https://doi.org/10.1007/s11432-024-4235-6}
}

@inproceedings{chen2025mprgui,
  title={{MPR-GUI}: Benchmarking and Enhancing Multilingual Perception and Reasoning in {GUI} Agents},
  author={Chen, Ruihan and Li, Qiming and Feng, Xiaocheng and Zhong, Weihong and Yang, Xiaoliang and Gu, Yuxuan and Zhou, Zekun and Lu, Yunfei and Ren, Haoyu and Chen, Kun and Tu, Dandan and Qin, Bing},
  booktitle={Proceedings of the 64th Annual Meeting of the Association for Computational Linguistics (Volume 1: Long Papers)},
  pages={29798--29832},
  year={2026},
  publisher={Association for Computational Linguistics},
  doi={10.18653/v1/2026.acl-long.1375},
  url={https://aclanthology.org/2026.acl-long.1375/}
}

@inproceedings{yang2025macosworld,
  title={{macOSWorld}: A Multilingual Interactive Benchmark for {GUI} Agents},
  author={Yang, Pei and Ci, Hai and Shou, Mike Zheng},
  booktitle={Advances in Neural Information Processing Systems},
  volume={38},
  pages={148529--148571},
  year={2025},
  doi={10.52202/085713-4468},
  url={https://doi.org/10.52202/085713-4468}
}
